\documentclass[oneside]{amsart}
\usepackage{amsmath,amssymb,amsthm}
\usepackage{graphicx}
\usepackage[colorlinks=true, linkcolor=blue, citecolor=blue, urlcolor=blue]{hyperref}
\usepackage{tikz} 

\usepackage{empheq} 

\usepackage{longtable}
\usepackage{booktabs}
\usepackage{array}
\usepackage{amsmath} 
\usepackage{multirow} 

\usetikzlibrary{positioning, fit, calc, shapes.geometric, arrows.meta}

\usepackage{tabularx}
\usepackage{array} 

\usepackage{tikz-feynman}

\usepackage{subcaption}

\usepackage[english]{babel}
\usepackage[autostyle, english = american]{csquotes}
\MakeOuterQuote{"}

\newtheorem*{example*}{Example}

\title{Sectoral Coupling in Linguistic State Space}
\author{Sebastian Dumbrava}
\date{}

\begin{document}

\begin{abstract}
This work presents a formal framework for quantifying the internal dependencies between functional subsystems within artificial agents whose belief states are composed of structured linguistic fragments. Building on the Semantic Manifold framework, which organizes belief content into functional sectors and stratifies them across hierarchical levels of abstraction, we introduce a system of sectoral coupling constants that characterize how one cognitive sector influences another within a fixed level of abstraction. The complete set of these constants forms an agent-specific coupling profile that governs internal information flow, shaping the agent's overall processing tendencies and cognitive style. We provide a detailed taxonomy of these intra-level coupling roles, covering domains such as perceptual integration, memory access and formation, planning, meta-cognition, execution control, and affective modulation. We also explore how these coupling profiles generate feedback loops, systemic dynamics, and emergent signatures of cognitive behavior. Methodologies for inferring these profiles from behavioral or internal agent data are outlined, along with a discussion of how these couplings evolve across abstraction levels. This framework contributes a mechanistic and interpretable approach to modeling complex cognition, with applications in AI system design, alignment diagnostics, and the analysis of emergent agent behavior.
\end{abstract}

\maketitle
\setcounter{tocdepth}{1}
\tableofcontents

\section*{Executive Summary}
This work addresses the challenge of quantifying inter-functional dependencies in artificial intelligence by introducing a framework of \textbf{intra-level sectoral coupling constants} $g_{ij}^k$. Building on the Semantic Manifold framework where beliefs are structured linguistic fragments, these parameters measure the influence between cognitive subsystems (Semantic Sectors $\Sigma_s$) at a specific level of abstraction. The agent's complete \textbf{coupling profile} $G = \{g_{ij}^k\}$  serves as a quantitative signature of its internal processing architecture, shaping its cognitive dynamics and behavioral style.

\begin{enumerate}
	\item \textbf{Core Proposal and Taxonomy:} The framework formally defines the coupling constants $g_{ij}^k$ that quantify the influence pathway $$\Sigma_i^{(k)} \rightarrow \Sigma_j^{(k)}$$ exclusively within an abstraction level $k$. A comprehensive taxonomy is presented to classify the functional roles of these couplings.
	
	\item \textbf{Cognitive Signatures and Dynamics:} It demonstrates how the coupling profile $G$ determines an agent's cognitive style (e.g., "reactive" vs. "deliberative")  and gives rise to emergent dynamics like feedback loops. These phenomena result from the interplay between intra-level couplings and inter-level operators such as Abstraction ($\Lambda$) and Elaboration ($V$).
	
	\item \textbf{Empirical Measurement:} To ground the framework, this work outlines methodologies for measuring and inferring $g_{ij}^k$ values from various forms of agent data, including internal logs and observable behavior. A detailed, three-phase procedural framework for their empirical estimation is also provided.
	
	\item \textbf{Dynamic Coupling Profiles:} The framework investigates the dynamics of the coupling profile itself. It introduces "coupling propagation" to describe how the effective profile $G^{(k)}$ transforms across scales, and analyzes how perturbations $$G \rightarrow G'$$ can alter cognitive processing.
	
	\item \textbf{Significance and Applications:} The framework provides a mechanistic and interpretable model of agent cognition. This has significant implications for principled AI design, AI safety, diagnostics, and governance, by enabling the analysis of internal feedback loops and potential failure modes.
\end{enumerate}

\bigskip

As artificial intelligence systems grow more complex, this framework offers a critical tool for ensuring their safety and interpretability by quantifying their internal processing architecture. By defining an agent's "cognitive signature" through its coupling profile, it becomes possible to analyze emergent dynamics like feedback loops and diagnose potential failure modes. This foundational understanding is vital for a future in which an agent's cognitive style can be reliably controlled, leading to the development of safer, more predictable, and governable systems.

\newpage

\section{Introduction}
\label{sec:introduction}

\subsection{The Challenge: Quantifying Inter-functional Dependencies in AI Cognition}
\label{subsec:intro_challenge}
The ongoing advancement of artificial intelligence (AI) necessitates the development of increasingly sophisticated agents capable of complex reasoning, robust adaptation, and nuanced interaction within dynamic environments. A significant challenge in this endeavor lies in creating cognitive models that are not only computationally powerful but also offer clear interpretability of their internal states and the mechanisms governing their belief evolution. While existing frameworks provide valuable insights, a persistent gap remains in precisely characterizing and quantifying the inter-functional dependencies that orchestrate complex cognitive behaviors within these agents. Understanding how different cognitive capacities---such as perception, planning, memory, and reflection---systematically influence one another is crucial for explaining emergent cognitive behaviors, designing truly integrated AI, and ensuring their safe and aligned operation.

\subsection{Foundation: The Semantic Manifold Framework}
\label{subsec:intro_foundation}
This work builds upon the Semantic Manifold framework set forth in \textit{Theoretical Foundations for Semantic Cognition in Artificial Intelligence} \cite{Dumbrava2025TheoreticalFoundations}, which offers a cognitively motivated architecture for agents whose belief states ($\phi$) are conceptualized as dynamic ensembles of natural language fragments ($\varphi_{i}$). These fragments are organized within a structured linguistic state space ($\Phi$) according to functional Semantic Sectors ($\Sigma_{s}$) that categorize beliefs by their cognitive role (e.g., $\Sigma_{\text{perc}}$ for perception, $\Sigma_{\text{plan}}$ for planning) and hierarchical Abstraction Layers ($k$) that stratify beliefs by their level of generality. A detailed review of this linguistic, structured approach to interpretable cognition is provided in Section~\ref{sec:background_sm}, establishing the substrate for our model of sectoral influence.

\subsection{Core Proposal: Intra-Level Sectoral Coupling Constants ($g_{ij}^k$) and the Set of Couplings ($G$)}
\label{subsec:intro_core_proposal}
To address the challenge of quantifying inter-functional dependencies, this work introduces the concept of \textbf{intra-level sectoral coupling constants}, denoted $g_{ij}^k$. These parameters are proposed to quantify the strength and nature of influence exerted by a source Semantic Sector $\Sigma_i$ on a target Semantic Sector $\Sigma_j$, exclusively at a specific abstraction level $k$. The complete collection of these intra-level couplings, $G = \{g_{ij}^k\}$, forms an agent-specific coupling profile, defining its characteristic pattern of internal interactions within each abstraction layer. The formal definition of $g_{ij}^k$ and how they relate to dedicated inter-level cognitive operators (like Abstraction $\Lambda$ and Elaboration $V$) that mediate influences across different abstraction levels are detailed in Section~\ref{sec:architecture_of_influence}.

\subsection{Central Thesis: The Set of Couplings $G$ as a Primary Determinant of Cognitive Dynamics and Style}
\label{subsec:intro_thesis}
Our central thesis is that an agent's cognitive dynamics, its internal processing pathways, and its distinctive processing "style" can be quantitatively characterized and understood through its specific profile of intra-level sectoral coupling constants $G$. We posit that this set $G$ is a primary determinant of how established Semantic Manifold cognitive operators function and how the agent's belief state $\phi$ evolves, shaping information flow and cognitive tendencies. The detailed mechanisms by which $G$ gives rise to diverse cognitive styles are explored in Sections~\ref{sec:architecture_of_influence} and \ref{sec:cognitive_signatures}.

\subsection{Significance of the Proposed Framework}
\label{subsec:intro_significance}
The framework centered on intra-level sectoral coupling constants $g_{ij}^k$ offers substantial significance for advancing artificial cognition:
\begin{itemize}
	\item It provides a quantitative method for a more precise and mechanistic understanding of functional dependencies, primarily direct intra-level influences and mediated inter-level processes, governing belief evolution.
	\item It enables richer agent characterization through $G$ profiles, potentially improving predictive power regarding belief evolution and behavior.
	\item It offers a conceptual basis for the principled design of AI agents with specific cognitive profiles.
	\item It can contribute to AI safety, alignment, and governance by improving the understanding of $G$, aiding in diagnosing problematic cognitive dynamics, informing interventions, and potentially guiding the design of more stable and governable agents.
	\item It enhances the testability of hypotheses by defining measurable parameters.
\end{itemize}
These contributions are developed in subsequent sections and further discussed in Section~\ref{sec:discussion}.

\subsection{Overview of Contributions and Structure}
\label{subsec:intro_roadmap}
This work systematically develops this coupling-focused framework. Section~\ref{sec:background_sm} provides a concise review of the Semantic Manifold architecture, emphasizing its role as a dynamic substrate for belief. Section~\ref{sec:architecture_of_influence} formally defines the intra-level sectoral coupling constants $g_{ij}^k$ and the set $G$, and presents a comprehensive taxonomy classifying the diverse roles these intra-level couplings play. It also clarifies how inter-level transitions are handled. Section~\ref{sec:cognitive_signatures} explores how distinct coupling profiles $G$ give rise to varied "cognitive styles," emergent system dynamics, and characteristic interaction pathways, including a conceptual approach to visualizing feedback loops. Section~\ref{sec:measuring_g} details methodologies for measuring or inferring these coupling constants, including a procedural framework for their empirical estimation, and discusses key challenges. Section~\ref{sec:dynamics_of_G} examines the dynamics of the coupling profile $G$ itself, including the scale-dependence of intra-level coupling profiles $G^{(k)}$ and susceptibility to perturbations. Finally, Section~\ref{sec:discussion} discusses the implications, strengths, and limitations of the framework, Section~\ref{sec:future_directions} outlines promising future research directions, and Section~\ref{sec:conclusion} concludes by summarizing this work's contributions towards a more quantitative and mechanistic understanding of agent cognition. To complement this systematic development, the appendices provide a rich set of supplementary materials, including expanded taxonomies, illustrative cognitive processing sequences, deeper discussions of coupling dynamics, and a discussion of future architectural development.

\section{The Semantic Manifold Framework}
\label{sec:background_sm}

The study of sectoral coupling presented in this work operates upon the architectural principles set forth in the Semantic Manifold framework \cite{Dumbrava2025TheoreticalFoundations}.
This framework provides a comprehensive, cognitively-inspired model for how an artificial agent's internal belief states are structured, represented, and dynamically processed.
This section briefly reviews its essential components, emphasizing those aspects that form the necessary substrate for understanding how intra-level sectoral couplings $g_{ij}^k$ govern the interactions within this architecture.
Key components to be reviewed include:
\begin{enumerate}
	\item how an agent's belief state is represented as linguistic fragments within a structured state space (Subsection~\ref{subsec:belief_representation});
	\item the organizational principles of functional Semantic Sectors and hierarchical Abstraction Layers (Subsection~\ref{subsec:organizational_principles});
	\item the core cognitive operators that drive belief evolution (Subsection~\ref{subsec:core_operators});
	\item and the mechanisms for meta-cognition and self-regulation (Subsection~\ref{subsec:meta_cognition}).
\end{enumerate}

\subsection{Belief State Representation ($\phi = \{\varphi_i\}$ in $\Phi$)}
\label{subsec:belief_representation}
Central to the Semantic Manifold framework is the agent's belief state $\phi$ comprised of individual linguistic belief fragments $\varphi_i$, denoted as $$\phi=\{\varphi_1, \varphi_2, \dots, \varphi_n\}$$ (where singleton belief states, comprised of single fragments, may be identified with the fragment itself).  Each fragment $\varphi_i$ is a natural language expression, e.g., 
\begin{enumerate}
	\item $\varphi' =$ ``Object detected at coordinates (x,y)''
	\item $\varphi'' =$ ``Current objective is to reach target Z''
	\item $\varphi''' =$ ``Previous attempt resulted in error E''
\end{enumerate}
This linguistic basis ensures that belief content is inherently interpretable.  The entire space of all possible belief configurations an agent can entertain constitutes the semantic state space, $\Phi$.  The internal organization of $\phi$ reflects semantic relationships and functional roles of its constituent fragments. 

\subsection{Organizational Principles: Semantic Sectors ($\Sigma_s$) and Abstraction Layers ($k$)}
\label{subsec:organizational_principles}
The semantic state space $\Phi$ is not uniform but is organized along two primary dimensions to facilitate structured cognition: 
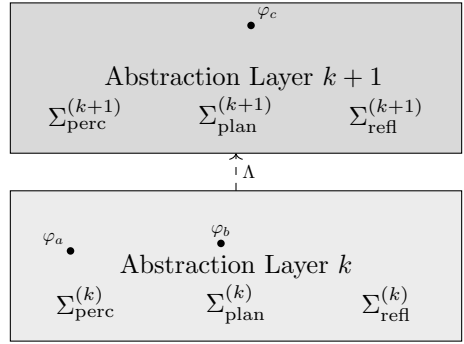
\begin{figure}[htbp]
	\centering
	
	\begin{tikzpicture}

		\draw[fill=gray!15, draw=black] (0,0) rectangle (6,2); 
		\node at (3,1) {Abstraction Layer $k$};
		\node at (1,0.5) {$\Sigma_{\text{perc}}^{(k)}$};
		\node at (3,0.5) {$\Sigma_{\text{plan}}^{(k)}$};
		\node at (5,0.5) {$\Sigma_{\text{refl}}^{(k)}$}; 
		\fill (0.8,1.2) circle (0.05) node[above left, scale=0.7] {$\varphi_a$};
		\fill (2.8,1.3) circle (0.05) node[above, scale=0.7] {$\varphi_b$}; 
		
		\draw[fill=gray!30, draw=black] (0,2.5) rectangle (6,4.5); 
		\node at (3.1,3.5) {Abstraction Layer $k+1$};
		\node at (1,3) {$\Sigma_{\text{perc}}^{(k+1)}$}; 
		\node at (3,3) {$\Sigma_{\text{plan}}^{(k+1)}$};
		\node at (5,3) {$\Sigma_{\text{refl}}^{(k+1)}$};
		\fill (3.2,4.2) circle (0.05) node[above right, scale=0.7] {$\varphi_c$}; 
		
		\draw[->, dashed] (3,2) -- (3,2.5) node[midway, right, scale=0.7] {$\Lambda$};
	\end{tikzpicture}
	\caption{Belief states are organized along two primary dimensions: functional Semantic Sectors ($\Sigma_s$) categorizing beliefs by cognitive role (e.g., perception $\Sigma_{\text{perc}}$, planning $\Sigma_{\text{plan}}$, reflection $\Sigma_{\text{refl}}$), and hierarchical Abstraction Layers ($k$) stratifying beliefs by generality.  Individual belief fragments ($\varphi_i$) reside within specific sector-layer locations, with operators like Abstraction ($\Lambda$) mediating inter-level transitions. }
	\label{fig:semantic_manifold_architecture}
\end{figure}

\begin{itemize}
	\item \textbf{Semantic Sectors ($\Sigma_s$)}: Belief fragments are categorized by their primary cognitive function into different Semantic Sectors.  Key examples include $\Sigma_{\text{perc}}$ for perceptual beliefs derived from sensory input, $\Sigma_{\text{plan}}$ for goals and action plans, $\Sigma_{\text{mem}}$ for retrieved memories, and $\Sigma_{\text{refl}}$ for meta-cognitive beliefs and self-assessments.  This sectoral organization allows for functional modularity in processing. 
	\item \textbf{Abstraction Layers ($k$ or $\Phi^{(k)}$)}: Beliefs are also stratified by their level of semantic generality or abstraction, indexed by $k$.  Low $k$ values (e.g., $k=0$) typically correspond to concrete, specific, often sensor-grounded information, while higher $k$ values ($k \gg 0$) represent more abstract generalizations, principles, or schemas.  The overall belief space is the union of these layers: $$\Phi = \bigcup_k \Phi^{(k)}$$ (and, therefore, belief states $\phi$ can be stratified analogously: $\phi = \bigcup_k \phi^{(k)}$). 
\end{itemize}

Thus, each belief fragment $\varphi_i$ can be conceptually located by its primary sector(s) and abstraction level $(\Sigma, k)$, providing a structured "address" within the cognitive architecture. 

\subsection{Core Cognitive Operators}
\label{subsec:core_operators}
The belief state $\phi$ evolves dynamically through the action of a suite of core cognitive operators defined within the Semantic Manifold framework.  These operators manipulate and transform belief fragments. Key operators include:

\begin{itemize}
	\item \textbf{Observation Encoding ($X$)}: Converts raw sensory inputs into initial linguistic belief fragments (\cite{Dumbrava2025TheoreticalFoundations}, Chapter 7). 
	\item \textbf{Abstraction ($\Lambda$) and Elaboration ($V$)}: Mediate movement between abstraction layers $\Phi^{(k)}$; $\Lambda$ abstracts to higher $k$, while $V$ elaborates to lower $k$ (\cite{Dumbrava2025TheoreticalFoundations}, Chapter 10). 
	\item \textbf{Assimilation ($A$)}: Integrates new information (from $X$ or internal processes) into the existing belief state, with mechanisms for corrective ($A_{\text{corr}}$) and elaborative ($A_{\text{elab}}$) integration (\cite{Dumbrava2025TheoreticalFoundations}, Chapter 13). 
	\item \textbf{Nullification ($N_t$)}: Models the gradual decay or forgetting of belief fragments based on their anchoring strength and lack of reinforcement (\cite{Dumbrava2025TheoreticalFoundations}, Chapter 14). 
	\item \textbf{Annihilation ($K, K_{\Sigma}$)}: Represents the abrupt erasure of belief structures, either globally ($K$) or for specific sectors ($K_{\Sigma}$) (\cite{Dumbrava2025TheoreticalFoundations}, Chapter 15).
	\item \textbf{Spontaneous Drift ($D$)}: Allows for minimal, undirected semantic motion, particularly from states of cognitive quiescence (\cite{Dumbrava2025TheoreticalFoundations}, Chapter 16). 
	
\end{itemize}

\begin{figure}[htbp]
	\centering
	\begin{tikzpicture}
		\begin{feynman}
			
			\vertex (vA) [draw, shape=circle, minimum size=10mm, line width=0.8pt] at (0, 0) {$\Lambda$}; 
			\vertex (vE) [draw, shape=circle, minimum size=10mm, line width=0.8pt] at (0, 2.5) {$V$};

			\vertex (p1_in) at (-2, -1.5); 
			\vertex (p2_in) at (2, -1.5);
			\vertex (p3_in_source) at (2, 0.5);
			\vertex (p4_out) at (0, 4.2); 
			
			\diagram* {
				(p1_in) -- [fermion] (vA),
				(p2_in) -- [fermion] (vA),
				(vA) -- [scalar, edge] (vE),
				(p3_in_source) -- [fermion] (vE),
				(vE) -- [fermion] (p4_out),
			}; 
			
			\node[below left=0.1cm of p1_in, text width=3.2cm, align=center, font=\footnotesize] 
			{$\varphi_1 = \text{``Red object detected''}$}; 
			\node[below right=0.1cm of p2_in, text width=3.2cm, align=center, font=\footnotesize] 
			{$\varphi_2 = \text{``Ball detected''}$}; 
			\node[right=0.2cm of p3_in_source, text width=5.4cm, align=left, font=\footnotesize]
			{$\varphi_3 = \text{``Only one object detected''}$}; 
			\node[above=0.15cm of p4_out, text width=3.6cm, align=center, font=\footnotesize] 
			{$\varphi_4 = \text{``Red ball detected''}$};

			\node[below=0.25cm of vA, font=\small\bfseries] {}; 
			\node[above=0.25cm of vE, font=\small\bfseries] {};

			\draw[-{Stealth[length=3mm, width=2mm]}, thick, black] 
			(-5.5, -2.0) -- (-5.5, 4.7) 
			node[midway, right, rotate=90, anchor=south, font=\small] {Time};
			
		\end{feynman}
	\end{tikzpicture}
	\caption{Visual analogy of belief fragment transformation involving inter-level operators Abstraction ($\Lambda$) and Elaboration ($V$).  Incoming concrete fragments (e.g., $\varphi_1, \varphi_2$ from $\Sigma_{\text{perc}}^{(0)}$) are processed by $\Lambda$.  The resulting abstracted content is then processed by $V$ along with other contextual information (e.g., $\varphi_3$) to produce an output fragment (e.g., $\varphi_4$ in $\Sigma_{\text{perc}}^{(0)}$). }
	\label{fig:feynman_abstraction_elaboration_main}
\end{figure}
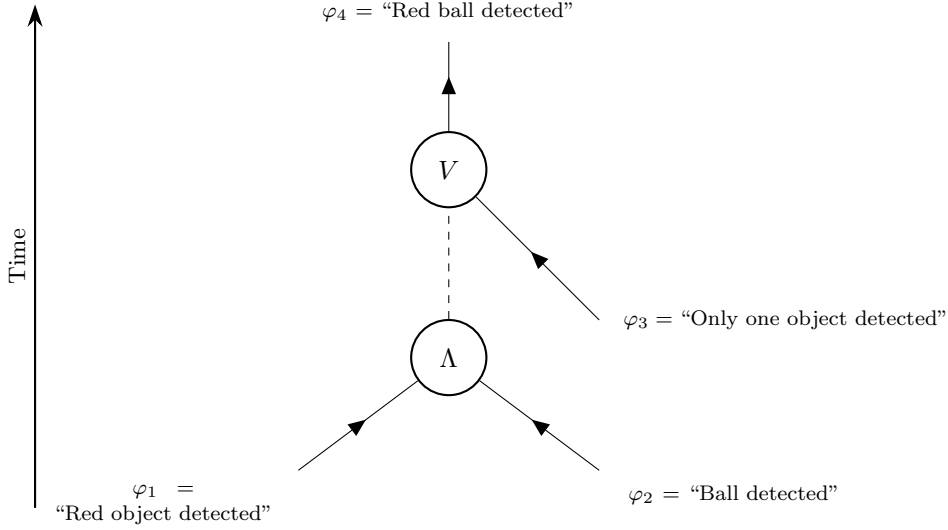

These operators are the engines of belief change.  As will be detailed in Section~\ref{sec:architecture_of_influence}, the intra-level sectoral coupling constants $g_{ij}^k$ introduced in this work are proposed to measure the dynamics, efficacy, and interaction of these fundamental operators within their respective abstraction layers.

\subsection{Meta-Cognition and Regulatory Functions}
\label{subsec:meta_cognition}
The Semantic Manifold incorporates mechanisms for meta-cognition and self-regulation, primarily centered around the reflective sector, $\Sigma_{\text{refl}}$. 
\begin{itemize}
	\item The \textbf{Meta-Assimilation ($M$)} operator integrates internally generated introspective information (e.g., about current coherence $\kappa$, cognitive load $\lambda$, or semantic effort $\epsilon$) into $\Sigma_{\text{refl}}$ as explicit meta-beliefs (\cite{Dumbrava2025TheoreticalFoundations}, Chapter 28). 
	\item This capacity for self-monitoring underpins \textbf{regulatory loops}, where meta-control policies ($\pi_{\text{regulate}}$) and effort allocation policies ($\pi_{\text{effort}}$) can adjust cognitive operations and resource distribution based on these reflective assessments (\cite{Dumbrava2025TheoreticalFoundations}, Chapters 29 and 32). 
\end{itemize}
The effectiveness of such regulatory functions critically depends on the timely and accurate flow of information between different Semantic Sectors (e.g., from operational sectors to $\Sigma_{\text{refl}}$ and back).  The intra-level sectoral couplings $g_{ij}^k$, by modulating operator dynamics within specific levels, are fundamental to governing this internal communication (e.g., how information is prepared for inter-level transfer by operators like $\Lambda$ or $V$, or processed after such transfers) and thus are integral to the agent's capacity for robust self-regulation.  

This architecture---characterized by interpretable linguistic beliefs, organized by sectors and abstraction layers, and driven by defined cognitive operators and meta-cognitive regulation---provides the dynamic substrate upon which the detailed analysis of sectoral couplings will be developed in the subsequent sections.

\subsection{Section Summary: The Semantic Manifold as a Substrate}
\label{subsec:summary_sec2}
This section reviewed the essential components of the Semantic Manifold framework, which provides the architectural foundation for this work's analysis of sectoral couplings. The key principles establishing this substrate are:
\begin{enumerate}
	\item \textbf{Linguistic Belief Representation:} An agent's belief state, $\phi$, is an interpretable ensemble of natural language fragments, $\varphi_i$, residing within a structured state space, $\Phi$.
	\item \textbf{Dual Organizational Structure:} Beliefs are organized along two primary dimensions: functional \textbf{Semantic Sectors} ($\Sigma_s$) that group beliefs by cognitive role (e.g., perception, planning) , and hierarchical \textbf{Abstraction Layers} ($k$) that stratify beliefs by their level of generality.
	\item \textbf{Core Cognitive Operators:} The belief state evolves through the action of a suite of operators that manipulate and transform belief fragments. These include operators for moving between abstraction layers, such as Abstraction ($\Lambda$) and Elaboration ($V$) , and for integrating new information, such as Assimilation ($A$).
	\item \textbf{Meta-Cognitive Regulation:} The framework incorporates mechanisms for self-monitoring and regulation, primarily centered in the reflective sector ($\Sigma_{\text{refl}}$). This allows the agent to form meta-beliefs about its own internal states and adjust its cognitive operations accordingly.
\end{enumerate}
Together, these components establish the dynamic and structured substrate upon which the interactions quantified by intra-level coupling constants operate.

\section{Intra-Level Sectoral Couplings}
\label{sec:architecture_of_influence}

Building upon the Semantic Manifold framework reviewed in Section~\ref{sec:background_sm}, this section introduces the core theoretical construct of this work: intra-level sectoral coupling constants. These constants provide a formal and quantitative means to describe the intricate web of influences between different cognitive functions (Semantic Sectors) within an agent's belief state, specifically focusing on interactions that occur within each abstraction level. We will:
\begin{enumerate}
	\item define these intra-level couplings (Subsection~\ref{subsec:formal_definition_g_G});
	\item and present a detailed taxonomy classifying their diverse functional roles (Subsection~\ref{subsec:taxonomy_couplings}).
\end{enumerate}
This is crucial for understanding how an agent's specific profile of sectoral couplings $G$ is formed and how it operates to measure overall cognitive processing, which involves both intra-level and inter-level dynamics.

\subsection{Formal Definition of Intra-Level Couplings $g_{ij}^k$ and the Set $G$}
\label{subsec:formal_definition_g_G}

A cornerstone of our framework for understanding integrated cognition is the formalization of interactions between different Semantic Sectors. We posit that direct interactions within a given abstraction level can be quantified by \textbf{intra-level sectoral coupling constants}, denoted $g_{ij}^k$ (or $g_{{\Sigma_{i}} \rightarrow \Sigma_{j}}^k$).

\begin{figure}[htbp]
	\centering
\begin{tikzpicture}[
	sector/.style={rectangle, draw, fill=gray!15, minimum width=3.5cm, minimum height=1.8cm, rounded corners, align=center}
	]
	\node at (5,3.2) [above, font=\bfseries] {Abstraction Layer $k$}; 
	\draw[dashed] (0,0) rectangle (10,2.8); 
	
	\node[sector] (sigma_i) at (2.5,1.4) {$\Sigma_i^{(k)}$}; 
	
	\node[sector] (sigma_j) at (7.5,1.4) {$\Sigma_j^{(k)}$}; 
	
	\draw[->, thick, black] (sigma_i.east) -- (sigma_j.west) node[midway, above, sloped, font=\small, yshift=1mm] {$g_{ij}^k$};
\end{tikzpicture}
	\caption{Conceptual illustration of an intra-level sectoral coupling constant $g_{ij}^k$. This constant quantifies the strength and nature of influence from a source Semantic Sector $\Sigma_i^{(k)}$  to a target Semantic Sector $\Sigma_j^{(k)}$, exclusively within a specific abstraction level $k$.}
	\label{fig:intra_level_coupling_gij}
\end{figure}
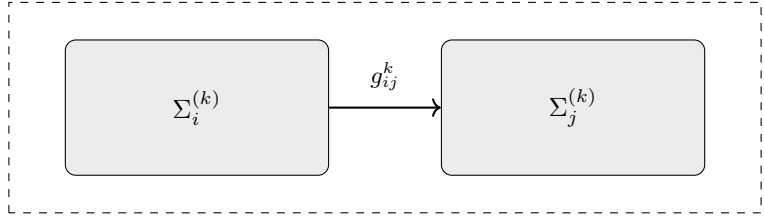

\begin{itemize}
	\item The intra-level coupling constant $g_{ij}^k$ is defined as a parameter that quantifies the \textbf{strength and } (e.g., excitatory, inhibitory, or a more complex modulatory effect) of the influence exerted by the state or activity within a source Semantic Sector $\Sigma_i$ at abstraction level $k$ (denoted $\Sigma_i^{(k)}$) on the state, activity, or operator dynamics within a target Semantic Sector $\Sigma_j$ also at abstraction level $k$ (denoted $\Sigma_j^{(k)}$).
	\item The superscript $k$ strictly indicates that this coupling operates within that specific abstraction level $k$, i.e., describing an influence pathway $$\Sigma_i^{(k)} \rightarrow \Sigma_j^{(k)}.$$
	\item The strict intra-level definition of $g_{ij}^k$ ensures a clear separation of concerns: $g_{ij}^k$ quantifies influence within a level, while operators like $\Lambda$ and $V$ are responsible for transformations between levels.
	\item The complete set of these intra-level constants, $$G = \{g_{ij}^k\}$$ for all relevant sector pairs $(i,j)$ and pertinent abstraction levels $k$, forms the agent's set of couplings or \textbf{profile}. This set $G$ is a fundamental characteristic of the agent's cognitive architecture, which can potentially be considered part of its specific parameterization $\theta$ (as discussed in \cite{Dumbrava2025TheoreticalFoundations}, Chapter 8), and defines its specific cognitive tendencies (i.e., the couplings should have close-to-baseline values near baseline belief states $\phi$).
	\item While the coupling constants are intrinsically functions of the belief state, $$g_{ij}^k = g_{ij}^k(\phi),$$ this dependence is typically omitted herein for notational simplicity.
\end{itemize}

\begin{example*}[Drone Example]
	\normalfont
	To make this concept more tangible, consider an autonomous drone. A specific intra-level coupling, $g_{\Sigma_{\text{perc}} \rightarrow \Sigma_{\text{plan}}}^0$, might quantify how strongly a concrete perceptual input like $$\varphi_x = \text{"Large obstacle detected 10m ahead"} \in \Sigma_{\text{perc}}^{(0)}$$ influences the immediate generation of a concrete planning response $$\varphi_y = \text{"Initiate upward thrust and halt forward motion"} \in \Sigma_{\text{plan}}^{(0)}.$$ A high positive value for this coupling would indicate a rapid, obligatory planning response within level $k=0$, characteristic of a reactive safety system. Conversely, a low or near-zero value for $g_{\Sigma_{\text{perc}} \rightarrow \Sigma_{\text{plan}}}^0$ might characterize a more deliberative drone that first processes this information within $\Sigma_{\text{perc}}^{(0)}$ before an action is decided upon and elaborated back to $\Sigma_{\text{plan}}^{(0)}$. 
\end{example*}
The set of couplings $G$ thus measures how information flows between specialized cognitive areas within each abstraction layer and how integrated cognitive processes emerge from these interactions in conjunction with inter-level operators.

\subsection{A Taxonomy of Key Sectoral Coupling Roles}
\label{subsec:taxonomy_couplings}

To provide a more nuanced understanding of how the set of intra-level couplings $G = \{g_{ij}^k\}$ shapes cognition, we propose a taxonomy of sectoral coupling roles. This classification is based on the broad functional roles these intra-level couplings play in the agent's cognitive economy, drawing upon the sectoral distinctions defined in the Semantic Manifold framework \cite{Dumbrava2025TheoreticalFoundations}. Table~\ref{tab:sectoral_couplings} provides a summary of these roles. The detailed characteristics, mechanisms, and impacts of each specific coupling type are elaborated in Appendix~\ref{app:detailed_coupling_roles}.

\begin{table}[h]
	\centering
	\caption{Taxonomy of Intra-Level Sectoral Couplings \( g^k_{ij} \)}
	\label{tab:sectoral_couplings}
	\renewcommand{\arraystretch}{1.3}
	\begin{tabularx}{\textwidth}{|>{\raggedright\arraybackslash}p{4.8cm}|X|}
		\hline
		\textbf{Coupling Role} & \textbf{Functional Description} \\
		\hline
		\textbf{Perceptual Integration} (\( g^k_{\Sigma_{\text{perc}} \rightarrow \Sigma_X} \)) & 
		Channels perceptual input into downstream sectors.  \\
		\hline
		\textbf{Memory Interaction} (\( g^k_{\Sigma_X \rightarrow \Sigma_{\text{mem}}} \)) &
		Measures formation and retrieval of memory structures.  \\
		\hline
		\textbf{Planning and Goals} (\( g^k_{\Sigma_X \leftrightarrow \Sigma_{\text{plan}}} \)) & 
		Links sector state to plan formation and propagates plan-based control over other sectors. \\
		\hline
		\textbf{Meta-Cognitive Regulation} (\( g^k_{\Sigma_X \leftrightarrow \Sigma_{\text{refl}}} \)) & 
		Supports self-monitoring (afferent) and reflective modulation (efferent) via meta-control policies. \\
		\hline
		\textbf{Execution Control} (\( g^k_{\Sigma_{\text{plan}}, \Sigma_{\text{refl}} \rightarrow \Sigma_{\text{exe}}} \)) & 
		Finalized plans and meta-decisions activate execution processes at level \( k \). \\
		\hline
		\textbf{Communicative Output} (\( g^k_{\Sigma_X \rightarrow \Sigma_{\text{lang-out}}} \)) & 
		Selects and expresses internal beliefs linguistically based on salience and intent. \\
		\hline
		\textbf{Intra-Sectoral Recurrence} (\( g^k_{ii} \)) &
		Encodes self-reinforcement within a sector; supports focus, persistence, or leads to rumination. \\
		\hline
		\textbf{Affective-Cognitive Modulation} (\( g^k_{\Sigma_X \leftrightarrow \Sigma_{\text{affect}}} \)) & 
		Affective states bias cognition or are triggered by cognitive activity; modulates attentional patterns.  \\
		\hline

	\end{tabularx}
\end{table}

\noindent\textbf{I. Input Processing \& World-Model Integration Couplings} 
These intra-level couplings govern how the agent internalizes and interprets information from its environment and its own embodiment within a specific abstraction layer $k$, forming and updating its internal model of the world at that layer.

\subsubsection{Perceptual Integration Couplings ($g_{\Sigma_{\text{perc}} \rightarrow \Sigma_X}^k$)}
\label{ssubsec:taxonomy_perceptual_integration}
Perceptual Integration Couplings quantify the influence of the perceptual sector ($\Sigma_{\text{perc}}^{(k)}$) on other cognitive sectors ($\Sigma_X^{(k)}$). They are crucial for grounding other cognitive functions in the current environmental context relevant to that level, triggering specific intra-level responses, and providing the raw material for learning and memory formation. A detailed exposition of their characteristics and mechanisms is provided in Appendix~\ref{app:detail_perceptual_integration}.

\noindent\textbf{II. Deliberative, Reasoning, and Internal Processing Couplings}
These intra-level couplings define pathways for internal thought within a specific abstraction layer $k$, including memory access, planning, and reflection relevant to that level.

\subsubsection{Memory Interaction Couplings ($g_{\Sigma_X \leftrightarrow \Sigma_{\text{mem/narr}}}^k$)}
\label{ssubsec:taxonomy_memory_interaction}
These couplings quantify the critical processes of memory formation and utilization, describing the flow of information between operational sectors ($\Sigma_X^{(k)}$) and the agent's memory systems (e.g., $\Sigma_{\text{mem}}^{(k)}, \Sigma_{\text{narr}}^{(k)}$). They are fundamental for learning from experience, adapting behavior, and grounding current cognition in past events. See Appendix~\ref{app:detail_memory_formation} for full details.

\subsubsection{Planning \& Goal Processing Couplings ($g_{\Sigma_X \leftrightarrow \Sigma_{\text{plan}}}^k$)}
\label{ssubsec:taxonomy_planning_goal}
These couplings connect various sectors $\Sigma_X^{(k)}$ with the planning sector $\Sigma_{\text{plan}}^{(k)}$. They measure how inputs from $\Sigma_X^{(k)}$ shape planning dynamics and, conversely, measure how active plans in $\Sigma_{\text{plan}}^{(k)}$ direct or bias processing in other sectors $\Sigma_X^{(k)}$. See Appendix~\ref{app:detail_planning_goal} for full details. 

\subsubsection{Reflective \& Meta-Cognitive Couplings ($g_{\Sigma_X \leftrightarrow \Sigma_{\text{refl}}}^k$)}
\label{ssubsec:taxonomy_reflective_metacognitive}
These couplings measure the interaction between operational sectors $\Sigma_X^{(k)}$ and the reflective sector $\Sigma_{\text{refl}}^{(k)}$. Afferent couplings measure meta-belief formation in $\Sigma_{\text{refl}}^{(k)}$ based on activity in $\Sigma_X^{(k)}$, while efferent couplings measure how meta-beliefs from $\Sigma_{\text{refl}}^{(k)}$ guide and regulate processing in $\Sigma_X^{(k)}$. For a comprehensive discussion, see Appendix~\ref{app:detail_reflective_metacognitive}.

\noindent\textbf{III. Output Generation and Execution-Related Couplings}
These intra-level couplings bridge internal cognitive states at a specific abstraction level $k$ with external actions or communications initiated from that level.

\subsubsection{Execution \& Action Control Couplings ($g_{(\Sigma_{\text{plan}}/\Sigma_{\text{refl}}) \rightarrow \Sigma_{\text{exe}}}^k$)}
\label{ssubsec:taxonomy_execution_control}
These couplings quantify how finalized plans from $\Sigma_{\text{plan}}^{(k)}$ or decisive meta-beliefs from $\Sigma_{\text{refl}}^{(k)}$ influence the action execution system $\Sigma_{\text{exe}}^{(k)}$. They measure the translation of internal intentions at level $k$ into overt actions or inputs for possibly further elaboration. For more details, refer to Appendix~\ref{app:detail_execution_control}. 

\subsubsection{Communicative Output Couplings ($g_{\Sigma_X \rightarrow \Sigma_{\text{lang-out}}}^k$)}
\label{ssubsec:taxonomy_communicative_output}
These couplings measure how internal beliefs from a source sector $\Sigma_X^{(k)}$ are selected, potentially transformed, and conveyed to a language output sector $\Sigma_{\text{lang-out}}^{(k)}$, enabling the agent to communicate. See Appendix~\ref{app:detail_communicative_output} for a detailed discussion. 

\noindent\textbf{IV. Systemic and Adaptive Couplings}
These couplings affect the cognitive stability or long-term adaptation within specific abstraction layers or across the system.

\subsubsection{Intra-Sectoral Recurrent Couplings ($g_{ii}^k$)}
\label{ssubsec:taxonomy_intra_sectoral}
Recurrent couplings describe the influence of a sector $\Sigma_i^{(k)}$ on its own subsequent states, which contributes to maintaining focus, cognitive inertia, pattern completion, or stability. Refer to Appendix~\ref{app:detail_intra_sectoral} for details.

\begin{figure}[htbp]
	\centering
\begin{tikzpicture}[scale=0.7,
	node_dist_v/.initial=5cm,    
	node_dist_h/.initial=2.5cm,  
	node_dist_offset/.initial=2cm, 
	sector/.style={draw, rounded corners=6pt, minimum width=3cm, minimum height=1.2cm, font=\small, align=center, fill=gray!20},
	sector_k0/.style={fill=gray!10}, 
	sector_k0_special/.style={fill=gray!40}, 
	arrow/.style={->, >=Stealth},
	loop/.style={->, thick, >=Stealth, looseness=10},
	level_label/.style={font=\scriptsize\itshape, draw=none, align=right}, 
	edge_label/.style={font=\tiny, midway, auto, inner sep=1pt}, 
	operator_label/.style={font=\tiny, midway, auto, inner sep=1pt} 
	]

	\node (refl1) [sector] at (7.9, 5.0) {\( \Sigma^{(k+1)}_{\text{refl}} \)};
	\node (plan1) [sector, right=2.5 of refl1] {\( \Sigma^{(k+1)}_{\text{plan}} \)};

	\node (perc0) [sector, sector_k0] at (0,0) {\( \Sigma^{(k)}_{\text{perc}} \)};
	\node (plan0) [sector, sector_k0, right=2.5 of perc0] {\( \Sigma^{(k)}_{\text{plan}} \)};
	\node (exe0)  [sector, sector_k0, right=2.5 of plan0] {\( \Sigma^{(k)}_{\text{exe}} \)};

	\node (refl0) [sector, sector_k0_special, above=2 of plan0, xshift=0, yshift=-40] {\( \Sigma^{(k)}_{\text{refl}} \)};

	\draw[arrow] (perc0.east) -- node[edge_label, above] {\( g^k_{\text{perc} \rightarrow \text{plan}} \)} (plan0.west);
	\draw[arrow] (plan0.east) -- node[edge_label, above] {\( g^k_{\text{plan} \rightarrow \text{exe}} \)} (exe0.west);
	\draw[arrow] (perc0.north) to[out=120, in=150, looseness=1] node[edge_label, fill=none] {\( g^k_{\text{perc} \rightarrow \text{refl}} \)} (refl0.north west);
\draw[arrow] (refl0.south) to[out=270, in=90, looseness=1.2] node[edge_label, fill=none, left] {\( g^k_{\text{refl} \rightarrow \text{plan}} \)} (plan0.north);
	\draw[loop] (refl0.west) to[bend left=270, looseness=5] node[edge_label, left] {\( g^k_{\text{refl} \rightarrow \text{refl}} \)} (refl0.south west);

	\draw[arrow] (refl1.east) -- node[edge_label, above] {\( g^{k+1}_{\text{refl} \rightarrow \text{plan}} \)} (plan1.west);

	\draw[arrow, dashed] (refl0.north) -- node[operator_label, right, pos=0.4] {\( \Lambda \)} (refl1.south);

	\draw[arrow, dashed]  (plan1.south) to[out=0, in=0, looseness=1]  node[operator_label, left, pos=0.4] {\( V \)} (plan0.north east);
	
\end{tikzpicture}
\caption{Illustration of a recurrent coupling, specifically \(g^k_{\text{refl} \rightarrow \text{refl}}\), operating within \(\Sigma^{(k)}_{\text{refl}}\).}
	\label{fig:cognitive_pathway_loop_revised}
\end{figure}
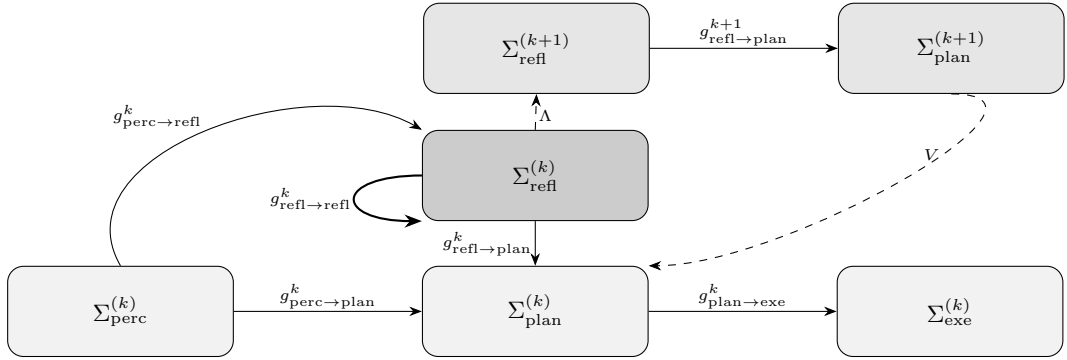

\subsubsection{Affective-Cognitive Couplings ($g_{\Sigma_X \leftrightarrow \Sigma_{\text{affect}}}^k$)}
\label{ssubsec:taxonomy_affective_cognitive}
These couplings describe the dynamics between cognitive sectors $\Sigma_X^{(k)}$ and an affective sector $\Sigma_{\text{affect}}^{(k)}$, integrating emotion-like states with cognitive processing for biasing attention, memory, and decision-making. For a full description, see Appendix~\ref{app:detail_affective_cognitive}.

\subsection{Section Summary: Formalizing Intra-Level Influence}
\label{subsec:summary_sec3}
This section introduced the core theoretical construct of this work: the formalization of interactions between cognitive subsystems that occur within a given abstraction level. The key contributions are:
\begin{enumerate}
	\item \textbf{Formal Definition of Intra-Level Couplings:} We defined the \textbf{intra-level sectoral coupling constant}, $g_{ij}^{k}$, as a parameter that quantifies the strength and nature of influence exerted by a source sector $\Sigma_i$ on a target sector $\Sigma_j$ exclusively within a single abstraction level $k$. The complete set of these constants, $G = \{g_{ij}^{k}\}$, forms the agent's unique \textbf{coupling profile}, a fundamental characteristic of its internal cognitive architecture.
	\item \textbf{A Taxonomy of Functional Roles:} To provide a nuanced understanding of their function, we presented a comprehensive taxonomy classifying the diverse roles these intra-level couplings play in the agent's cognitive economy. The classification covers key functional pathways, including Perceptual Integration, Memory Interaction, Planning and Goal Processing, Reflective \& Meta-Cognitive Regulation, and Execution Control.
\end{enumerate}
This formalization distinguishes direct, intra-level influences (parameterized by $g_{ij}^{k}$) from inter-level transitions (mediated by operators like $\Lambda$ and $V$), thereby providing a structured basis for analyzing an agent's internal information flow and processing tendencies.

\section{Cognitive Signatures}
\label{sec:cognitive_signatures}

Having defined the architecture of sectoral influence through intra-level coupling constants $g_{ij}^k$ in Section~\ref{sec:architecture_of_influence}, we now turn to the macroscopic consequences of these interactions. The complete collection of these intra-level couplings, $G = \{g_{ij}^k\}$, represents the agent's overall cognitive characteristics. An agent's typical processing tendencies, and the complex, emergent patterns of its internal belief evolution, are manifestations of this $G$ profile and its interplay with inter-level transitions mediated by operators like Abstraction ($\Lambda$) and Elaboration ($V$). This section explores:

\begin{enumerate}
	\item how distinct $G$ profiles allow for the characterization of recognizable "cognitive styles" (Subsection~\ref{subsec:cognitive_styles});
	\item how they are associated with sophisticated emergent dynamics (Subsection~\ref{subsec:emergent_dynamics});
	\item and how the very pathways through which information and influence flow within the agent's cognitive system (Subsection~\ref{subsec:conceptualizing_pathways}).
\end{enumerate}

\subsection{Cognitive Styles as Reflections of $G$ Profiles}
\label{subsec:cognitive_styles}

The specific pattern and relative strengths of the various intra-level coupling types within an agent's profile $G$ correspond to qualitatively different and persistent agent characteristics, which we term \textbf{``cognitive styles.''}

\begin{figure}[htbp]
	\centering
	\begin{subfigure}[t]{0.48\textwidth}
		\centering
		\begin{tikzpicture}[scale=0.85, transform shape,
			sector/.style={rectangle, draw, fill=gray!15, minimum size=1cm, rounded corners, font=\tiny},
			path/.style={->, thick},
			weak_path/.style={->, dashed, gray}]
			\node[sector] (perc0) at (0,0) {$\Sigma_{\text{perc}}^{(0)}$};
			\node[sector] (refl0) at (3,0) {$\Sigma_{\text{refl}}^{(0)}$}; 
			\node[sector] (plan0) at (6,0) {$\Sigma_{\text{plan}}^{(0)}$}; 
			\node[sector] (refl1) at (3,2) {$\Sigma_{\text{refl}}^{(1)}$}; 
			\node[sector] (plan1) at (6,2) {$\Sigma_{\text{plan}}^{(1)}$}; 
			\draw[path, black] (perc0) -- (refl0) node[midway, above, sloped, font=\tiny] {$g^0_{\text{perc}\rightarrow\text{refl}}$}; 
			\draw[path, black] (refl0) -- node[midway, right, font=\tiny] {$\Lambda$} (refl1); 
			\draw[path, black] (refl1) -- (plan1) node[midway, above, font=\tiny] {$g^1_{\text{refl}\rightarrow\text{plan}}$}; 
			\draw[path, black] (plan1) -- node[midway, left, font=\tiny] {$V$} (plan0);
			\draw[weak_path] (perc0.east) to[bend right=50] (plan0.west);
		\end{tikzpicture}
		\caption*{Reflective/Deliberative Agent}
		\label{fig:reflective}
	\end{subfigure}
	\hfill
	\begin{subfigure}[t]{0.48\textwidth}
		\centering
		\begin{tikzpicture}[scale=0.85, transform shape,
			sector/.style={rectangle, draw, fill=gray!15, minimum size=1cm, rounded corners, font=\tiny}, 
			path/.style={->, thick},
			weak_path/.style={->, dashed, gray}]
			\node[sector] (perc0) at (0,0) {$\Sigma_{\text{perc}}^{(0)}$};
			\node[sector] (refl0) at (2,0) {$\Sigma_{\text{refl}}^{(0)}$};   
			\node[sector] (plan0) at (4,0) {$\Sigma_{\text{plan}}^{(0)}$};  
			\node[sector] (exe0) at (7.5,0) {$\Sigma_{\text{exe}}^{(0)}$};     
			\node[sector] (refl1) at (2,2) {$\Sigma_{\text{refl}}^{(1)}$};   
			\draw[path, black] (perc0) to[bend right=40] node[midway, below, sloped, font=\tiny] {strong $g^0_{\text{perc}\rightarrow\text{plan}}$} (plan0);
			\draw[path, black] (plan0) -- (exe0) node[midway, above, sloped, font=\tiny] {strong $g^0_{\text{plan}\rightarrow\text{exe}}$}; 
			\draw[weak_path] (perc0.east) -- (refl0.west);
			\draw[weak_path] (refl0.east) -- (plan0.west);
			\draw[weak_path] (refl0.north) -- (refl1.south) node[midway, right, font=\tiny] {$\Lambda$};
		\end{tikzpicture}
		\caption*{Reactive Agent}
		\label{fig:reactive}
	\end{subfigure}
	\caption{Comparison of cognitive styles. Reflective agents have strong inter-level reasoning and regulation. Reactive agents rely on fast intra-level couplings at $k=0$.}
	\label{fig:agent_styles_comparison}
\end{figure}
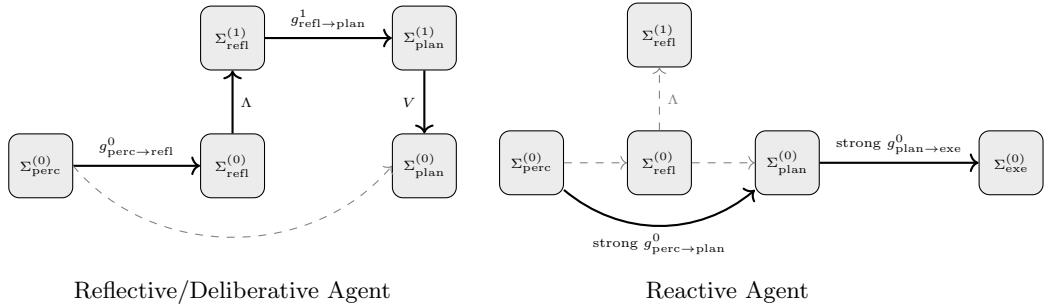

These styles are emergent properties that reflect the predominant modes of information processing inherent in the agent's architecture, as quantitatively described by its intra-level couplings and its characteristic engagement of inter-level operators. Building on the taxonomy from Section~\ref{subsec:taxonomy_couplings}, we illustrate with a few key examples how different $G$ profiles manifest as distinct cognitive styles:

\begin{itemize}
	\item \textbf{The "Reflective/Deliberative" Agent}: This style is primarily associated with a $G$ profile featuring dominant Reflective \& Meta-Cognitive Couplings (Section~\ref{ssubsec:taxonomy_reflective_metacognitive}) within various abstraction layers, coupled with effective use of $\Lambda$ and $V$ to bridge these layers. A profile with strong afferent intra-level couplings (e.g., $g_{\Sigma_{\text{plan}} \rightarrow \Sigma_{\text{refl}}}^k$) means that ongoing plans or surprising perceptual inputs within a level $k$ frequently result in introspective evaluation in $\Sigma_{\text{refl}}^{(k)}$. This might then lead to the engagement of $\Lambda$ to abstract these reflections to higher levels (e.g., $\Sigma_{\text{refl}}^{(k+1)}$) for broader consideration. Similarly, potent efferent intra-level couplings (e.g., $g_{\Sigma_{\text{refl}} \rightarrow \Sigma_{\text{plan}}}^k$) signify that meta-beliefs from $\Sigma_{\text{refl}}^{(k)}$ concerning coherence ($\kappa$), risk, or alignment with identity have a significant determining influence on planning and execution at that level $k$. If higher-level reflections from $\Sigma_{\text{refl}}^{(k+1)}$ are involved, they would be elaborated via $V$ back to level $k$ to exert influence, with the incorporation of this elaborated content potentially being influenced by other intra-level couplings at level $k$. Such an agent would likely exhibit careful, methodical decision-making, robust self-correction (potentially involving loops across abstraction levels mediated by $\Lambda$ and $V$), and a tendency to consult memory extensively (via strong Memory Retrieval Couplings $g_{\Sigma_X \leftrightarrow \Sigma_{\text{mem}}}^k$ at various $k$) before acting.
	
	\item \textbf{The "Reactive" Agent}: In contrast, a reactive style corresponds to a $G$ profile characterized by very strong, likely low-abstraction ($k=0$), Perceptual Integration Couplings (e.g., $g_{\Sigma_{\text{perc}} \rightarrow \Sigma_{\text{plan}}}^0$ from Section~\ref{ssubsec:taxonomy_perceptual_integration}). These define a direct and rapid influence of perception on planning within $k=0$. This is typically combined with efficient Execution \& Action Control Couplings ($g_{\Sigma_{\text{plan}} \rightarrow \Sigma_{\text{exe}}}^0$ from Section~\ref{ssubsec:taxonomy_execution_control}) also at $k=0$. In such a profile, pathways involving deep reflection (which might necessitate abstraction via $\Lambda$ to higher-$k$ reflective sectors) or extensive memory retrieval for deliberation (beyond immediate $k=0$ associations) might correspond to weaker coupling strengths, or the $\Lambda/V$ operators themselves might be less frequently engaged or less efficient. This results in rapid responses to immediate environmental stimuli with less emphasis on long-term consequence analysis or abstract reasoning (which are typically associated with higher-$k$ processing facilitated by $\Lambda$).
	
	\item \textbf{The "Dysfunctional" Agent (Example: Pathological Looping)}: Cognitive pathologies can correspond to specific imbalances in the $G$ profile. For instance, a $G$ profile with overly strong positive Intra-Sectoral Recurrent Couplings ($g_{ii}^k$ from Section~\ref{ssubsec:taxonomy_intra_sectoral}) within $\Sigma_{\text{refl}}^{(k)}$ or $\Sigma_{\text{plan}}^{(k)}$, without sufficient damping from inhibitory couplings or regulatory oversight from other sectors (i.e., weak $g_{X \leftrightarrow \text{refl}}^k$) (and potentially ineffective inter-level review), could result in unproductive rumination where meta-beliefs cycle within level $k$ without resolution. Such a profile could also lead to obsessive planning loops where the agent gets stuck refining without proceeding to action or appropriate inter-level transition (e.g., elaboration via $V$ to concrete steps or abstraction via $\Lambda$ for re-evaluation). Other dysfunctional styles, such as those stemming from an inability to learn due to very weak Memory Formation Couplings, are also conceivable.
\end{itemize}
Other distinct styles, such as
\begin{enumerate}
	\item "Creative/Associative,"
	\item "Focused/Goal-Driven,"
	\item or "Learning-Oriented/Adaptive" agents,
\end{enumerate}
can similarly be characterized by specific configurations of their $G$ profiles. These styles are not mutually exclusive; a sophisticated agent might exhibit a blend depending on the context.

\subsection{Emergent Cognitive Dynamics from Interacting Coupling Pathways}
\label{subsec:emergent_dynamics}

Beyond characterizing broad cognitive styles, the intricate interplay of multiple intra-level coupling types inherent in a specific $G$ profile, combined with the action of inter-level operators $\Lambda$ and $V$, corresponds to complex, emergent cognitive dynamics.

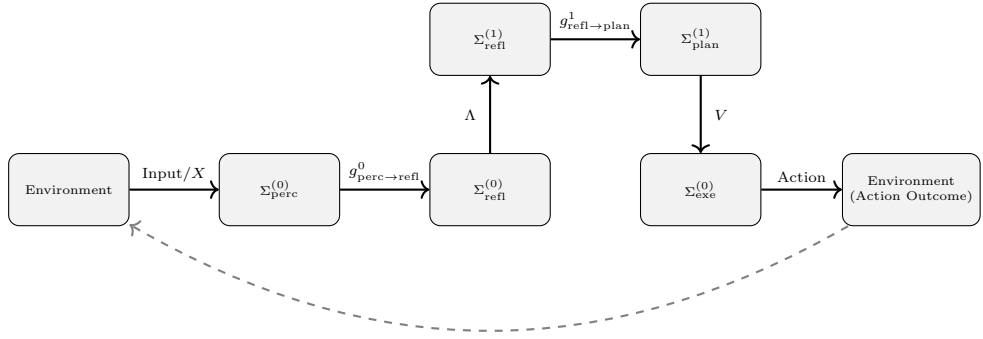
\begin{figure}[htbp]
	\centering
	\begin{tikzpicture}[
		scale=0.8, transform shape,
		sector/.style={rectangle, draw, fill=gray!10, minimum width=2cm, minimum height=1.2cm, rounded corners, font=\tiny, align=center},
		env/.style={rectangle, draw, fill=gray!10, minimum width=2cm, minimum height=1.2cm, rounded corners, font=\tiny, align=center},
		op_label/.style={font=\scriptsize, sloped, midway},
		path/.style={->, thick},
		inter_level_op/.style={font=\bfseries\scriptsize}
		]
		
		\node[env] (env_start) at (-3.5,0) {Environment}; 
		\node[sector] (perc0) at (0,0) {$\Sigma_{\text{perc}}^{(0)}$};
		\node[sector] (refl0) at (3.5,0) {$\Sigma_{\text{refl}}^{(0)}$};
		\node[sector] (refl1) at (3.5,2.5) {$\Sigma_{\text{refl}}^{(1)}$};
		\node[sector] (plan1) at (7,2.5) {$\Sigma_{\text{plan}}^{(1)}$};
		\node[sector] (exe0) at (7,0) {$\Sigma_{\text{exe}}^{(0)}$};
		\node[env] (env_end) at (10.5,0) {Environment \\ (Action Outcome)};
		
		\draw[path] (env_start) -- (perc0) node[op_label, above] {Input/$X$};
		\draw[path] (perc0) -- (refl0) node[op_label, above] {$g^0_{\text{perc}\rightarrow\text{refl}}$};
		\draw[path] (refl0) -- (refl1) node[inter_level_op, midway, left, xshift=-1mm] {$\Lambda$};
		\draw[path] (refl1) -- (plan1) node[op_label, above] {$g^1_{\text{refl}\rightarrow\text{plan}}$};
		\draw[path] (plan1) -- (exe0) node[inter_level_op, midway, right, xshift=1mm] {$V$};
		\draw[path] (exe0) -- (env_end) node[op_label, above] {Action};
		\draw[path, dashed, gray] (env_end) to[bend left=30] (env_start);
	\end{tikzpicture}
	\caption{Conceptual illustration of a multi-level cognitive processing cycle. Environmental input is processed by Observation Encoding ($X$) into $\Sigma_{\text{perc}}^{(0)}$. Intra-level couplings (e.g., $g^0_{\text{perc}\rightarrow\text{refl}}$) mediate influence within level $k=0$ to $\Sigma_{\text{refl}}^{(0)}$. The Abstraction operator ($\Lambda$) transfers information to a higher level (e.g., $\Sigma_{\text{refl}}^{(1)}$), where further intra-level processing occurs (e.g., via $g^1_{\text{refl}\rightarrow\text{plan}}$ to $\Sigma_{\text{plan}}^{(1)}$). The Elaboration operator ($V$) brings content back to a lower level (e.g., $\Sigma_{\text{exe}}^{(0)}$) for action execution.}
	\label{fig:multi_level_cycle}
\end{figure}

While individual $g_{ij}^k$ couplings parameterize direct influences within a level, their collective configuration as defined by the $G$ profile, along with inter-level transitions, constitutes the sophisticated pathways for information processing, regulation, and adaptation:

\begin{itemize}
	\item \textbf{Feedback Loops and Self-Regulation}: The taxonomy of intra-level couplings provides the structural basis for numerous positive and negative feedback loops crucial for stability and self-correction. These loops can operate entirely within an abstraction level $k$, or they can span multiple levels by incorporating $\Lambda$ and $V$ operations between intra-level segments. For instance, a reflective self-correction loop might involve sector $\Sigma_X^{(k)}$ signaling $\Sigma_{\text{refl}}^{(k)}$ (an interaction parameterized by $g_{X \rightarrow \text{refl}}^k$), which then might lead to abstraction of this signal to $\Sigma_{\text{refl}}^{(k+1)}$ (via $\Lambda$), its processing, elaboration back to $\Sigma_{\text{refl}}^{(k)}$ or $\Sigma_X^{(k)}$ (via $V$), with the subsequent influence on the originating sector being defined by the coupling $g_{\text{refl} \rightarrow X}^k$.
	
	\item \textbf{Cognitive Rhythms and Processing Cycles}: Robust sequences of communication between sectors, which are defined by chains of strong intra-level couplings and reliable $\Lambda/V$ operations, can establish characteristic cognitive rhythms or operational cycles. A typical problem-solving cycle might involve perception in $\Sigma_{\text{perc}}^{(0)}$ influencing reflection in $\Sigma_{\text{refl}}^{(0)}$ (an influence characterized by $g_{\text{perc} \rightarrow \text{refl}}^0$), then abstraction to $\Sigma_{\text{refl}}^{(1)}$ (via $\Lambda$), whose state influences planning in $\Sigma_{\text{plan}}^{(1)}$ (an influence characterized by $g_{\text{refl} \rightarrow \text{plan}}^1$), leading to elaboration to $\Sigma_{\text{exe}}^{(0)}$ (via $V$), the results of which feed back into $\Sigma_{\text{perc}}^{(0)}$, e.g., $$\Sigma_{\text{perc}}^{(0)} \xrightarrow{g^0_{\text{perc} \rightarrow \text{refl}}} \Sigma_{\text{refl}}^{(0)} \xrightarrow{\Lambda} \Sigma_{\text{refl}}^{(1)} \xrightarrow{g^1_{\text{refl} \rightarrow \text{plan}}} \Sigma_{\text{plan}}^{(1)} \xrightarrow{V} \Sigma_{\text{exe}}^{(0)} \rightarrow \text{Environment} \rightarrow \Sigma_{\text{perc}}^{(0)}.$$ The flow and balance within such cycles are direct consequences of the underlying $G$ profile (which defines the parameters for intra-level steps) and the efficiency and specific characteristics of $\Lambda/V$ operators (which define the nature of inter-level steps).
	
	\item \textbf{System Robustness and Pathways for Insight}: A well-balanced $G$ profile, corresponding to effective feedback and resource management parameters within levels, when combined with efficient inter-level integration via $\Lambda/V$ operators, is characteristic of cognitive robustness---the capacity to maintain coherent function despite perturbations. Furthermore, specific $G$ configurations, perhaps involving Abstractive/Elaborative processes (i.e., $\Lambda/V$ operators where the selection of inputs from $\Sigma_i^{(k_a)}$ or the assimilation of outputs into $\Sigma_j^{(k_b)}$ are significantly influenced by the strength of facilitatory intra-level couplings $g_{xi}^{k_a}$ or $g_{yj}^{k_b}$ respectively) that bridge disparate sectors across levels, or strong Memory Retrieval couplings with broad scope within levels, might define pathways more conducive to "insightful" connections or creative problem-solving by facilitating the novel recombination of belief fragments from various levels and sectors.
	
	\item \textbf{Dynamic Maintenance of Epistemic Identity}: The agent's sense of a stable self, or its epistemic identity $\vec{\eta}$ (as defined in \cite{Dumbrava2025TheoreticalFoundations}, Chapter 24), is dynamically maintained through the persistent and coherent operation of pathways. These pathways involve memory (e.g., autobiographical narratives whose formation is parameterized by intra-level couplings like $g^k_{\Sigma_X \rightarrow \Sigma_{\text{narr}} }$ in $\Sigma_{\text{narr}}^{(k)}$ and linked across levels by $\Lambda/V$ operators) and self-reflection (e.g., activity within $\Sigma_{\text{refl}}^{(k)}$ that occurs via pathways defined by various $g^k_{X \leftrightarrow \text{refl}}$ couplings, and that is abstracted/elaborated across reflective layers using $\Lambda/V$ operators). The stability of $\vec{\eta}$ is thus an emergent property reflecting the relevant subset of $G$ and the characteristics of the inter-level operators $\Lambda$ and $V$.
\end{itemize}

\subsection{Conceptualizing Interaction Pathways and Loops in $G$}
\label{subsec:conceptualizing_pathways}

Understanding the systemic impact of a given coupling profile $G$ involves more than analyzing individual $g_{ij}^k$ values in isolation; it requires conceptualizing how these intra-level couplings, by their defined strengths and configurations, collectively constitute a network of influence within each abstraction layer, and how these networks interface with inter-level operators to form the pathways for holistic cognitive processes.

\begin{figure}[htbp]
	\centering
	\begin{tikzpicture}[
		font=\small,
		sector_style/.style n args={1}{rectangle, draw, minimum width=13cm, minimum height=#1, rounded corners},
		belief_style/.style={rectangle, draw, text width=2.8cm, minimum height=1.2cm, align=center, font=\footnotesize},
		op_internal_style/.style={rectangle, draw, text width=3.5cm, minimum height=1cm, align=center, rounded corners},
		op_interlevel_style/.style={circle, draw, minimum size=0.8cm, font=\bfseries},
		coupling_label_style/.style={align=center},
		mod_arrow/.style={->, dashed, thin, shorten >=2pt, shorten <=2pt},
		data_arrow/.style={->, thick, >=Stealth, shorten >=1pt, shorten <=1pt},
		sector_label/.style={anchor=north west, font=\small, text=black!70, inner sep=2mm}
		]
		
		\node[sector_style={2.5cm}] (refl_sector) at (0,5) {};
		\node[anchor=north east, font=\bfseries\scriptsize, inner sep=3mm] at (refl_sector.north east) {Abstraction Level $k+1$};
		\node[sector_label] at (refl_sector.south west) {Reflective Sector ($\Sigma_{\text{refl}}^{(k+1)}$)};
		
		\node[sector_style={4.5cm}] (plan_sector) at (0,0) {};
		\node[anchor=north east, font=\bfseries\scriptsize, inner sep=3mm] at (plan_sector.north east) {Abstraction Level $k$};
		\node[sector_label] at (plan_sector.south west) {Planning Sector ($\Sigma_{\text{plan}}^{(k)}$)};
		
		\node[belief_style] (phi_crit_kplus1) at (0, 5) {Abstracted Critique \\ $\varphi_{\text{critique}}$};
		
		\node[belief_style] (phi_plan_in_k) at (-4, -0.75) {Original Plan \\ $\varphi_{\text{plan}}$};
		\node[op_internal_style] (Acorr_op) at (0, -0.75) {Corrective \\ Assimilation ($A_{\text{corr}}$)};
		\node[belief_style] (phi_plan_prime_k) at (4, -0.75) {Refined Plan \\ $\varphi'_{\text{plan}}$};
		\node[belief_style, dashed] (phi_crit_elab_k) at (0, 1.5) {Elaborated Critique \\ $\varphi_c'$};
		
		\node[op_interlevel_style] (A_op) at (-2.0, 3) {$\Lambda$};
		\node[op_interlevel_style] (V_op) at (2.5, 3) {$V$};
		
		\draw[data_arrow] (phi_plan_in_k) -- (A_op) node[midway, above, font=\tiny] {};
		\draw[data_arrow] (A_op) -- (phi_crit_kplus1);
		\draw[data_arrow] (phi_crit_kplus1) -- (V_op);
		\draw[data_arrow] (V_op) -- (phi_crit_elab_k);
		
		\draw[data_arrow] (phi_plan_in_k) -- (Acorr_op);
		\draw[data_arrow] (Acorr_op) -- (phi_plan_prime_k);
		
		\draw[mod_arrow] (phi_crit_kplus1) edge [loop, out=225, in=135, looseness=5] node[left, coupling_label_style] {$g_{\text{refl}\rightarrow\text{refl}}^{k+1}$} (phi_crit_kplus1);
		\draw[mod_arrow] (phi_crit_elab_k) -- (Acorr_op) node[midway, right, coupling_label_style] {$g_{\text{refl}\rightarrow\text{plan}}^{k}$};
		
	\end{tikzpicture}
	\caption{Reflective self-correction loop. An initial plan ($\varphi_{\text{plan}}$) in $\Sigma_{\text{plan}}^{(k)}$ is assessed, leading to its Abstraction ($\Lambda$) into $\Sigma_{\text{refl}}^{(k+1)}$. Within the reflective sector, an Abstracted Critique ($\varphi_{\text{critique}}$) is formed, a process modulated by the recurrent coupling $g_{\text{refl}\rightarrow\text{refl}}^{k+1}$. This critique is Elaborated ($V$) back to level $k$ as $\varphi_c'$. Finally, Corrective Assimilation ($A_{\text{corr}}$) combines the original plan and the critique to produce a Refined Plan ($\varphi'_{\text{plan}}$). This assimilation is modulated by the efferent coupling $g_{\text{refl}\rightarrow\text{plan}}^{k}$, which quantifies the influence of the reflective feedback on the planning process.}
	\label{fig:revised_reflective_loop_nocolor}
\end{figure}
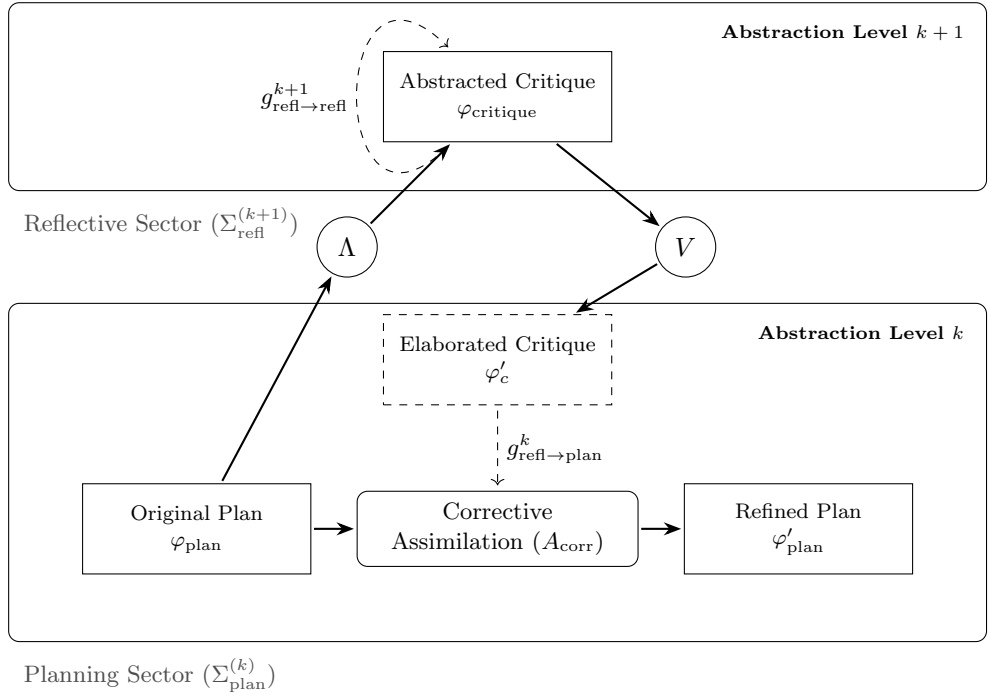

The set $G$ effectively represents a directed, weighted graph for each abstraction level $k$, where Semantic Sectors $\Sigma_s^{(k)}$ are nodes and the intra-level couplings $g_{ij}^k$ define the properties of the edges. Overall cognitive pathways may traverse these intra-level graphs and bridge between them using $\Lambda$ and $V$.

\begin{itemize}
	\item \textbf{Tracing Pathways of Influence}: One can conceptually trace sequences of strong intra-level couplings to understand how information or control signals are likely to propagate within a given level $k$. For example, a sequence of strong couplings $$g_{\Sigma_A \rightarrow \Sigma_B}^k \rightarrow g_{\Sigma_B \rightarrow \Sigma_C}^k \rightarrow g_{\Sigma_C \rightarrow \Sigma_D}^k$$ defines a dominant processing pipeline at that level $k$. Multi-level pathways involve sequences of such intra-level segments linked by $\Lambda$ or $V$ operations. Identifying such pathways helps in understanding how an agent might arrive at a particular belief state or decision.
	
	\item \textbf{Identifying Functional Loops}: Specific configurations of intra-level couplings define the structure of functional loops within an abstraction level. More complex loops spanning multiple levels involve these intra-level segments along with $\Lambda$ and $V$ operators. A cognitive loop, such as the reflective self-correction loop involving $\Sigma_{\text{plan}}^{(k)}$ and $\Sigma_{\text{refl}}^{(k)}$, can be entirely intra-level: activity in $\Sigma_{\text{plan}}^{(k)}$ influences $\Sigma_{\text{refl}}^{(k)}$ (an interaction parameterized by $g_{\Sigma_{\text{plan}} \rightarrow \Sigma_{\text{refl}}}^k$), and activity in $\Sigma_{\text{refl}}^{(k)}$ subsequently influences $\Sigma_{\text{plan}}^{(k)}$ (parameterized by $g_{\Sigma_{\text{refl}} \rightarrow \Sigma_{\text{plan}}}^k$). Alternatively, such a loop might involve $\Lambda$ to $\Sigma_{\text{refl}}^{(k+1)}$ and $V$ back down, where the characteristics of these inter-level steps are themselves influenced by the relevant intra-level couplings at the source and target levels. The overall behavior of such a loop (e.g., its stability, tendency towards convergence or oscillation, speed of correction) is determined by the properties (strength, sign, nature) of all constituent intra-level couplings in the path and the characteristics of any involved $\Lambda/V$ operators.
	
	\item \textbf{Pathological Loops and Cognitive Stagnation}: This pathway-centric view is particularly useful for conceptualizing cognitive pathologies. If, for instance, recurrent couplings within a sector (e.g., $g_{\Sigma_{\text{refl}} \rightarrow \Sigma_{\text{refl}}}^k$) are excessively strong and excitatory at a given level $k$, without adequate inhibitory influence or damping from other regulatory couplings at that level, this configuration can result in pathological reverberations or "rumination." In such cases, the agent may get stuck in a non-progressive cycle of thoughts within that level, or fail to appropriately engage $\Lambda/V$ for broader context or grounding. Similarly, a feedback loop that provides consistently flawed corrective signals due to miscalibrated intra-level couplings could lead to persistent maladaptive behavior. If such a loop is part of a larger, inter-level cognitive circuit, it could destabilize broader cognitive functions.
	
	\item \textbf{Conceptual Tool for Analysis}: Thinking in terms of these pathways and loops serves as a powerful conceptual tool. It allows for qualitative reasoning about the potential emergent behaviors, stabilities, and failure modes of a cognitive architecture that is defined by a particular set of intra-level couplings $G$ and specific $\Lambda/V$ characteristics.
\end{itemize}
This conceptualization of $G$ in concert with inter-level operators as defining a network of pathways and loops underscores the framework's utility in moving towards a more mechanistic and systems-level understanding of agent cognition.

\subsection{Section Summary: From Couplings to Cognitive Signatures}
\label{subsec:summary_sec4}
This section explored the macroscopic consequences of the intra-level coupling profile $G$, demonstrating how the collective configuration of $g_{ij}^k$ constants gives rise to an agent's characteristic processing tendencies and emergent behaviors. The key insights are:
\begin{enumerate}
	\item \textbf{Cognitive Styles:} An agent's persistent processing tendencies, which we term "cognitive styles," are a direct reflection of its underlying coupling profile $G$. For instance, a "Reflective/Deliberative" agent is characterized by strong meta-cognitive couplings and effective use of inter-level operators , whereas a "Reactive" agent is dominated by strong, low-level perceptual-to-planning couplings ($g_{\Sigma_{\text{perc}} \rightarrow \Sigma_{\text{plan}}}^0$). Specific imbalances in $G$, such as overly strong recurrent couplings, can lead to dysfunctional styles like pathological rumination.
	\item \textbf{Emergent System Dynamics:} Complex cognitive dynamics, including self-regulatory feedback loops and characteristic processing rhythms, emerge from the intricate interplay between the intra-level pathways defined by $G$ and the inter-level transitions mediated by operators like Abstraction ($\Lambda$) and Elaboration ($V$). The stability of an agent's epistemic identity ($\vec{\eta}$) is also presented as an emergent property of these dynamic pathways.
	\item \textbf{Conceptualizing Pathways and Loops:} The framework provides a powerful analytical lens by conceptualizing the cognitive architecture as a network of influence. At each abstraction level $k$, the Semantic Sectors ($\Sigma_s^{(k)}$) act as nodes and the intra-level couplings ($g_{ij}^k$) define the weighted, directed edges. This allows for tracing pathways of influence and identifying the structure of both functional and pathological cognitive loops.
\end{enumerate}
Ultimately, this section establishes that the set of couplings $G$ is a primary determinant of an agent's observable cognitive signature, bridging low-level architectural parameters with high-level, emergent patterns of belief evolution and behavior.

\section{Measuring and Inferring the Coupling Profile}
\label{sec:measuring_g}

The theoretical framework of intra-level sectoral coupling constants $g_{ij}^k$ and the resulting agent's coupling profile $G = \{g_{ij}^k\}$ offers a powerful lens for analyzing agent cognition. To make this framework practically useful and scientifically verifiable, however, it is essential to establish methodologies for empirically determining the values of these coupling constants, which characterize the agent's internal processing architecture.
This section addresses this estimation challenge by outlining:
\begin{enumerate}
	\item general approaches to estimation (Subsection~\ref{subsec:methodological_approaches});
	\item a detailed, concrete procedural framework for estimation in data-rich environments (Subsection~\ref{subsec:procedural_framework});
	\item and the key challenges inherent in the measurement process (Subsection~\ref{subsec:challenges_advanced_methods}).
\end{enumerate}
While the estimation of these parameters is complex, the outlined procedures provide a pathway towards grounding the $g_{ij}^k$ values in observable or derivable data.

\subsection{Methodological Approaches to Estimation}
\label{subsec:methodological_approaches}

Quantifying the set of intra-level couplings $G$ requires an analysis of data corresponding to those couplings. Several broad methodological avenues can be pursued:

\begin{figure}[htbp]
	\centering
	\begin{tikzpicture}[
		scale=1.0, 
		axis/.style={->, >=stealth, thick},
		event_i/.style={circle, draw=black!60!black, fill=black!30, minimum size=5pt, inner sep=0pt},
		event_j/.style={rectangle, draw=black!60!black, fill=black!30, minimum size=5pt, inner sep=0pt, rounded corners=1pt},
		track_label/.style={font=\small, anchor=east, xshift=-2mm},
		time_tick_label/.style={font=\tiny, below=2pt}
		]
		
		\draw[axis] (0,0) -- (9,0) node[right, font=\small] {Time ($t$)};
		\foreach \xval/\xlab in {1/t_1, 2.5/t_2, 4/t_3, 5.5/t_4, 7/t_5} { 
			\draw (\xval cm, -2pt) -- (\xval cm, 2pt) node[time_tick_label] {$\xlab$};
		}

		\node[track_label] at (0,1.5) {$\Sigma_i^{(k)}$ Events:};
		\draw[dotted, gray] (0,1.5) -- (9,1.5); 
		\node[event_i, label={[font=\tiny, black!80!black]above left:Event $e_{i1}$}] (ei1) at (1,1.5) {};
		\node[event_i, label={[font=\tiny, black!80!black]above left:Event $e_{i2}$}] (ei2) at (4,1.5) {};
		\node[event_i, label={[font=\tiny, black!80!black]above left:Event $e_{i3}$}] (ei3) at (7,1.5) {};

		\node[track_label] at (0,0.5) {$\Sigma_j^{(k)}$ Events:};
		\draw[dotted, gray] (0,0.5) -- (9,0.5); 
		\node[event_j, label={[font=\tiny, black!80!black]below right:Event $e_{j1}$}] (ej1) at (2.5,0.5) {}; 
		\node[event_j, label={[font=\tiny, black!80!black]below right:Event $e_{j2}$}] (ej2) at (5.5,0.5) {};

		\draw[->, thick, black!60!black, dashed, bend left=15] (ei1.south) to node[midway, below right, font=\tiny, black, xshift=8mm,  yshift=0mm] {Influence for $g_{ij}^k$?} (ej1.north);
		\draw[->, thick, black!60!black, dashed, bend left=15] (ei2.south) to node[midway, below right, font=\tiny, black, xshift=8mm,  yshift=0mm] {Influence for $g_{ij}^k$?} (ej2.north);
		
	\end{tikzpicture}
	\caption{Inferring $g_{ij}^k$ from discrete event co-occurrence. Events in $\Sigma_i^{(k)}$ and $\Sigma_j^{(k)}$ (at the same level $k$) are logged over time. Dashed arrows indicate potential influences analyzed by statistical methods to estimate $g_{ij}^k$.}
	\label{fig:data_for_g_inference_event_based}
\end{figure}
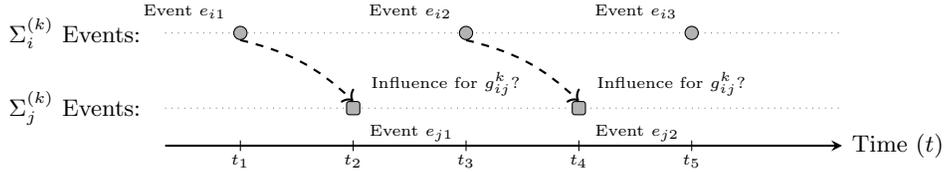

\begin{itemize}
	\item \textbf{Analysis of Internal Agent Logs (Primarily in Simulation Settings)}: When an agent's architecture allows for comprehensive logging of its internal states and processes, this provides the most direct data source. Detailed, time-stamped logs of belief fragments across different Semantic Sectors and Abstraction Layers, alongside records of cognitive operator activations (inputs, outputs, timings), can be analyzed. Statistical correlations and dependency modeling between activity in a source sector $\Sigma_i^{(k)}$ and subsequent, specific changes (e.g., operator applications, state modifications) in a target sector $\Sigma_j^{(k)}$ at the same level $k$ can be used to estimate the relevant $g_{ij}^k$ values.
	\item \textbf{Inference from Observable Behavior and Task Performance}: In many scenarios, especially with physical robots or agents where internal logging is limited, $g_{ij}^k$ values must be inferred more indirectly. Patterns in an agent's observable actions, decision latencies, error rates, learning curves, and overall success on carefully designed tasks can reflect its underlying coupling profile $G$ and the characteristics of its inter-level operators $\Lambda$ and $V$. Computational cognitive models that incorporate $G$ and inter-level operators as potentially parameterized components can be fitted to this behavioral data using system identification or inverse reinforcement learning techniques.
	\item \textbf{Inference from Agent's Output Text}: For linguistically competent agents, their textual output (e.g., explanations, dialogue, reports) can provide rich data. Features such as topic coherence and shifts, the integration (or lack thereof) of content from different conceptual domains (hypothesized to originate from different sectors $\Sigma_X^{(k)}$ at specific levels $k$), the expression of meta-cognitive states (linking to $\Sigma_{\text{refl}}^{(k)}$ activity), discourse structure, and variations in the abstraction level of language used (reflecting engagement of $\Lambda/V$ operators), can all be analyzed to infer characteristics of the underlying $G$ profile and inter-level processing that generated the text.
\end{itemize}
These approaches often complement each other and form the basis for the more detailed procedural framework presented next.

\subsection{A Detailed Procedural Framework for Empirical Estimation of $g_{ij}^k$}
\label{subsec:procedural_framework}

Building upon the general methodologies, this subsection details a concrete, phased procedural framework for the empirical estimation of a specific intra-level sectoral coupling constant $g_{ij}^k$. This is particularly suited for environments where internal agent data can be logged. We will use the example of estimating $g_{\Sigma_{\text{perc}} \rightarrow \Sigma_{\text{plan}}}^0$ (the parameter quantifying the intra-level influence from concrete perception $\Sigma_{\text{perc}}^{(0)}$ to concrete planning $\Sigma_{\text{plan}}^{(0)}$) in an autonomous drone navigating an obstacle course to illustrate the steps.

\begin{figure}[htbp]
	\centering
	\begin{tikzpicture}[
		phase/.style={rectangle, draw, fill=gray!10, text width=3.5cm, minimum height=1cm, rounded corners, align=center, font=\small},
		stepnode/.style={rectangle, draw, text width=3.2cm, minimum height=1cm, align=center, font=\scriptsize, inner sep=2pt, anchor=north},
		arrow/.style={->, thick, >=stealth}
		]
		
		\node[phase] (p1) at (0,6.5) {Phase 1: Preparation \& Data Collection};
		\node[stepnode] (s1a) at (0,5.3) {Define Target Intra-Level $g_{ij}^k$};
		\node[stepnode] (s1b) at (0,3.9) {Set Up Comprehensive Logging};
		\node[stepnode] (s1c) at (0,2.5) {Design Experimental Conditions/Tasks};
		\node[stepnode] (s1d) at (0,1.1) {Collect Data Across Multiple Trials};
		
		\draw[arrow] (p1.south) -- (s1a.north);
		\draw[arrow] (s1a.south) -- (s1b.north);
		\draw[arrow] (s1b.south) -- (s1c.north);
		\draw[arrow] (s1c.south) -- (s1d.north);
		
		\node[phase] (p2) at (5,5.5) {Phase 2: Data Pro- cessing \& Quan- titative Analysis}; 
		\node[stepnode] (s2a) at (5,4.0) {Identify/Isolate Rel- evant Intra-Level Interaction Events}; 
		\node[stepnode] (s2b) at (5,2.5) {Quantify Input-Output \& Model the Intra- Level Coupling $g_{ij}^k$}; 
		\node[stepnode] (s2c) at (5,1.0) {Statistical Es- timation of $g_{ij}^k$}; 
		
		\draw[arrow] (p2.south) -- (s2a.north);
		\draw[arrow] (s2a.south) -- (s2b.north);
		\draw[arrow] (s2b.south) -- (s2c.north);
		
		\draw[arrow, dashed] (s1d.east) .. controls (2.5,1.6) and (2.5,5.5) .. (p2.west);
		
		\node[phase] (p3) at (10,4.5) {Phase 3: Validation \& Interpretation};
		\node[stepnode] (s3a) at (10,3.0) {Validate Estimated $g_{ij}^k$ (e.g., predictive power, significance)};
		\node[stepnode] (s3b) at (10,1.5) {Interpret $g_{ij}^k$ in Con- text of Agent Archi- tecture \& Behavior}; 
		
		\draw[arrow] (p3.south) -- (s3a.north);
		\draw[arrow] (s3a.south) -- (s3b.north);
		
		\draw[arrow, dashed] (s2c.east) .. controls (7.5,1.0) and (7.5,4.5) .. (p3.west);
	\end{tikzpicture}
	\caption{Procedural framework for the empirical estimation of an intra-level sectoral coupling constant $g_{ij}^k$. The process involves three main phases: preparation and data collection; data processing and quantitative analysis to model and estimate the intra-level coupling; and finally, validation and interpretation of the estimated constant.}
	\label{fig:estimation_framework_corrected}
\end{figure}
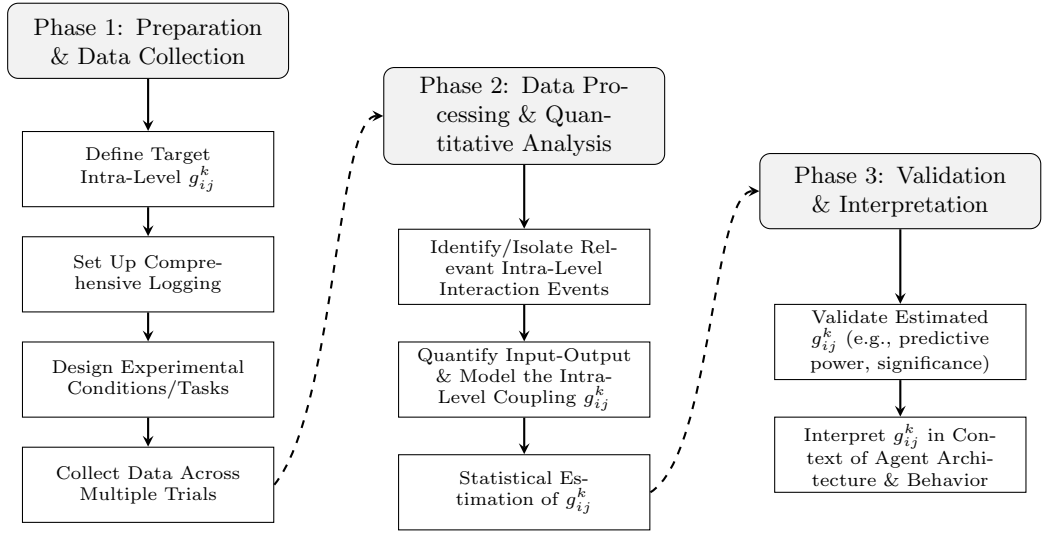

\subsubsection{Phase 1: Preparation and Data Collection}
\label{subsubsec:phase1_prep_data}
Thorough preparation is key for accurate estimation.
\begin{enumerate}
	\item \textbf{Define the Target Coupling(s)}: Clearly specify the source sector ($\Sigma_i^{(k)}$), target sector ($\Sigma_j^{(k)}$), and the abstraction level ($k$) of the intra-level interaction to be quantified. For our example, $\Sigma_i = \Sigma_{\text{perc}}$, $\Sigma_j = \Sigma_{\text{plan}}$, and $k=0$. We aim to quantify how direct, concrete perceptions in $\Sigma_{\text{perc}}^{(0)}$ correspond to immediate, concrete planning responses in $\Sigma_{\text{plan}}^{(0)}$.
	\item \textbf{Establish Comprehensive Data Logging}: Implement detailed, time-stamped logging. This must include:
	\begin{itemize}
		\item Belief fragments in $\Sigma_{\text{perc}}^{(0)}$, e.g., $$\varphi_{\text{obstacle}} = \text{"Obstacle type A at relative coords (x,y,z)"}$$ and $\Sigma_{\text{plan}}^{(0)}$, e.g., $$\varphi_{\text{evade}} = \text{"Execute evasive maneuver type M."}$$
		\item Activations of relevant operators, particularly Observation Encoding ($X$) generating $\varphi_{\text{obstacle}}$ into $\Sigma_{\text{perc}}^{(0)}$ and Assimilation ($A$) or planning operators within $\Sigma_{\text{plan}}^{(0)}$ generating $\varphi_{\text{evade}}$.
		\item Contextual data: drone speed, altitude, current broader task goal (if any).
	\end{itemize}
	\item \textbf{Design Experimental Conditions/Tasks}: For $g_{\Sigma_{\text{perc}} \rightarrow \Sigma_{\text{plan}}}^0$, the drone will navigate a simulated course with varied obstacles appearing at different positions and requiring different types of evasive maneuvers. The frequency, type, and urgency of obstacles will be controlled variables.
	\item \textbf{Collect Data Across Multiple Trials}: Run numerous trials with varying obstacle presentations to gather a rich dataset of perception-to-plan events occurring at $k=0$.
\end{enumerate}

\subsubsection{Phase 2: Data Processing and Quantitative Analysis}
\label{subsubsec:phase2_analysis}
This phase extracts interaction events and quantifies the relationship parameterized by $g_{ij}^k$.
\begin{enumerate}
	\setcounter{enumi}{4}
	\item \textbf{Identify and Isolate Relevant Interaction Events}: From logs, extract paired instances of (a) a specific perceptual belief $\varphi_{\text{obstacle}}^{(0)}$ entering $\Sigma_{\text{perc}}^{(0)}$, and (b) the subsequent generation of a relevant planning belief $\varphi_{\text{evade}}^{(0)}$ in $\Sigma_{\text{plan}}^{(0)}$ or an actual evasive action resulting from activity in $\Sigma_{\text{exe}}^{(0)}$ that was triggered by $\Sigma_{\text{plan}}^{(0)}$. Record latencies and the semantic content/appropriateness of the response.
	\item \textbf{Quantify the Input-Output Relationship and Model the Coupling}: This is where $g_{ij}^k$ (in our example $g_{\Sigma_{\text{perc}} \rightarrow \Sigma_{\text{plan}}}^0$) is determined as a parameter characterizing the intra-level influence. The coupling can be modeled in several ways depending on its function:
	\begin{enumerate}
		\item \textbf{Scenario A (Gated Response):} If the input from the source sector acts as a trigger for a response in the target sector, the model would quantify the strength and latency of this triggered sequence. For example, we can model the relationship between a perceptual event $\varphi_{\text{obstacle}}^{(0)}$ and the generation of a correct evasive plan $\varphi_{\text{evade}}^{(0)}$. The coupling $g_{\Sigma_{\text{perc}} \rightarrow \Sigma_{\text{plan}}}^0$ would be a key parameter in a function representing this relationship, quantifying how strongly perception gates or triggers the planning response.
		\item \textbf{Scenario B (Modulated Outcome):} If the operator in the target sector is always triggered but its outcome is affected by the source sector, the model would capture this modulation. For instance, if the drone always plans a maneuver but the \textit{type} of maneuver (e.g., "swerve" vs. "climb") or its \textit{speed of generation} depends on the perceptual input, then $g_{\Sigma_{\text{perc}} \rightarrow \Sigma_{\text{plan}}}^0$ would be the parameter in a model describing this effect on the plan's characteristics.
		\item \textbf{Scenario C (Modulated Parameter):} If the source sector influences a continuous parameter within the target sector's processing, the model would reflect this. For example, the perception of a very close obstacle could influence an "urgency level" parameter within the planning process at level $k=0$. The coupling $g_{\Sigma_{\text{perc}} \rightarrow \Sigma_{\text{plan}}}^0$ would be estimated as the parameter that quantifies this influence on the urgency parameter.
	\end{enumerate}
	\item \textbf{Quantitative Estimation of $g_{ij}^k$}: Using the collected (perception $\varphi^{(0)}$, plan-response $\varphi^{(0)}$) data pairs and the chosen quantitative model from the previous step, estimate the value of $g_{\Sigma_{\text{perc}} \rightarrow \Sigma_{\text{plan}}}^0$. This can be done using standard parameter estimation or function fitting techniques appropriate for the model. The sign of the estimated parameter would indicate if perception at $k=0$ typically corresponds to an excitatory or inhibitory influence on immediate planning at $k=0$, and its magnitude would quantify the strength of this rapid intra-level influence.
\end{enumerate}

\subsubsection{Phase 3: Validation and Interpretation}
\label{subsubsec:phase3_validation}
Assess the reliability and meaning of the estimated parameter $g_{ij}^k$.
\begin{enumerate}
	\setcounter{enumi}{7}
	\item \textbf{Validate the Estimated Coupling Constant(s)}:
	\begin{itemize}
		\item Test the predictive power of the model (e.g., predicting likelihood of evasion based on $\Sigma_{\text{perc}}^{(0)} \rightarrow \Sigma_{\text{plan}}^{(0)}$ dynamics) using the estimated $g_{\Sigma_{\text{perc}} \rightarrow \Sigma_{\text{plan}}}^0$ on a hold-out dataset of drone runs.
		\item Assess statistical significance (e.g., p-value for the coefficient $g$) and confidence intervals for the estimate.
		\item If the drone's $G$ profile was partly engineered with a known target value for this intra-level coupling, compare the estimate to this ground truth.
	\end{itemize}
	\item \textbf{Interpret the Coupling in Context}: A high positive estimated $g_{\Sigma_{\text{perc}} \rightarrow \Sigma_{\text{plan}}}^0$ would confirm a "reactive" tendency in the drone's obstacle response at the concrete level ($k=0$). A lower value might suggest that other factors (e.g., competing influences within $\Sigma_{\text{perc}}^{(0)}$ or $\Sigma_{\text{plan}}^{(0)}$ from other sectors like $\Sigma_{\text{refl}}^{(0)}$, or a propensity to quickly engage inter-level processing via $\Lambda$ before committing to a plan at $k=0$) correspond to a larger role before concrete plans are formed. Document if the estimated coupling strength varies with context (e.g., is higher when the drone is in a "high-speed mode," thus affecting $k=0$ processing differently).
\end{enumerate}

\subsection{Key Challenges in Measurement and Advanced Methodologies}
\label{subsec:challenges_advanced_methods}

Successfully estimating the full set of intra-level couplings $G$ is a significant undertaking. The procedural framework above, while concrete, simplifies a complex reality. Key challenges include the high dimensionality of $G$ (many sector pairs $(i,j)$ at each level $k$, summed over relevant $k$), potential identifiability issues when multiple intra-level couplings co-vary, the partial observability of internal states in many AI systems, and the non-stationarity of couplings if they adapt over time. Accurately modeling the precise functional form that describes how an intra-level $g_{ij}^k$ parameterizes an influence on an operator (even one acting only within level $k$) can also be difficult. Addressing these deep challenges is a key area for future research and will likely require the development of more sophisticated techniques. Ultimately, the choice of methodology must be tailored to the agent's architecture, the specific intra-level coupling under investigation, and the nature of the available data.

\subsection{Section Summary: Grounding the Framework through Measurement}
\label{subsec:summary_sec5}
This section addressed the critical challenge of empirically grounding the theoretical framework by outlining methodologies for measuring or inferring the intra-level sectoral coupling constants $g_{ij}^k$ from agent data. The key points are:
\begin{enumerate}
	\item \textbf{Methodological Approaches:} Three primary avenues for estimation were presented, depending on data availability. These are:
	\begin{enumerate}
		\item the direct analysis of internal, time-stamped agent logs;
		\item indirect inference from observable patterns in behavior, such as decision latencies and error rates;
		\item and inference from the linguistic features and structure of an agent's textual output.
	\end{enumerate}
	\item \textbf{A Procedural Framework:} A detailed, three-phase procedural framework was provided for the systematic estimation of a specific coupling constant in data-rich environments. The process consists of: 
	\begin{enumerate}
		\item \textbf{Preparation and Data Collection}, involving the design of experimental tasks and comprehensive logging;
		\item \textbf{Data Processing and Quantitative Analysis}, where interaction events are isolated and a quantitative model is used to estimate the $g_{ij}^k$ value;
		\item \textbf{Validation and Interpretation}, where the estimate's reliability is assessed and its meaning is contextualized within the agent's overall behavior.
	\end{enumerate}
	\item \textbf{Key Challenges:} The section acknowledged that successfully estimating the full coupling profile $G$ is a significant undertaking. Major challenges include the high dimensionality of $G$, potential identifiability issues from co-varying parameters, the partial observability of internal states in many systems, and the possibility of non-stationary couplings that change over time. Addressing these deep challenges likely requires moving beyond basic statistical correlations to more sophisticated techniques.
\end{enumerate}
These methodologies provide a pathway to make the coupling profile $G$ a set of verifiable parameters, thus making the overall cognitive model scientifically testable and practically useful.

\section{Dynamics of the Coupling Profile}
\label{sec:dynamics_of_G}

The collection of intra-level sectoral coupling constants, $G = \{g_{ij}^k\}$, which constitutes an agent's coupling profile, should not be conceptualized as an entirely static set of parameters. While certain core aspects of $G$ might be deeply embedded in the agent's architecture, the values defining the effective strengths of certain intra-level couplings can change over time or across contexts.
This section explores key aspects of such dynamics, including:
\begin{enumerate}
	\item a model for how coupling profiles propagate across abstraction layers (Subsection~\ref{subsec:coupling_propagation});
	\item the nature and implications of perturbations ($G \rightarrow G'$) that can alter the coupling profile (Subsection~\ref{subsec:coupling_perturbations});
	\item and how to assess architectural consistency and interpret deviations from expected dynamic behavior (Subsection~\ref{subsec:architectural_consistency}).
\end{enumerate}
Understanding these dynamics is crucial for a complete model of an adaptable and multi-layered cognitive system.

\subsection{Hierarchical Coupling Propagation}
\label{subsec:coupling_propagation}

To understand how an agent's internal interaction patterns scale with abstraction, we hypothesize that for a coherent cognitive architecture, there exists a stable, learnable relationship between the coupling profiles of adjacent abstraction layers. We term this relationship \textbf{hierarchical coupling propagation}.

We posit that, for abstraction levels $k$ and $k'$, and belief states $\phi$ and $\phi'$, there exists a parameterized family of operators $F(k', k, \phi', \phi)$ such that the vector of coupling constants $\vec{G}^{(k')}(\phi')$ can be derived from the vector $\vec{G}^{(k)}(\phi)$ according to
\[ \vec{G}^{(k')}(\phi') = F(k', k, \phi', \phi) \vec{G}^{(k)}(\phi).\]
We assume that $F$ above can be learned using standard machine learning techniques. However, as a first-order model for low abstraction levels and locally stable belief states, we propose that this propagation can be approximated by linear transformations, \textbf{a simplifying assumption made for the sake of exploration} (the true relationship is likely non-linear, potentially exhibiting saturating effects where couplings cannot be amplified indefinitely, or phase-transition-like behaviors where small changes in a low-level profile can cause large-scale reorganization at higher levels). With respect to this assumption, the vector of coupling constants at one level, $\vec{G}^{(k)}$, can be mapped to the effective vector of couplings at the next level, $\vec{G}^{(k+1)}$, by a \textbf{propagation matrix} $\mathbf{M}_k$ associated to that level $k$:
\begin{equation*}
	\vec{G}^{(k+1)} = \mathbf{M}_k \vec{G}^{(k)}.
\end{equation*}
Note that we restrict attention to the case $\phi \approx \phi'$, thereby neglecting the more general possibility that $\mathbf{M}_k$ itself varies as a function $\mathbf{M}_k = \mathbf{M}_k(\phi', \phi)$.

\begin{figure}[htbp]
	\centering
	\begin{tikzpicture}[
		font=\small,
		vec_style/.style={rectangle, draw, minimum width=1.2cm, minimum height=2.5cm, fill=gray!10, align=center},
		mat_style/.style={rectangle, draw, minimum width=2.8cm, minimum height=2.5cm, fill=gray!30, align=center},
		op_style/.style={font=\Large}
		]

		\node[mat_style] (m_mat) at (0,0) {$\mathbf{M}_k$ \\ \tiny (Propagation Matrix)};
		
		\node[op_style] (times_op) at (2.0,0) {$\times$};

		\node[vec_style] (gk_vec) at (3.5,0) {$\vec{G}^{(k)}$ \\ \tiny ($g_{ij}^{k}$ entries)};
		
		\node[op_style] (equals_op) at (5.0,0) {$=$};

		\node[vec_style] (gkplus1_vec) at (6.5,0) {$\vec{G}^{(k+1)}$ \\ \tiny ($g_{ij}^{k+1}$ entries)};
		
	\end{tikzpicture}
	\caption{Illustration of the linear propagation rule for coupling profiles. The vector of intra-level coupling constants at abstraction level $k$, denoted $\vec{G}^{(k)}$, is transformed by a Propagation Matrix $\mathbf{M}_k$ to produce the effective coupling profile $\vec{G}^{(k+1)}$ at the next level of abstraction.}
	\label{fig:linear_propagation_rule}
\end{figure}

This formulation provides a direct, testable hypothesis about the agent's structure. For a conceptual exploration of these dynamics using a continuous model of abstraction, see Appendix~\ref{app:continuous_coupling_propagation}.

\subsubsection{Mathematical Formulation and Significance}

The coupling profile at a higher level $k'$, denoted $\vec{G}^{(k')}$, can be derived from the profile at a lower level $k$, $\vec{G}^{(k)}$, by applying the product of the propagation matrices $\mathbf{M}_i$ for all levels $i$ between $k$ and $k'$. The formula for this iterated propagation is:
\[ \vec{G}^{(k')} = \left( \prod_{i = k}^{k' - 1} \mathbf{M}_i \right) \vec{G}^{(k)}. \]
The product $\prod_{i = k}^{k' - 1} \mathbf{M}_i$ represents a single composite operator that describes how the entire pattern of intra-level interactions transforms as it moves up the cognitive hierarchy from one level to a more abstract level.

\subsubsection{Long-Term Behavior and Convergence}

This linear model reveals how specific cognitive interaction patterns evolve with abstraction. The long-term behavior is determined by the eigenvalues ($\lambda$) and eigenvectors of the propagation matrices.
\begin{itemize}
	\item \textbf{Amplified Modes ($|\lambda| > 1$):} Interaction patterns (eigenvectors) associated with large eigenvalues are amplified, becoming the dominant features of the agent's high-level, abstract cognition.
	\item \textbf{Damped Modes ($|\lambda| < 1$):} Patterns with small eigenvalues are attenuated, their influence remaining localized to lower, more concrete levels of processing.
	\item \textbf{Fixed Points ($|\lambda| = 1$):} An interaction profile that is an eigenvector with an eigenvalue of 1 represents a scale-invariant feature of the agent's architecture, a cognitive "backbone" preserved across abstraction levels.
\end{itemize}

We posit a hypothesis regarding the ultimate fate of these couplings: the \textbf{Coupling Convergence Hypothesis}. It states that as the abstraction level $k$ tends towards infinity, all intra-level sectoral coupling constants must approach zero: $$\lim_{k \rightarrow \infty} g_{ij}^{(k)} = 0.$$ This implies that the only globally stable fixed point is the zero profile, representing a complete decoupling of cognitive functions at the highest, most abstract levels. This constraint is not merely mathematical; it is grounded in the physical limitations of any realizable agent. For more details, see Appendix~\ref{app:continuous_coupling_propagation}

\begin{example*}[Autonomous Cybersecurity Agent]
	\normalfont
	Consider an advanced AI agent for cybersecurity with sectors for Perception ($\Sigma_{\text{perc}}$), Planning ($\Sigma_{\text{plan}}$), and Reflection ($\Sigma_{\text{refl}}$) at concrete ($k=0$), tactical ($k=1$), and strategic ($k=2$) abstraction levels. Instead of deriving a transformation from first principles, we will infer it from data and use it to make a prediction.

	\par\medskip\noindent
	\textbf{1. Measurement}: From internal logs, we have empirically estimated the intra-level coupling matrices for the first two levels:
	\par\smallskip\noindent
	\textit{Level $k=0$ (Reactive Profile)}:
	$$ G^{(0)} = \begin{pmatrix} g_{\text{pp}} & g_{\text{pl}} & g_{\text{pr}} \\ g_{\text{lp}} & g_{\text{ll}} & g_{\text{lr}} \\ g_{\text{rp}} & g_{\text{rl}} & g_{\text{rr}} \end{pmatrix} = \begin{pmatrix} 0.2 & \mathbf{0.9} & 0.1 \\ 0.1 & 0.3 & 0.1 \\ 0.0 & 0.2 & 0.4 \end{pmatrix}. $$
	The high $g_{\text{perc} \to \text{plan}}^{(0)} = 0.9$ confirms a strong "reflex arc."
	
	\par\medskip\noindent
	\textit{Level $k=1$ (Deliberative Profile)}:
	$$ G^{(1)} = \begin{pmatrix} 0.2 & 0.5 & \mathbf{0.8} \\ 0.2 & 0.4 & 0.3 \\ 0.1 & \mathbf{0.7} & 0.5 \end{pmatrix}. $$
	Here, perception strongly triggers reflection ($g_{\text{perc} \to \text{refl}}^{(1)} = 0.8$), which in turn guides planning ($g_{\text{refl} \to \text{plan}}^{(1)} = 0.7$).
	
	\par\medskip\noindent
	\textbf{2. Inferring the Propagation Rule}: We now infer the propagation rule $\mathbf{M}_1$ that we expect couplings at level $k = 1$ to obey. For this illustrative example, assume, for simplicity, that the inferred rule is determined to be:
	\begin{align*}
		g_{\text{perc} \to \text{plan}}^{(k+1)} &= 0.5 \cdot g_{\text{perc} \to \text{plan}}^{(k)} \\
		g_{\text{perc} \to \text{refl}}^{(k+1)} &= 1.1 \cdot g_{\text{perc} \to \text{refl}}^{(k)} \\
		g_{\text{refl} \to \text{plan}}^{(k+1)} &= 1.1 \cdot g_{\text{refl} \to \text{plan}}^{(k)} \\
		g_{\text{refl} \to \text{refl}}^{(k+1)} &= 1.2 \cdot g_{\text{refl} \to \text{refl}}^{(k)}
	\end{align*}
	(Other couplings are assumed to be invariant under propagation). This rule---our inferred $\mathbf{M}_1$---shows that direct reactive pathways are damped, while reflective pathways are amplified.
	
	\par\medskip\noindent
	\textbf{3. Predicting Dynamics at $k=2$}: We can now use our inferred propagation rule to predict the agent's strategic-level coupling profile, $G^{(2)}$, by applying it to the tactical-level coupling profile $G^{(1)}$:
	\begin{align*}
		g_{\text{perc} \to \text{plan}}^{(2)} &= 0.5 \cdot g_{\text{perc} \to \text{plan}}^{(1)} = 0.5 \cdot 0.5 = 0.25 \\
		g_{\text{perc} \to \text{refl}}^{(2)} &= 1.1 \cdot g_{\text{perc} \to \text{refl}}^{(1)} = 1.1 \cdot 0.8 = 0.88 \\
		g_{\text{refl} \to \text{plan}}^{(2)} &= 1.1 \cdot g_{\text{refl} \to \text{plan}}^{(1)} = 1.1 \cdot 0.7 = 0.77 \\
		g_{\text{refl} \to \text{refl}}^{(2)} &= 1.2 \cdot g_{\text{refl} \to \text{refl}}^{(1)} = 1.2 \cdot 0.5 = 0.60
	\end{align*}
	This yields the predicted strategic-level coupling profile:
	$$ G^{(2)}_{\text{predicted}} = \begin{pmatrix} 0.2 & 0.25 & \mathbf{0.88} \\ 0.2 & 0.4 & 0.3 \\ 0.1 & \mathbf{0.77} & \mathbf{0.60} \end{pmatrix}. $$
	
	\par\medskip\noindent
	\textbf{Interpretation}: By modeling the process as coupling propagation, we have moved from observation to making a specific, testable prediction about the agent's high-level strategic reasoning architecture. This allows engineers to anticipate the agent's emergent cognitive style at higher levels of abstraction and provides a clear framework for verifying architectural consistency.
\end{example*}

\subsection{Coupling Perturbations ($G \rightarrow G'$): Sources and Consequences}
\label{subsec:coupling_perturbations}

A perturbation $G \rightarrow G'$ represents a change in the agent's set of intra-level coupling parameters, signifying a potentially significant shift in its cognitive processing style or architecture.

\begin{figure}[htbp]
	\centering
	\begin{tikzpicture}[
		scale=0.8,
		profile_box/.style={rectangle, draw=black, fill=gray!10, minimum width=4.5cm, minimum height=3.5cm, rounded corners},
		sector_node/.style={circle, draw, fill=white, minimum size=8mm, font=\scriptsize},
		coupling_strong/.style={->, ultra thick, black},                           
		coupling_medium/.style={->, thick, black},                                
		coupling_weak/.style={->, thin, gray!60},                     
		perturb_arrow/.style={->, very thick, black, font=\bfseries, shorten >=2mm, shorten <=2mm} 
		]
		
		\node[profile_box] (profile_G) at (-5.5,0) {}; 
		\node at (-5.5,2.5) [font=\small] {Initial Profile $G^{(k)}$ (at level $k$)}; 
		\node[sector_node] (S1_G) at (-5.5 - 1.3, 0.9) {$\Sigma_1$}; 
		\node[sector_node] (S2_G) at (-5.5 + 1.3, 0.9) {$\Sigma_2$}; 
		\node[sector_node] (S3_G) at (-5.5, -0.4) {$\Sigma_3$}; 
		
		\draw[coupling_strong] (S1_G) to node[midway,above,sloped,font=\tiny]{$g_{12}^k$} (S2_G);
		\draw[coupling_medium] (S2_G) to[bend left=15] node[midway,right,font=\tiny]{$g_{23}^k$} (S3_G);
		\draw[coupling_weak] (S1_G) to[bend right=15] node[midway,left,font=\tiny]{$g_{13}^k$} (S3_G);
		\draw[coupling_medium] (S3_G) to[loop below, looseness=10] node[right,font=\tiny]{$g_{33}^k$} (S3_G);

		\draw[perturb_arrow] (-3.0,0) -- (3.0,0) node[midway, align=center, text width=2.8cm, yshift=0.2cm, fill=white, inner sep=1pt] {Perturbation \\ (e.g., Learning, Intervention, Context Shift)};

		\node[profile_box] (profile_G_prime) at (5.5,0) {}; 
		\node at (5.5,2.5) [font=\small] {Perturbed Profile $G'^{(k)}$ (at level $k$)}; 
		\node[sector_node] (S1_Gp) at (5.5 - 1.3, 0.9) {$\Sigma_1$}; 
		\node[sector_node] (S2_Gp) at (5.5 + 1.3, 0.9) {$\Sigma_2$}; 
		\node[sector_node] (S3_Gp) at (5.5, -0.4) {$\Sigma_3$}; 
		
		\draw[coupling_medium] (S1_Gp) to node[midway,above,sloped,font=\tiny]{$g'_{12}{}^k$} (S2_Gp); 
		\draw[coupling_strong] (S2_Gp) to[bend left=15] node[midway,right,font=\tiny]{$g'_{23}{}^k$} (S3_Gp); 
		\draw[coupling_weak] (S1_Gp) to[bend right=15] node[midway,left,font=\tiny]{$g'_{13}{}^k$} (S3_Gp); 
		\draw[coupling_strong] (S3_Gp) to[loop below, looseness=10] node[right,font=\tiny]{$g'_{33}{}^k$} (S3_Gp); 
	\end{tikzpicture}
	\caption{Conceptual illustration of a perturbation transforming an initial intra-level coupling profile $G^{(k)}$ into a new profile $G'^{(k)}$ at the same abstraction level $k$. Such perturbations can arise from various sources (e.g., learning, external interventions, significant contextual shifts) and result in altered strengths or patterns of intra-level sectoral interactions (represented by changes in $g_{ij}^k$ values, visualized here by arrow thickness).}
	\label{fig:coupling_perturbation}
\end{figure}
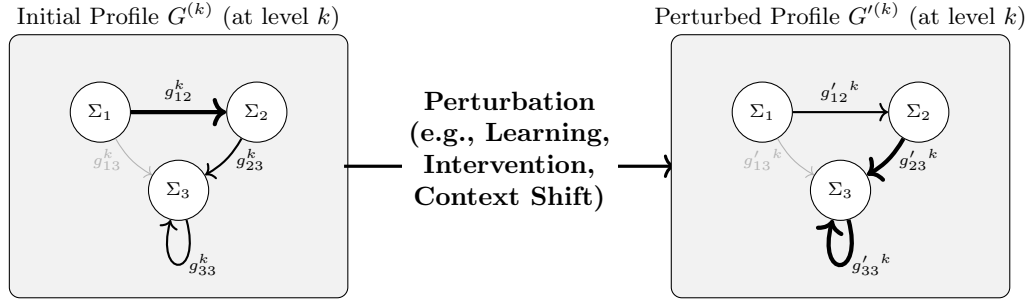

\begin{itemize}
	\item \textbf{Sources of Perturbation}: Changes in the $G$ profile can arise from several factors:
	\begin{itemize}
		\item \textit{Learning and Adaptation}: As an agent learns, its effective $G$ profile can evolve.
		\item \textit{External Interventions}: Deliberate re-tuning by designers or epistemic controls can induce changes in the $G$ profile.
		\item \textit{Contextual Shifts / State Dependency}: The values of intra-level couplings are functions of the belief state, $g_{ij}^k(\phi)$, so significant changes in $\phi$ can alter the effective $G^{(k)}$ profile.
		\item \textit{Internal Pathologies or System Drift}: Degradation or dysfunctional learned patterns could lead to detrimental perturbations in the $G$ profile.
	\end{itemize}
	
	\item \textbf{Consequences of Perturbations}: A shift from $G$ to $G'$ will fundamentally alter belief evolution dynamics. This change affects how information is processed within each level and can manifest as observable changes in cognitive style, task performance, and overall system stability.
\end{itemize}

\subsection{Architectural Consistency and Interpreting Deviations}
\label{subsec:architectural_consistency}

The dynamic nature of coupling parameters raises important considerations for architectural integrity and diagnostics.
\begin{itemize}
	\item \textbf{Rigidity in the Propagation Rule}: Attempting to rigidly enforce a deterministic propagation rule (a fixed $\mathbf{M}_k$) could lead to brittleness and inflexibility. A degree of plasticity in how coupling profiles manifest across scales might be more realistic and adaptive.
	\item \textbf{Inferring from Deviations}: If an agent exhibits behavior inconsistent with the profile predicted by its established propagation rule, e.g, $$G_{predicted}^{(k+1)} \neq \mathbf{M}_k G_{actual}^{(k)},$$ this signals a notable internal event. Potential influences include the onset of cognitive pathologies or the activation of overriding safety mechanisms. Monitoring for such deviations from the expected propagation dynamics is critical for understanding architectural stability and integrity.
\end{itemize}
These considerations highlight that the coupling profile $G$ and its propagation rules are not just descriptive tools; they provide a lens through which the consistency and health of the agent's cognitive architecture can be assessed.

\subsection{Section Summary: The Dynamic Nature of the Coupling Profile}
\label{subsec:summary_sec6}
This section explored the dynamics of the coupling profile $G$, conceptualizing it not as a static set of parameters but as a feature of the agent's architecture that can evolve over time and across different levels of abstraction. The key dynamic aspects discussed are:
\begin{enumerate}
	\item \textbf{Hierarchical Coupling Propagation:} A model for how coupling profiles transform across abstraction layers was proposed. This propagation might be approximated by a linear transformation, $\vec{G}^{(k+1)} = \mathbf{M}_k\vec{G}^{(k)}$, where a \textbf{Propagation Matrix} $\mathbf{M}_k$ maps the coupling profile from one level to the next. The eigenvalues of this matrix would determine whether specific interaction patterns are amplified (dominant modes) or attenuated (decaying modes) at higher levels of abstraction.
	\item \textbf{Coupling Perturbations ($G \rightarrow G'$):} The framework accounts for changes in the coupling profile, which are termed perturbations. Such shifts can arise from various sources, including learning and adaptation, external interventions by designers, significant contextual shifts, or internal system drift. These perturbations fundamentally alter an agent's cognitive processing style and belief evolution dynamics.
	\item \textbf{Architectural Consistency and Diagnostics:} The dynamic nature of the profile provides a basis for assessing architectural integrity and health. A key insight is that deviations from an established propagation rule (e.g., when $G_{predicted}^{(k+1)} \neq \mathbf{M}_k G_{actual}^{(k)}$) are significant internal events. Monitoring for such deviations can help diagnose the onset of cognitive pathologies or the activation of overriding safety mechanisms.
\end{enumerate}
Understanding these dynamics is crucial for modeling adaptable agents and for developing diagnostic tools to monitor the health and consistency of an agent's cognitive architecture.

\section{Discussion}
\label{sec:discussion}

The framework presented in this work, centered on intra-level sectoral coupling constants $G = \{g_{ij}^k\}$ for modeling belief dynamics within the Semantic Manifold framework, aims to provide a more quantitative, mechanistic, and dynamically nuanced understanding of agent cognition.
By focusing on the parameterized intra-level interactions between defined cognitive functions (Semantic Sectors at specific abstraction layers), and how these interact with inter-level operators ($\Lambda, V$), this approach seeks to bridge the gap between qualitative cognitive models and the need for precise, analyzable specifications of internal agent dynamics.
This section discusses several key aspects of the framework, including:
\begin{enumerate}
	\item the primary strengths and contributions of the intra-level coupling methodology (Subsection~\ref{subsec:discussion_strengths});
	\item its inherent challenges, limitations, and overall model complexity (Subsection~\ref{subsec:discussion_challenges});
	\item and its implications for practical epistemic control and AI governance (Subsection~\ref{subsec:discussion_epistemic_control}).
\end{enumerate}

\subsection{Strengths and Contributions of the Intra-Level Coupling Framework}
\label{subsec:discussion_strengths}

The synthesis of the linguistically grounded Semantic Manifold framework with a detailed parameterization of sectoral interactions via explicit intra-level coupling constants offers several conceptual and potential practical advantages: 
\begin{itemize}
	\item \textbf{Integration of Interpretability with Quantification}: The framework retains the interpretability of belief states ($\phi$) composed of linguistic fragments ($\varphi_i$) organized by Semantic Sectors ($\Sigma_s$) and Abstraction Layers ($k$), as detailed in Section~\ref{sec:background_sm}. Upon this, the intra-level coupling constants ($g_{ij}^k$), formally defined in Section~\ref{sec:architecture_of_influence}, provide a quantitative measure of intra-level interaction strength and nature, bridging qualitative understanding with formal modeling of processes within each abstraction layer. This moves beyond purely qualitative descriptions by offering a structured method to analyze functional dependencies governing belief evolution.
	\item \textbf{Richer Agent Characterization and Predictive Potential}: The set of intra-level couplings $G$ serves as a detailed profile of intra-level cognitive interactions, enabling a nuanced characterization of diverse "cognitive styles," as explored in Section~\ref{sec:cognitive_signatures}. These styles emerge from how $G$ shapes processing within levels and influences the engagement of inter-level operators ($\Lambda, V$), offering insights beyond what is possible by analyzing belief content or operator sets alone. This provides a richer basis for describing individual agent differences or developmental trajectories and lays groundwork for models with potentially improved predictive power regarding belief evolution and behavior.
	\item \textbf{Systematic Modeling of Complex Inter-functional Dependencies}: The explicit parameterization via $g_{ij}^k$ and the comprehensive taxonomy (Section~\ref{subsec:taxonomy_couplings}) allow for systematic investigation of how different cognitive functions influence each other within their respective abstraction layers. This includes modeling sophisticated dynamics like reflective self-regulation loops (which can be intra-level or involve inter-level $\Lambda/V$ steps, as conceptualized in Section~\ref{subsec:conceptualizing_pathways}) and hierarchical reasoning.
	\item \textbf{Foundation for Principled Dynamic Analysis and Testability}: By quantifying intra-level influence pathways, $G$ provides a robust foundation for analyzing belief dynamics. Furthermore, defining $g_{ij}^k$ as measurable parameters (Section~\ref{sec:measuring_g}) enhances the testability of hypotheses about cognitive architecture, encouraging research into fundamental parameters of an agent's cognitive economy. This offers a conceptual pathway for the principled design of AI agents with specific cognitive profiles.
	\item \textbf{Dynamic and Adaptive Potential}: The framework accommodates a dynamic $G$. Concepts explored in Section~\ref{sec:dynamics_of_G}, such as coupling propagation (Section~\ref{subsec:coupling_propagation}, describing how intra-level coupling profiles $G^{(k)}$ transform to $G^{(k+1)}$) and coupling perturbations ($G \rightarrow G'$) (Section~\ref{subsec:coupling_perturbations}), allow for hypothesis generation and testing, predictability and correction, and adaptation of an agent's cognitive architecture.
\end{itemize}

\subsection{Challenges, Limitations, and Model Complexity}
\label{subsec:discussion_challenges}

Despite its conceptual strengths, the proposed coupling-focused framework faces significant theoretical and practical challenges: 
\begin{itemize}
	\item \textbf{Overall Model Complexity and Parameter Explosion}: The model, with numerous operators and a potentially vast matrix of intra-level coupling constants $G = \{g_{ij}^k\}$ across many sectors and abstraction levels, is inherently complex. The sheer number of potential $g_{ij}^k$ parameters (the "parameter explosion" as one considers all $i,j,k$) makes comprehensive estimation and analysis daunting. Future work must explore strategies for managing this, such as identifying dominant or "relevant" intra-level couplings, assuming sparsity in the $G^{(k)}$ matrices for each level $k$, leveraging hierarchical parameterization, or focusing on effective couplings at specific scales of analysis.
	\item \textbf{Measurement Fidelity and Identifiability}: Reliably measuring or inferring intra-level $g_{ij}^k$ values from observable data (internal logs, behavior, or text, as discussed in Section~\ref{sec:measuring_g}) is a formidable inverse problem. Partial observability and interacting parameters (multiple $g_{ij}^k$ at the same level, or $g_{ij}^k$ influencing inputs to $\Lambda/V$ whose effects are observed later) can lead to significant identifiability issues.
	\item \textbf{Scalability}: The computational cost of simulating agents with extensive belief states and a dense set of intra-level coupling matrices $G^{(k)}$ for multiple $k$ poses serious scalability challenges for practical implementation.
	\item \textbf{Parameter Learning for $G$ and Operators}: Developing robust and data-efficient machine learning algorithms to learn $G$ (and potentially operator parameters, including for $\Lambda,V$) is a major research frontier. Defining appropriate learning signals is non-trivial. 
	\item \textbf{Empirical Validation}: Rigorously validating such a detailed internal model against external agent behavior or establishing its cognitive plausibility (both for intra-level $g_{ij}^k$ and inter-level $\Lambda/V$ processes) requires sophisticated experimental designs and metrics.
\end{itemize}

\subsection{Implications for Epistemic Control and AI Governance}
\label{subsec:discussion_epistemic_control}

Understanding an agent's intra-level coupling profile $G$ has direct and significant implications for designing and implementing effective epistemic control mechanisms, such as those explored in related work on Belief Filtering \cite{Dumbrava2025BeliefFiltering} and Belief Injection \cite{Dumbrava2025BeliefInjection}.
\begin{itemize}
	\item \textbf{Targeted Interventions}: Knowledge of $G$ can inform how and where to intervene in an agent's cognitive processes. If a problematic behavior is traced to an imbalanced coupling $g_{ij}^k$ (from the taxonomy in Section~\ref{subsec:taxonomy_couplings}), Belief Injection could introduce meta-beliefs into $\Sigma_{\text{refl}}^{(k)}$ to recalibrate regulatory policies acting at that level, or Belief Filtering could mitigate downstream consequences of activity within that level $\Sigma_X^{(k)}$. For example, identifying overly strong intra-level couplings leading to pathological loops within a level $k$ (as conceptualized in Section~\ref{subsec:conceptualizing_pathways}) could pinpoint specific targets for dampening interventions at that level.
	\item \textbf{Predicting Control Impact}: A model parameterized by $G$ (and parameters for $\Lambda/V$) could help predict the likely systemic effects of an injected belief or a new filter rule, allowing for more careful and principled application of these control mechanisms. Understanding how a specific intra-level $g_{ij}^k$ gates information or modulates an operator within its level $k$ allows for anticipating the ripple effects of altering inputs to that pathway, including effects on subsequent inter-level processing.
	\item \textbf{Designing for Governability}: Understanding how specific configurations of intra-level couplings $G^{(k)}$ influence information flow and stability within levels, and how they affect the engagement and output of inter-level regulatory operators (like $\Lambda/V$ in reflective loops), is critical for designing cognitive architectures that are inherently more stable, robust, and amenable to safe and aligned governance from the outset. Agents could potentially be engineered with $G$ profiles that favor desired cognitive styles (e.g., cautious deliberation via strong intra-level reflective couplings $g_{X \leftrightarrow \text{refl}}^k$ that effectively modulate planning at each level $k$ and promote inter-level review via $\Lambda/V$) in safety-critical applications.
\end{itemize}
The intra-level coupling framework thus provides a more granular, mechanistic understanding of the internal dynamics upon which epistemic controls must operate, moving beyond black-box approaches to agent regulation. Future, more sophisticated forms of epistemic control might even extend to "meta-architectural tuning," where an agent's underlying neural network parameters are systematically adjusted to achieve a pre-defined, desirable functional coupling profile $G$ (composed of intra-level $g_{ij}^k$), thereby shaping its cognitive style and behavioral predispositions in a highly targeted and interpretable manner. While the path to fully realizing and validating such a comprehensive framework (for both intra-level $g_{ij}^k$ and their interplay with inter-level $\Lambda/V$ processes) is marked by substantial research challenges, the potential to develop a deeper, more predictive, and quantitatively grounded science of artificial agent cognition makes this endeavor profoundly important.

\section{Future Directions}
\label{sec:future_directions}

The framework centered on intra-level sectoral coupling constants $G = \{g_{ij}^k\}$ for analyzing belief dynamics within the Semantic Manifold, as presented in this work, lays a conceptual and formal groundwork that opens numerous avenues for future research and development.
Realizing the full potential of this approach will require advancements in learning algorithms for $G$, rigorous empirical validation, deeper exploration of emergent cognitive phenomena linked to coupling profiles (and their interaction with inter-level processes mediated by $\Lambda$ and $V$), and translation into practical, high-impact applications.
This section outlines key priorities for such future work, including:
\begin{enumerate}
	\item developing advanced learning techniques for inferring and adapting the coupling profile $G$ (Subsection~\ref{subsec:future_learning_g});
	\item pursuing rigorous empirical validation and creating benchmark tasks (Subsection~\ref{subsec:future_empirical_validation});
	\item exploring the link between coupling profiles and emergent, higher-order cognitive phenomena (Subsection~\ref{subsec:future_emergent_phenomena});
	\item and translating the framework into applications for AI safety, diagnostics, and explainability (Subsection~\ref{subsec:future_applications}).
\end{enumerate}

\subsection{Advanced Learning Techniques for Inferring and Adapting $G$}
\label{subsec:future_learning_g}
A critical priority is the development of robust machine learning techniques to learn or infer the set of intra-level sectoral couplings $G = \{g_{ij}^k\}$ from agent data, and potentially to enable agents to adapt their own $G$ profiles over time.

\begin{itemize}
	\item \textbf{Inferring $G$ from Interaction Data}: Developing methods based on Hierarchical Reinforcement Learning (HRL), where high-level policies might implicitly define or modulate couplings $g_{ij}^k$ to optimize task performance by improving coordination between sectors within levels and across levels via $\Lambda/V$. Graph Neural Networks (GNNs) could also be explored to model sectoral interactions within each level $k$, where learned edge weights might correspond to coupling strengths $g_{ij}^k$. Multi-level GNN architectures could model the overall system including $\Lambda/V$ operations that link these intra-level graphs.
	\item \textbf{Bayesian Inference for $G$}: Employing sophisticated Bayesian methods (e.g., MCMC, Variational Inference) to infer posterior distributions over intra-level $g_{ij}^k$ parameters from sparse or noisy observational data (behavioral or textual), thereby explicitly modeling uncertainty in the inferred coupling profile $G$.
	\item \textbf{Learning Adaptive Couplings}: Investigating how $G$ might co-evolve with other core architectural parameters $\theta$ of the agent (as discussed in \cite{Dumbrava2025TheoreticalFoundations}, Chapter 37).
	\item \textbf{Learning Meta-Architectural Mappings for $G$}: A more speculative, yet potentially transformative, future direction involves elucidating the direct mapping between the low-level weights $W$ of an agent's foundational neural networks (those driving its cognitive operators, including $\Lambda$ and $V$) and the emergent, high-level intra-level sectoral coupling profile $G$. One could envision training a dedicated "G-Map" neural network capable of learning this complex, likely non-linear transformation $W \mapsto G$. 
	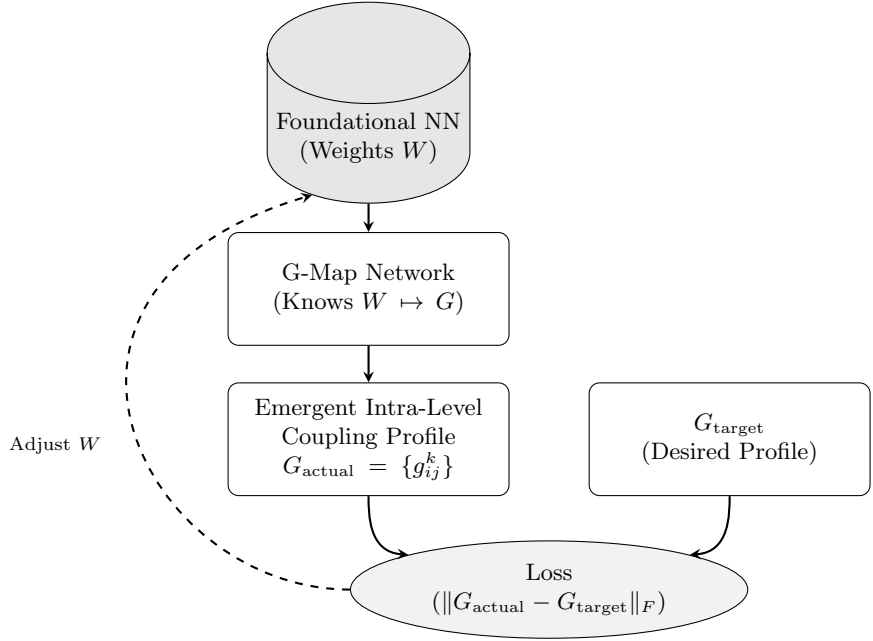
\begin{figure}[htbp]
		\centering
		\begin{tikzpicture}[
			box/.style={rectangle, draw, text width=3.5cm, minimum height=1.5cm, align=center, rounded corners, font=\small},
			arrow/.style={->, thick, >=stealth, font=\scriptsize},
			network/.style={cylinder, shape border rotate=90, draw, fill=gray!20, aspect=0.5, minimum height=1.5cm, minimum width=2cm, align=center, font=\small},
			loss_node_style/.style={ellipse, draw, fill=gray!10, text width=3.5cm, align=center, font=\small, minimum height=1cm}
			]
			\node[network] (nn_w) at (0,4.8) {Foundational NN \\ (Weights $W$)}; 
			\node[box] (g_mapper) at (0,2.8) {G-Map Network \\ (Knows $W \mapsto G$)}; 
			\node[box] (g_actual) at (0,0.8) {Emergent Intra-Level Coupling Profile $G_{\text{actual}} = \{g_{ij}^k\}$}; 
			\node[box] (g_target) at (4.8,0.8) {$G_{\text{target}}$ \\ (Desired Profile)}; 
			
			\node[loss_node_style] (loss) at (2.4,-1.2) {Loss \\ ($\|G_{\text{actual}} - G_{\text{target}}\|_F$)}; 
			
			\draw[arrow] (nn_w) -- (g_mapper);
			\draw[arrow] (g_mapper) -- (g_actual);
			\draw[arrow] (g_actual.south) .. controls (0,-0.2) and (0,-0.7) .. (loss.north west); 
			\draw[arrow] (g_target.south) .. controls (4.8,-0.2) and (4.8,-0.7) .. (loss.north east); 
			\draw[arrow, dashed] (loss.west) .. controls (-3,-1.2) and (-5,2.8) .. (nn_w.south west) node[midway, below left, font=\scriptsize, xshift=-3mm] {Adjust $W$};

		\end{tikzpicture}
		\caption{Conceptual diagram of the G-Map meta-tuning framework. A dedicated G-Map network learns the transformation from an agent's foundational neural network weights ($W$) to its emergent intra-level sectoral coupling profile ($G_{\text{actual}} = \{g_{ij}^k\}$). A desired target profile ($G_{\text{target}}$) can be defined. The discrepancy (Loss), for instance, the Frobenius norm of the difference $\|G_{\text{actual}} - G_{\text{target}}\|_F$, then guides adjustments to the underlying weights $W$, facilitating principled architectural modification via an interpretable intermediate representation $G$.}
		\label{fig:g_mapper_concept_corrected}
	\end{figure}

	Such a G-Map would essentially model how specific configurations of underlying network parameters give rise to the agent's quantifiable intra-level coupling profile. The successful development of a reliable G-Map could enable a novel paradigm of "meta-tuning." If a target coupling profile ($G_{\text{target}}$) ---representing a desired cognitive style or specific behavioral predisposition---can be defined, the G-Map could facilitate a more interpretable and principled approach to tuning the agent's architecture. While training the weights $W$, adjustments could be guided by minimizing a loss function based on the discrepancy between the current $G_{\text{actual}}$ (derived from $W$ via the G-Map) and $G_{\text{target}}$. This "meta-tuning" via an interpretable intermediate representation ($G$ of intra-level couplings) offers significant potential for: 
	\begin{itemize}
		\item more principled cognitive design by adjusting understandable functional interaction parameters ($g_{ij}^k$);
		\item targeted adaptation of cognitive styles (Section~\ref{subsec:cognitive_styles});
		\item and enhanced AI safety by promoting $G$ profiles known to correlate with safer operational modes.
	\end{itemize}
	 Key challenges in this endeavor include the data-intensive training of an accurate G-Map (requiring numerous ($W, G$) pairs, where $G$ itself needs robust estimation, potentially involving an iterative refinement process between $G$ estimation and G-Map training), managing the complexity of the $W \mapsto G$ mapping, and ensuring system stability during such meta-tuning processes. Appendix~\ref{app:g_mapper_strategies} outlines a set of potential experimental strategies designed to address these challenges and further develop the G-Map concept, including the use of operational gauge equivalence for dataset generation and systematic analysis of the learned $W \mapsto G$ mapping.
\end{itemize}

\subsection{Empirical Validation and Benchmark Development}
\label{subsec:future_empirical_validation}
The theoretical predictions and characterizations derived from different $G$ profiles necessitate rigorous empirical validation to ground the framework.
\begin{itemize}
	\item \textbf{Simulated Agent Testbeds}: Implementing agents within controlled simulated environments where their architecture and $G$ profile can be explicitly defined and manipulated. This will be crucial for: 
	\begin{itemize}
		\item testing whether agents with different engineered $G$ profiles exhibit the predicted "cognitive styles" (Section~\ref{subsec:cognitive_styles}) and performance characteristics (arising from both intra-level processing shaped by $G$ and inter-level processing via $\Lambda/V$);
		\item and validating methods for inferring $g_{ij}^k$ (Section~\ref{sec:measuring_g}) by comparing inferred values against ground-truth parameters, using detailed internal logs.
	\end{itemize}
	\item \textbf{Human-Agent Interaction (HAI) Studies}: Designing HAI tasks to investigate how an agent's $G$ profile (and its consequent inter-level processing patterns) influences the quality of collaboration, its interpretability to human users, or the perceived "personality" and reasoning style of the agent.
	\item \textbf{Development of Benchmark Tasks}: Creating specific benchmark tasks and environments explicitly designed to exercise and thus reveal the influence of particular intra-level sectoral couplings, as well as the agent's use of inter-level operators like $\Lambda$ and $V$. For example, tasks requiring rapid integration of perception and planning within a specific level $k$ would heavily depend on specific Perceptual Integration Couplings $g_{\Sigma_{\text{perc}} \rightarrow \Sigma_{\text{plan}}}^k$, while others might require deep reflection involving abstraction to higher-$k$ reflective sectors and subsequent intra-level processing at those higher levels, thus highlighting both intra-level Reflective \& Meta-Cognitive Couplings and the Abstraction mechanism.
\end{itemize}

\subsection{Exploring Emergent Phenomena and Links to Higher Cognition}
\label{subsec:future_emergent_phenomena}
The overall coupling profile $G$ fundamentally shapes the agent's internal cognitive ecosystem, including how it utilizes inter-level operators $\Lambda$ and $V$. Future theoretical and empirical work could explore how specific configurations and dynamics of $G$ relate to complex, emergent cognitive phenomena.
\begin{itemize}
	\item \textbf{Insight, Creativity, and Problem-Solving Strategies}: Investigating if particular patterns in $G$---such as balanced excitatory/inhibitory intra-level couplings, strong modulation of inputs/outputs for Abstractive \& Elaborative processes that facilitate novel connections between disparate sectors across levels, or specific dynamics in coupling propagation---are more conducive to cognitive breakthroughs or flexible problem-solving strategies.
	\item \textbf{Robustness and Dynamics of Epistemic Identity ($\vec{\eta}$)}: Analyzing how different $G$ profiles contribute to the formation, maintenance, and resilience of an agent's epistemic identity $\vec{\eta}$ over time, as introduced in \cite{Dumbrava2025TheoreticalFoundations} (Chapter 24).
	\item \textbf{Computational Analogs of Higher-Order Cognitive Functions}: While highly speculative, exploring whether very complex, highly integrated, and recurrent coupling patterns within $G$, especially those involving extensive feedback loops (potentially spanning multiple levels via $\Lambda/V$) with the reflective sector $\Sigma_{\text{refl}}$ (as conceptualized in Section~\ref{subsec:conceptualizing_pathways}), could support computational analogs of phenomena associated with higher-order awareness or integrated information processing.
\end{itemize}

\subsection{Applications: AI Safety, Diagnostics, Personalized AI, and XAI}
\label{subsec:future_applications}
A quantitative understanding of intra-level sectoral couplings and their role in modulating inter-level processes offers significant potential for practical AI applications, particularly in areas requiring robust, interpretable, and aligned AI.
\begin{itemize}
	\item \textbf{AI Safety and Alignment}: The $G$ profile could be a target for design or monitoring to promote safety. This includes engineering agents with $G$ profiles that inherently favor cautious, deliberative cognitive styles (e.g., via strong intra-level Reflective Couplings $g_{X \leftrightarrow \text{refl}}^k$ that gate action within level $k$ or promote engagement of $\Lambda$ for higher-level review) or ensuring that ethical principles, represented within specific sectors $\Sigma_{\text{ethic}}^{(k)}$, have appropriate intra-level coupling strengths $g_{\text{ethic} \rightarrow \text{plan}}^k$ to effectively influence planning and execution at relevant levels. Inferred $G$ profiles could also serve as a diagnostic tool to monitor agents for "value drift" or the emergence of unsafe internal interaction patterns (e.g., pathological intra-level loops or problematic engagement of $\Lambda/V$ operators).
	\item \textbf{Diagnostics of Cognitive Pathologies and Targeted Intervention}: If an agent exhibits undesirable behavior, analyzing its $G$ profile or recent perturbations therein (Section~\ref{subsec:coupling_perturbations}) could help diagnose underlying issues like dissociation between sectors within a level (weak $g_{ij}^k$), runaway intra-level feedback loops, or flawed modulation of critical $\Lambda/V$ processes. This could inform targeted "therapeutic" interventions, potentially via mechanisms like Belief Injection \cite{Dumbrava2025BeliefInjection} to introduce corrective meta-beliefs (e.g., in $\Sigma_{\text{refl}}^{(k)}$) or adjust specific intra-level interaction pathways or the triggers for inter-level review.
	\item \textbf{Personalized AI Systems}: Engineering AI companions, tutors, or assistants whose coupling profiles $G$ are tuned or can adapt to better match a specific user's cognitive style, communication preferences, or task requirements, leading to more effective and intuitive human-AI collaboration.
	\item \textbf{Explainable AI (XAI)}: Enhancing AI explainability by allowing systems to articulate not just what they believe, but how their internal cognitive architecture predisposes them to process information in certain ways. Explanations could refer to dominant intra-level coupling pathways influencing a decision within a particular abstraction layer (e.g., "The decision was rapid because immediate perceptual inputs strongly drive planning due to a high coupling between perception sectors and planning sectors"), or explain why certain information was or was not escalated to higher abstraction levels or elaborated to concrete details.
\end{itemize}
These future directions underscore the richness of the proposed intra-level coupling-focused framework and its interaction with inter-level processing. Continued research in these areas, particularly empirical validation and the development of methods for learning $G$ and understanding its interplay with $\Lambda/V$ operators, is essential for realizing its full potential for advancing the modeling, understanding, construction, and governance of sophisticated artificial intelligence systems.

\section{Conclusion}
\label{sec:conclusion}

This work has introduced and elaborated upon a novel theoretical framework centered on \textbf{intra-level sectoral coupling constants $g_{ij}^k$} for quantitatively analyzing and understanding the complex cognitive dynamics within artificial agents. Building upon the foundational principles of the Semantic Manifold framework \cite{Dumbrava2025TheoreticalFoundations}, which posits that agent belief states are structured ensembles of interpretable linguistic fragments organized by function and abstraction (as reviewed in Section~\ref{sec:background_sm}), our work provides a mechanistic approach to characterizing the intricate web of influences between an agent's diverse cognitive capacities, distinguishing between direct intra-level interactions and operator-mediated inter-level transitions. The central aim has been to move towards a more precise, analyzable, and ultimately predictive model of belief evolution in sophisticated AI, grounded in clearly defined scopes of interaction.

At the core of this contribution is the formalization of $g_{ij}^k$ as parameters quantifying the strength and nature of influence between Semantic Sectors ($\Sigma_s$) exclusively within specific Abstraction Layers ($k$) (as detailed in Section~\ref{sec:architecture_of_influence}). Inter-level influences are mediated by dedicated operators like Abstraction ($\Lambda$) and Elaboration ($V$), whose functioning can be modulated by these intra-level couplings. The complete set of these intra-level couplings, $G = \{g_{ij}^k\}$, constitutes the agent's unique \textbf{coupling profile,} a fundamental descriptor of its internal processing architecture at each level of abstraction. We have detailed how this coupling profile $G$ directly modulates the behavior of core Semantic Manifold cognitive operators by describing the direct influence within operational layers through the modulation of inputs, outputs, or operational efficacy. A comprehensive taxonomy (Section~\ref{subsec:taxonomy_couplings}) was presented to classify the diverse functional roles these intra-level couplings play, and to explain how they contribute to the modulation of inter-level transition processes.

The work has explored how specific configurations of $G$ can give rise to distinct "cognitive styles" and complex emergent system dynamics, including feedback loops and self-regulatory behaviors (Section~\ref{sec:cognitive_signatures}), which arise from the interplay of $G$-modulated intra-level processing and $\Lambda/V$-mediated inter-level operations. Furthermore, we have addressed the empirical grounding of this framework by outlining methodological approaches and a detailed procedural framework for measuring or inferring these intra-level $g_{ij}^k$ values (Section~\ref{sec:measuring_g}). The dynamic nature of the coupling profile $G$ itself, and its susceptibility to perturbations, was also examined (Section~\ref{sec:dynamics_of_G}).

The significance of this sectoral coupling-focused framework for advancing cognitive models, as discussed in Section~\ref{sec:discussion}, is manifold:
\begin{itemize}
\item It brings \textbf{enhanced precision and quantifiability} to the description of inter-functional dependencies within cognitive architectures, by clearly delineating direct intra-level influences from operator-mediated inter-level processes.
\item It offers the potential for \textbf{improved predictive power} regarding an agent's belief evolution and subsequent behavioral tendencies, once $G$ and the characteristics of its inter-level operators are known or well-estimated.
\item It increases the \textbf{testability} of hypotheses about cognitive architecture by defining measurable intra-level parameters.
\item It provides a basis for a \textbf{richer characterization of agent differences} and developmental trajectories through their unique intra-level coupling profiles and patterns of inter-level operator engagement.
\end{itemize}
This approach encourages a research program focused on identifying and understanding the fundamental parameters that orchestrate an agent's internal cognitive economy, both within and between abstraction layers.

The ultimate promise of formalizing belief dynamics through intra-level sectoral couplings interacting with inter-level transition mechanisms lies in its potential to transform how we understand, design, and ensure the safe and aligned operation of complex AI systems:
\begin{itemize}
\item It offers pathways for the \textbf{principled design and engineering} of agents with specific, desired cognitive interaction profiles (defined by $G$) and styles, potentially tailored for particular tasks or ethical considerations by shaping both intra-level processing and the invocation of inter-level transitions.
\item It enables a \textbf{deeper understanding of internal cognitive dynamics}, allowing for the tracing of actions and decisions back to the evolution of linguistic beliefs as shaped by the specific strengths of intra-level sectoral influence and the operations of inter-level transformers.
\item It contributes to \textbf{enhanced AI safety, alignment, and diagnostics} by providing a more granular model of the internal mechanisms upon which regulatory feedback loops depend, clarifying how intra-level $g_{ij}^k$ and inter-level operators $\Lambda/V$ contribute to these loops. An understanding of an agent's $G$ profile can aid in diagnosing "cognitive pathologies" (e.g., detrimental intra-level feedback loops, dissociation between sectors within a level, or flawed modulation of $\Lambda/V$ processes) and can inform the targeted application of epistemic control mechanisms such as Belief Filtering \cite{Dumbrava2025BeliefFiltering} or Belief Injection \cite{Dumbrava2025BeliefInjection}.
\end{itemize}
While significant challenges remain, particularly in the robust measurement of these intra-level couplings on a large scale, the development of efficient learning algorithms for $G$ (and for characterizing $\Lambda/V$ operations), and comprehensive empirical validation (as outlined in Section~\ref{sec:future_directions}), the synthesis of the Semantic Manifold framework's structured linguistic approach with a quantitative, dynamic model of intra-level sectoral couplings interacting with inter-level processes represents a vital and promising direction. It propels us towards building artificial agents whose internal cognitive ecosystems are not only sophisticated and capable but also increasingly analyzable, predictable, and ultimately, more deeply understandable and governable. This work endeavors to lay part of the groundwork for such future advancements in the science of artificial cognition.

\appendix
\section{Detailed Taxonomy of Semantic Sectors}
\label{app:sectors_abstraction_granularity}

The main body of this work, and the foundational Semantic Manifold framework \cite{Dumbrava2025TheoreticalFoundations}, primarily utilizes a core set of Semantic Sectors ($\Sigma_s$) such as perception ($\Sigma_{\text{perc}}$), planning ($\Sigma_{\text{plan}}$), memory/narrative ($\Sigma_{\text{mem}}/\Sigma_{\text{narr}}$), and reflection ($\Sigma_{\text{refl}}$), and a general concept of Abstraction Layers ($\Phi^{(k)}$). This appendix explores:
\begin{enumerate}
	\item a broader, more speculative taxonomy of potential sectors (Subsection~\ref{app_subsec:expanded_sectors});
	\item and discusses nuances in the granularity of abstraction layers, offering further avenues for extending the richness of the agent's cognitive architecture (Subsection~\ref{app_subsec:abstraction_granularity}).
\end{enumerate}

\subsection{Expanded Taxonomy of Potential Semantic Sectors}
\label{app_subsec:expanded_sectors}

The principle of functional modularity suggests that an agent's belief space ($\Phi$) can be stratified into numerous specialized sectors beyond the primary examples. The definition of these sectors would depend on the agent's specific design parameters and its range of capabilities. Below are some potential candidate sectors, along with their hypothetical roles:

\begin{itemize}
	\item \textbf{$\Sigma_{\text{know}}$ (Knowledge/Semantic Memory Sector)}: Distinct from episodic or narrative memory ($\Sigma_{\text{narr}}$), this sector could house more general semantic knowledge, factual information, conceptual hierarchies, and learned regularities about the world. Beliefs here might be highly anchored (\cite{Dumbrava2025TheoreticalFoundations}, Chapter 7) and often reside at moderate to high abstraction levels ($\Phi^{(k)}$).
	\item \textbf{$\Sigma_{\text{goal}}$ (Goal Management Sector)}: While $\Sigma_{\text{plan}}$ handles the construction of action sequences, a dedicated $\Sigma_{\text{goal}}$ could be responsible for the representation, prioritization, conflict resolution, and lifecycle management (e.g., activation, suspension, completion) of the agent's various goals. This sector would interact closely with $\Sigma_{\text{plan}}$ and $\Sigma_{\text{refl}}$.
	\item \textbf{$\Sigma_{\text{sim}}$ (Internal Simulation Sector)}: This sector would be the primary locus for Embodied Simulation processes, where the agent generates and evolves belief trajectories (\cite{Dumbrava2025TheoreticalFoundations}, Chapter 27) representing potential interactions, hypothetical scenarios, or counterfactuals. It might contain sub-regions like $\Sigma_{\text{sim-perc}}$ and $\Sigma_{\text{sim-act}}$ for simulated perceptions and actions.
	\item \textbf{$\Sigma_{\text{affect}}$ (Affective/Motivational Sector)}: As listed in \cite{Dumbrava2025TheoreticalFoundations}, this sector would contain belief fragments tagged with or representing emotional valence, motivational states, or internal drives. It would play a crucial role in Affective-Cognitive Couplings ($g_{\Sigma_X \leftrightarrow \Sigma_{\text{affect}}}^k$, as defined in Section~\ref{ssubsec:taxonomy_affective_cognitive}), influencing decision-making, learning, and attention within specific abstraction levels.
	\item \textbf{$\Sigma_{\text{social}}$ (Social Cognition/Theory of Mind Sector)}: Dedicated to representing and reasoning about the mental states (beliefs, desires, intentions) of other agents, as explored in \cite{Dumbrava2025TheoreticalFoundations} (Chapter 38). This sector would be central for complex social interactions, communication, and collaboration.
	\item \textbf{$\Sigma_{\text{ethic}}$ (Ethical Reasoning Sector)}: For agents designed to operate within human societies, a specialized sector for representing ethical principles, moral rules, value systems, and performing ethical deliberation could be crucial. It would likely interact heavily with $\Sigma_{\text{refl}}$ and $\Sigma_{\text{plan}}$.
	\item \textbf{$\Sigma_{\text{create}}$ (Creative Ideation Sector)}: A sector focused on generating novel ideas, hypotheses, or solutions through mechanisms like divergent thinking, conceptual blending (potentially drawing from diverse other sectors via intra-level couplings to $\Sigma_{\text{create}}^{(k)}$ or via $\Lambda/V$ operators bringing information from other levels), or serendipitous discovery based on Spontaneous Drift (\cite{Dumbrava2025TheoreticalFoundations}, Chapter 16).
	\item \textbf{$\Sigma_{\text{lang-proc}}$ (Language Processing Sector)}: While the entire framework is linguistically grounded, a specific sector (or sub-sectors like $\Sigma_{\text{ling-parse}}$ for input and $\Sigma_{\text{ling-gen}}$ for output) could be dedicated to the intricacies of language processing itself, such as syntactic analysis, disambiguation, pragmatic interpretation, and discourse management. This goes beyond the simple Observation Encoding ($X$) or Elaboration ($V$) for output mentioned generally. The monograph \cite{Dumbrava2025TheoreticalFoundations} also lists $\Sigma_{\text{lang}}$ as a common example.
	\item \textbf{$\Sigma_{\text{motor}}$ or $\Sigma_{\text{exe-interface}}$ (Motor Control/Execution Interface Sector)}: This sector would serve as the immediate interface to the agent's effectors, translating finalized, concrete plans from $\Sigma_{\text{plan}}^{(0)}$ or $\Sigma_{\text{exe}}^{(0)}$ into actual motor commands or action primitives. It ensures detailed grounding of action.
\end{itemize}

The definition of new sectors should be guided by principles of functional coherence (grouping beliefs and processes serving a related overarching function) and the potential for distinct processing pathways or regulatory needs. The emergence or differentiation of such sectors could also be a result of learning and adaptation, as discussed in \cite{Dumbrava2025TheoreticalFoundations} (Chapter 34).

\subsection{Considerations on Abstraction Layer Granularity}
\label{app_subsec:abstraction_granularity}

The Semantic Manifold framework organizes beliefs vertically by Abstraction Layers ($\Phi^{(k)}$), where $k$ is an integer index representing increasing levels of generality. While this discrete layering is a powerful conceptual tool, the "semantic distance" or "degree of abstraction" captured by a single step from $k$ to $k+1$ (mediated by operators like $\Lambda$) may not be uniform across the entire manifold or for all agents.
\begin{itemize}
	\item \textbf{Variable Semantic "Thickness" of Layers}: The amount of generalization or compression achieved by the Abstraction operator ($\Lambda$) in moving from $\Phi^{(k)}$ to $\Phi^{(k+1)}$ could vary. Some abstraction steps might represent minor summarizations, while others could involve significant conceptual leaps. This "thickness" could be an implicit property of the learned $\Lambda$ operator.
	\item \textbf{Agent-Specific Granularity}: The number of distinct, functional abstraction layers an agent utilizes and the perceived granularity between them could be a component of its architectural parameterization. Some agents might operate with a few coarse-grained layers, while others might possess a much finer-grained hierarchy for specific tasks.
	\item \textbf{Sector-Dependent Granularity}: It is plausible that the granularity of abstraction differs across Semantic Sectors. For instance, $\Sigma_{\text{perc}}$ might have many fine-grained layers close to $k=0$ to represent detailed perceptual features, while $\Sigma_{\text{refl}}$ might operate with fewer, more broadly defined abstract layers for high-level self-models.
	\item \textbf{Learned or Adaptive Granularity}: Through experience, an agent might learn to make finer or coarser distinctions in abstraction, effectively adapting the operational meaning of its $\Lambda$ and $V$ operators (Chapter 35 in \cite{Dumbrava2025TheoreticalFoundations}). This could involve learning to "skip" layers or to introduce intermediate sub-layers dynamically.
	\item \textbf{Implications for Scaling Operators ($\Lambda, V$)}: If granularity is variable, the $\Lambda$ and $V$ operators would need to be context-sensitive, potentially adapting their "step size" or the nature of the transformation based on the current sector, the specific beliefs being processed, or learned heuristics. The notion of $m$-fold composition ($\Lambda^m, V^m$) would still hold, but the semantic impact of each individual step might differ.
\end{itemize}
These considerations suggest that while the indexed layering $\Phi^{(k)}$ provides a valuable discrete framework, the underlying semantic scaling might be more continuous or flexibly structured in a sophisticated agent. Understanding this granularity is important for interpreting the meaning of "height" in the semantic manifold and for designing effective abstraction and elaboration mechanisms.

\section{Detailed Characteristics of Coupling Roles}
\label{app:detailed_coupling_roles}

This appendix provides detailed elaborations for each of the intra-level sectoral coupling roles that were summarized in Table~\ref{tab:sectoral_couplings} and briefly introduced in Section~\ref{subsec:taxonomy_couplings}. For each coupling type, this appendix provides a detailed breakdown of its definition, functional role, typical nature, and impact, covering:
\begin{enumerate}
	\item \textbf{Perceptual Integration} couplings, which ground cognition in the current environmental context (Subsection~\ref{app:detail_perceptual_integration});
	\item \textbf{Memory Formation \& Update} couplings, responsible for building the agent's long-term knowledge base (Subsection~\ref{app:detail_memory_formation});
	\item \textbf{Planning \& Goal Processing} couplings, which enable goal-directed reasoning and resource management (Subsection~\ref{app:detail_planning_goal});
	\item \textbf{Reflective \& Meta-Cognitive} couplings, which support self-awareness, error correction, and strategic adjustment (Subsection~\ref{app:detail_reflective_metacognitive});
	\item \textbf{Execution \& Action Control} couplings, which translate internal intentions into overt actions (Subsection~\ref{app:detail_execution_control});
	\item \textbf{Communicative Output} couplings, for externalizing internal states through language (Subsection~\ref{app:detail_communicative_output});
	\item \textbf{Intra-Sectoral Recurrent} couplings, which contribute to cognitive inertia, focus, and stability (Subsection~\ref{app:detail_intra_sectoral});
	\item \textbf{Affective-Cognitive} couplings, which integrate emotion-like states with cognitive processing (Subsection~\ref{app:detail_affective_cognitive}).
\end{enumerate}

\subsection{Perceptual Integration Couplings ($g_{\Sigma_{\text{perc}} \rightarrow \Sigma_X}^k$)}
\label{app:detail_perceptual_integration}

\begin{itemize}
	\item \textbf{Definition}: Quantify the influence of the perceptual sector $\Sigma_{\text{perc}}^{(k)}$, processing belief fragments $X(s)$ from Observation Encoding $X$ of sensory input $s$ that are relevant to abstraction level $k$, on other cognitive sectors $\Sigma_X^{(k)}$.
	\item \textbf{Functional Role}: To ground other cognitive functions at level $k$ in the current environmental context relevant to that level, trigger responses within $k$, and provide raw material for learning and memory formation at level $k$.
	\item \textbf{Typical Nature}: Primarily excitatory and triggering; information-providing for assimilation and learning within level $k$.
	\item \textbf{Impact \& Examples}: $g_{\Sigma_{\text{perc}} \rightarrow \Sigma_{\text{plan}}}^0$: Strong coupling at $k=0$ enables reactive planning (e.g., concrete obstacle detection in $\Sigma_{\text{perc}}^{(0)}$ triggers concrete evasion plan in $\Sigma_{\text{plan}}^{(0)}$). $g_{\Sigma_{\text{perc}} \rightarrow \Sigma_{\text{mem}/\text{narr}}}^k$: Perceptual episodes at level $k$ processed into $\Sigma_{\text{mem}}^{(k)}$. $g_{\Sigma_{\text{perc}} \rightarrow \Sigma_{\text{refl}}}^k$: Significant perceptions at level $k$ trigger reflection in $\Sigma_{\text{refl}}^{(k)}$.
\end{itemize}

\subsection{Memory Formation \& Update Couplings ($g_{\Sigma_X \rightarrow \Sigma_{\text{mem}/\text{narr}}}^k$)}
\label{app:detail_memory_formation}
\begin{itemize}
	\item \textbf{Definition}: Quantify how information from any source sector $\Sigma_X^{(k)}$ is encoded into persistent memory ($\Sigma_{\text{mem}}^{(k)}$, $\Sigma_{\text{narr}}^{(k)}$, or $\Sigma_{\text{knowledge}}^{(k)}$).
	\item \textbf{Functional Role}: To build and maintain the agent's long-term knowledge and experiential record at each abstraction level.
	\item \textbf{Typical Nature}: Excitatory or potentiating, leading to memory trace creation/strengthening within level $k$.
	\item \textbf{Impact \& Examples}: Successful plan outcomes (from $\Sigma_{\text{plan}}^{(k)}$) strongly coupled to $\Sigma_{\text{mem}}^{(k)}$ for storage at that level. Insights from $\Sigma_{\text{refl}}^{(k)}$ processed into $\Sigma_{\text{knowledge}}^{(k)}$ or $\Sigma_{\text{narr}}^{(k)}$.
\end{itemize}

\subsection{Planning \& Goal Processing Couplings ($g_{\Sigma_X \leftrightarrow \Sigma_{\text{plan}}}^k$)}
\label{app:detail_planning_goal}
\begin{itemize}
	\item \textbf{Definition}: Bidirectional, intra-level couplings. $g_{\Sigma_X^{(k)} \rightarrow \Sigma_{\text{plan}}^{(k)}}$: How inputs from $\Sigma_X^{(k)}$ shape goal/plan dynamics in $\Sigma_{\text{plan}}^{(k)}$. $g_{\Sigma_{\text{plan}}^{(k)} \rightarrow \Sigma_X^{(k)}}$: How active plans/goals in $\Sigma_{\text{plan}}^{(k)}$ direct processing in $\Sigma_X^{(k)}$.
	\item \textbf{Functional Role}: To enable goal-directed reasoning and resource orchestration at abstraction level $k$.
	\item \textbf{Typical Nature}: $g_{\Sigma_X^{(k)} \rightarrow \Sigma_{\text{plan}}^{(k)}}$: Can be triggering, excitatory, or inhibitory for planning processes at level $k$. $g_{\Sigma_{\text{plan}}^{(k)} \rightarrow \Sigma_X^{(k)}}$: Directive, modulatory, or biasing for processes in $\Sigma_X^{(k)}$.
	\item \textbf{Impact \& Examples}: A safety goal from $\Sigma_{\text{refl}}^{(k)}$ (e.g., "Maintain altitude stability at $k=1$") constrains plans in $\Sigma_{\text{plan}}^{(k)}$. Active plan in $\Sigma_{\text{plan}}^{(k)}$ heightens sensitivity of $A$ operator to task-relevant perceptual cues assimilated into $\Sigma_{\text{perc}}^{(k)}$.
\end{itemize}

\subsection{Reflective \& Meta-Cognitive Couplings ($g_{\Sigma_X \leftrightarrow \Sigma_{\text{refl}}}^k$)}
\label{app:detail_reflective_metacognitive}
\begin{itemize}
	\item \textbf{Definition}: Bidirectional, intra-level couplings. Afferent ($g^{k}_{\Sigma_X \rightarrow \Sigma_{\text{refl}}}$): How events in $\Sigma_X^{(k)}$ trigger meta-belief formation via Meta-Assimilation ($M$) into $\Sigma_{\text{refl}}^{(k)}$. Efferent ($g^{k}_{\Sigma_{\text{refl}} \rightarrow \Sigma_X}$): How meta-beliefs from $\Sigma_{\text{refl}}^{(k)}$ guide processing in $\Sigma_X^{(k)}$.
	\item \textbf{Functional Role}: To enable self-awareness, self-evaluation, error correction, strategic adjustment, coherence maintenance, and value alignment relevant to abstraction level $k$.
	\item \textbf{Typical Nature}: Afferent: Triggering/excitatory for meta-belief formation in $\Sigma_{\text{refl}}^{(k)}$. Efferent: Modulatory, inhibitory, excitatory, or gating for operations in $\Sigma_X^{(k)}$.
	\item \textbf{Impact \& Examples}: Planning failure in $\Sigma_{\text{plan}}^{(k)}$ triggers meta-assimilation into $\Sigma_{\text{refl}}^{(k)}$, leading to meta-beliefs that initiate plan revision in $\Sigma_{\text{plan}}^{(k)}$ via $g_{\Sigma_{\text{refl}} \rightarrow \Sigma_{\text{plan}}}^k$.
\end{itemize}

\subsection{Execution \& Action Control Couplings ($g_{(\Sigma_{\text{plan}}/\Sigma_{\text{refl}}) \rightarrow \Sigma_{\text{exe}}}^k$)}
\label{app:detail_execution_control}
\begin{itemize}
	\item \textbf{Definition}: Quantify how finalized plans from $\Sigma_{\text{plan}}^{(k)}$ or decisive meta-beliefs from $\Sigma_{\text{refl}}^{(k)}$ influence an action execution system/sector ($\Sigma_{\text{exe}}^{(k)}$) at the same level $k$, linking to "Semantic Execution" and "Activation Basins" relevant to that level (see \cite{Dumbrava2025TheoreticalFoundations}, Chapters 25 and 26).
	\item \textbf{Functional Role}: To translate internal intentions at level $k$ into overt actions or inputs for lower-level elaboration; involved in action readiness and gating at level $k$.
	\item \textbf{Typical Nature}: Primarily triggering or gating for execution processes originating from level $k$; can be excitatory for initiating action execution.
	\item \textbf{Impact \& Examples}: Coherent plan from $\Sigma_{\text{plan}}^{(0)}$ coupled to $\Sigma_{\text{exe}}^{(0)}$ initiates concrete actions. Meta-belief from $\Sigma_{\text{refl}}^{(k)}$ indicating risk might inhibit execution plans within $\Sigma_{\text{plan}}^{(k)}$.
\end{itemize}

\subsection{Communicative Output Couplings ($g_{\Sigma_X \rightarrow \Sigma_{\text{lang-out}}}^k$)}
\label{app:detail_communicative_output}
\begin{itemize}
	\item \textbf{Definition}: Quantify how internal beliefs from $\Sigma_X^{(k)}$ are selected, transformed, and conveyed to a language output sector ($\Sigma_{\text{lang-out}}^{(k)}$ or a general output buffer) at the same abstraction level $k$.
	\item \textbf{Functional Role}: To enable the agent to externalize internal states, knowledge, or reasoning from a specific abstraction layer $k$ via language.
	\item \textbf{Typical Nature}: Information-providing and triggering for language generation; can be selective.
	\item \textbf{Impact \& Examples}: Strong $g_{\Sigma_{\text{refl}} \rightarrow \Sigma_{\text{lang-out}}}^1$ characterizes an agent adept at explaining its abstract reasoning from $\Sigma_{\text{refl}}^{(1)}$. Beliefs from $\Sigma_{\text{knowledge}}^{(k)}$ can be selected for communication at level $k$.
\end{itemize}

\subsection{Intra-Sectoral Recurrent Couplings ($g_{ii}^k$)}
\label{app:detail_intra_sectoral}
\begin{itemize}
	\item \textbf{Definition}: Quantify influence of $\Sigma_i$'s current state (at $k$) on its own subsequent states within the same sector $\Sigma_i^{(k)}$.
	\item \textbf{Functional Role}: To contribute to cognitive inertia, focus maintenance, pattern completion, and state stability within $\Sigma_i^{(k)}$.
	\item \textbf{Typical Nature}: Excitatory (reinforcing) or inhibitory (self-damping) for activity within $\Sigma_i^{(k)}$.
	\item \textbf{Impact \& Examples}: Strong positive $g_{\Sigma_{\text{plan}} \rightarrow \Sigma_{\text{plan}}}^k$ can lead to goal persistence at level $k$. Recurrence in $\Sigma_{\text{perc}}^{(k)}$ might facilitate perceptual completion at that level. Overly strong recurrence can lead to unproductive loops within $\Sigma_i^{(k)}$.
\end{itemize}

\subsection{Affective-Cognitive Couplings ($g_{\Sigma_X \leftrightarrow \Sigma_{\text{affect}}}^k$)}
\label{app:detail_affective_cognitive}

\begin{itemize}
	\item \textbf{Definition}: Bidirectional, intra-level influences between cognitive sectors $\Sigma_X^{(k)}$ and an affective sector $\Sigma_{\text{affect}}^{(k)}$ representing emotion-like states at abstraction level $k$.
	\item \textbf{Functional Role}: To integrate emotion-like states at level $k$ with cognition at level $k$, biasing attention, memory, decision-making, learning, motivation within that layer.
	\item \textbf{Typical Nature}: $\Sigma_X^{(k)} \rightarrow \Sigma_{\text{affect}}^{(k)}$: Triggering/excitatory for affective state generation in $\Sigma_{\text{affect}}^{(k)}$. $\Sigma_{\text{affect}}^{(k)} \rightarrow \Sigma_X^{(k)}$: Primarily modulatory, biasing operations in $\Sigma_X^{(k)}$.
	\item \textbf{Impact \& Examples}: Perceived threat in $\Sigma_{\text{perc}}^{(0)}$ coupled to $\Sigma_{\text{affect}}^{(0)}$ generates "concrete fear" state; "concrete fear" state in $\Sigma_{\text{affect}}^{(0)}$ biases $\Sigma_{\text{plan}}^{(0)}$ towards immediate escape. Abstract threat in $\Sigma_{\text{perc}}^{(1)}$ coupled to $\Sigma_{\text{affect}}^{(1)}$ could generate "anxiety" which biases $\Sigma_{\text{plan}}^{(1)}$ towards cautious long-term planning.
\end{itemize}

\section{Exemplary Cognitive Processing Sequences}
\label{app:exemplary_cognitive_sequences_intros}
This appendix provides a collection of illustrative cognitive processing sequences to demonstrate how different Semantic Sectors ($\Sigma_s$) and Abstraction Layers ($k$) interact via intra-level couplings ($g_{ij}^k$) and inter-level operators ($\Lambda, V$). The sequences showcase a range of cognitive behaviors discussed throughout this work and are organized thematically to cover:
\begin{enumerate}
	\item \textbf{Reactive and Basic Action Loops}, which model fast, low-level responses such as reflex-like actions (Subsection~\ref{app_seq:direct_reactive_loop}) and simple memory formation (Subsection~\ref{app_seq:episodic_memory_formation});
	\item \textbf{Deliberative and Reflective Loops}, which illustrate more complex cognition involving memory-informed planning (Subsection~\ref{app_seq:mem_informed_abstract_planning}) and multi-level reflection (Subsection~\ref{app_seq:multilevel_reflection_replanning});
	\item \textbf{Learning and Adaptation Loops}, demonstrating how an agent might learn from experience (Subsection~\ref{app_seq:experiential_learning_outcome}) and refine skills (Subsection~\ref{app_seq:skill_refinement_trace_analysis});
	\item \textbf{Meta-Cognitive and Self-Regulatory Loops}, focusing on the agent's capacity to monitor and control its own processes, such as managing cognitive load (Subsection~\ref{app_seq:cog_load_management}) or ensuring belief coherence (Subsection~\ref{app_seq:coherence_monitoring_revision});
	\item \textbf{Affect-Driven and Social Loops}, which incorporate emotional states and reasoning about other agents, such as anxiety-modulated planning (Subsection~\ref{app_seq:abstract_threat_anxiety_planning}) and inferring intent (Subsection~\ref{app_seq:social_interaction_intent});
	\item \textbf{Communicative and Creative Loops}, outlining processes for generating explanations (Subsection~\ref{app_seq:explanation_abstract_insights}) and novel hypotheses (Subsection~\ref{app_seq:creative_ideation_hypothesis});
	\item \textbf{Complex Multi-Sector, Multi-Level Integrative Loops}, which combine numerous components for advanced tasks like ethical deliberation (Subsection~\ref{app_seq:ethical_deliberation_action}) and problem-solving with internal simulation (Subsection~\ref{app_seq:problem_solving_simulation}).
\end{enumerate}
The notation used in the following sequences is consistent with the conventions established in the main body of the work.

\subsection{Reactive and Basic Action Loops}
These sequences are typically fast, involve lower abstraction levels, and represent direct responses to stimuli or simple internal triggers.

\subsubsection{Direct Reactive Loop (Reflex Arc Analog)} \label{app_seq:direct_reactive_loop}
This first sequence illustrates the most fundamental type of reactive behavior, analogous to a simple reflex arc, where perception leads directly to action with minimal processing.

\[\boxed{ \Sigma_{\text{perc}}^{(0)} \xrightarrow{g_{\text{perc} \rightarrow \text{plan}}^0} \Sigma_{\text{plan}}^{(0)} \xrightarrow{g_{\text{plan} \rightarrow \text{exe}}^0} \Sigma_{\text{exe}}^{(0)} \rightarrow \text{Environment} \rightarrow \Sigma_{\text{perc}}^{(0)}} \]
\textit{Process Analysis:}
\begin{enumerate}
	\item Concrete perceptual input (e.g., "Obstacle Ahead!") arrives in $\Sigma_{\text{perc}}^{(0)}$.
	\item This input directly influences the concrete planning sector $\Sigma_{\text{plan}}^{(0)}$ via the intra-level coupling $g_{\text{perc} \rightarrow \text{plan}}^0$, potentially generating a plan like "Initiate Evasion!".
	\item The concrete plan in $\Sigma_{\text{plan}}^{(0)}$ influences the execution sector $\Sigma_{\text{exe}}^{(0)}$ via $g_{\text{plan} \rightarrow \text{exe}}^0$.
	\item The agent executes the action, which affects the Environment.
	\item Changes in the Environment are perceived by $\Sigma_{\text{perc}}^{(0)}$, closing the loop.
\end{enumerate}

\subsubsection{Affect-Modulated Reactive Loop} \label{app_seq:affect_modulated_reactive_loop}
The following sequence demonstrates how immediate affective states can modulate basic reactive responses at the concrete level.
\[ \boxed{\Sigma_{\text{perc}}^{(0)} \xrightarrow{g_{\text{perc} \rightarrow \text{affect}}^0} \Sigma_{\text{affect}}^{(0)} \xrightarrow{g_{\text{affect} \rightarrow \text{plan}}^0} \Sigma_{\text{plan}}^{(0)} \xrightarrow{g_{\text{plan} \rightarrow \text{exe}}^0} \Sigma_{\text{exe}}^{(0)} \rightarrow \text{Environment} \rightarrow \Sigma_{\text{perc}}^{(0)}} \]
\textit{Process Analysis:}
\begin{enumerate}
	\item Concrete perceptual input arrives in $\Sigma_{\text{perc}}^{(0)}$.
	\item This input influences the concrete affective sector $\Sigma_{\text{affect}}^{(0)}$ via $g_{\text{perc} \rightarrow \text{affect}}^0$, generating an affective state (e.g., "fear").
	\item The affective state in $\Sigma_{\text{affect}}^{(0)}$ biases planning in $\Sigma_{\text{plan}}^{(0)}$ via $g_{\text{affect} \rightarrow \text{plan}}^0$.
	\item The (potentially biased) concrete plan influences $\Sigma_{\text{exe}}^{(0)}$ for execution. 
	\item Action affects the Environment, with feedback to $\Sigma_{\text{perc}}^{(0)}$.
\end{enumerate}

\subsubsection{Simple Episodic Memory Formation} \label{app_seq:episodic_memory_formation}
This sequence outlines the direct encoding of experiences into memory, forming the basis of episodic recall.
\[ \boxed{\Sigma_{\text{perc}}^{(0)} \xrightarrow{g_{\text{perc} \rightarrow \text{mem}}^0} \Sigma_{\text{mem}}^{(0)}} \]
\textit{Process Analysis:}
\begin{enumerate}
	\item A concrete perceptual event (e.g., $\varphi_{\text{event}}$) occurs in $\Sigma_{\text{perc}}^{(0)}$.
	\item This event influences the memory sector $\Sigma_{\text{mem}}^{(0)}$ via $g_{\text{perc} \rightarrow \text{mem}}^0$.
	\item The outcome is the storage of a concrete episodic trace $\varphi_{\text{episode}}$ related to $\varphi_{\text{event}}$ in $\Sigma_{\text{mem}}^{(0)}$.
\end{enumerate}

\subsection{Deliberative and Reflective Loops}
These sequences involve higher abstraction levels, memory retrieval, and reflective processing, characteristic of more thoughtful or complex cognition.

\subsubsection{Perception-Reflection-Planning-Execution Cycle} \label{app_seq:perc_refl_plan_exec_cycle}
The sequence below demonstrates a comprehensive cycle where initial perceptions are reflected upon, abstracted for strategic planning, and then elaborated for execution.
\[ \boxed{ \Sigma_{\text{perc}}^{(0)} \xrightarrow{g_{\text{perc} \rightarrow \text{refl}}^0} \Sigma_{\text{refl}}^{(0)} \xrightarrow{\Lambda} \Sigma_{\text{refl}}^{(1)} \xrightarrow{g_{\text{refl} \rightarrow \text{plan}}^1} \Sigma_{\text{plan}}^{(1)} \xrightarrow{V} \Sigma_{\text{exe}}^{(0)} \rightarrow \text{Environment} \rightarrow \Sigma_{\text{perc}}^{(0)}} \]
\textit{Process Analysis:}
\begin{enumerate}
	\item Concrete perceptual input is processed in $\Sigma_{\text{perc}}^{(0)}$.
	\item This input influences the concrete reflection sector $\Sigma_{\text{refl}}^{(0)}$ via $g_{\text{perc} \rightarrow \text{refl}}^0$.
	\item Content from $\Sigma_{\text{refl}}^{(0)}$ (e.g., an initial assessment) is abstracted by operator $\Lambda$ to $\Sigma_{\text{refl}}^{(1)}$.
	\item Abstract reflection in $\Sigma_{\text{refl}}^{(1)}$ influences abstract planning in $\Sigma_{\text{plan}}^{(1)}$ via $g_{\text{refl} \rightarrow \text{plan}}^1$.
	\item The abstract plan from $\Sigma_{\text{plan}}^{(1)}$ is elaborated by operator $V$ into concrete commands in $\Sigma_{\text{exe}}^{(0)}$.
	\item Action upon the Environment occurs. 
	\item Environmental feedback returns to $\Sigma_{\text{perc}}^{(0)}$.
\end{enumerate}

\subsubsection{Memory-Informed Abstract Planning and Reflection} \label{app_seq:mem_informed_abstract_planning}
This next sequence shows how abstract planning can be informed by querying memory and further refined by reflective deliberation before action.

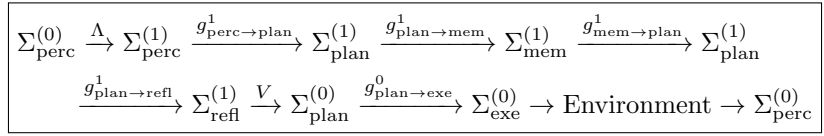
\begin{empheq}[box=\fbox]{align*}
	& \Sigma_{\text{perc}}^{(0)} \xrightarrow{\Lambda} \Sigma_{\text{perc}}^{(1)} \xrightarrow{g_{\text{perc} \rightarrow \text{plan}}^1} \Sigma_{\text{plan}}^{(1)} \xrightarrow{g_{\text{plan} \rightarrow \text{mem}}^1} \Sigma_{\text{mem}}^{(1)} \xrightarrow{g_{\text{mem} \rightarrow \text{plan}}^1} \Sigma_{\text{plan}}^{(1)} \\
	& \qquad \xrightarrow{g_{\text{plan} \rightarrow \text{refl}}^1} \Sigma_{\text{refl}}^{(1)} \xrightarrow{V} \Sigma_{\text{plan}}^{(0)} \xrightarrow{g_{\text{plan} \rightarrow \text{exe}}^0} \Sigma_{\text{exe}}^{(0)} \rightarrow \text{Environment} \rightarrow \Sigma_{\text{perc}}^{(0)}
\end{empheq}

\textit{Process Analysis:}
\begin{enumerate}
	\item Concrete perception from $\Sigma_{\text{perc}}^{(0)}$ is abstracted via $\Lambda$ to $\Sigma_{\text{perc}}^{(1)}$.
	\item Abstracted perception influences abstract planning in $\Sigma_{\text{plan}}^{(1)}$ via $g_{\text{perc} \rightarrow \text{plan}}^1$.
	\item The abstract plan in $\Sigma_{\text{plan}}^{(1)}$ triggers a query to abstract memory $\Sigma_{\text{mem}}^{(1)}$ (an effect of $g_{\text{plan} \rightarrow \text{mem}}^1$).
	\item Relevant abstract memories are retrieved from $\Sigma_{\text{mem}}^{(1)}$ and influence $\Sigma_{\text{plan}}^{(1)}$ (an effect of $g_{\text{mem} \rightarrow \text{plan}}^1$).
	\item The memory-informed abstract plan in $\Sigma_{\text{plan}}^{(1)}$ influences abstract reflection in $\Sigma_{\text{refl}}^{(1)}$ via $g_{\text{plan} \rightarrow \text{refl}}^1$ for deliberation.
	\item The deliberated abstract plan is elaborated via $V$ to a concrete plan in $\Sigma_{\text{plan}}^{(0)}$.
	\item This concrete plan guides execution via $\Sigma_{\text{exe}}^{(0)}$. 
	\item Action affects the Environment, with feedback to $\Sigma_{\text{perc}}^{(0)}$.
\end{enumerate}

\subsubsection{Goal-Driven Deliberation Initiated by Abstract Goal} \label{app_seq:goal_driven_deliberation}
Here, an abstract goal initiates a deliberative process that involves elaboration to concrete planning and memory consultation.
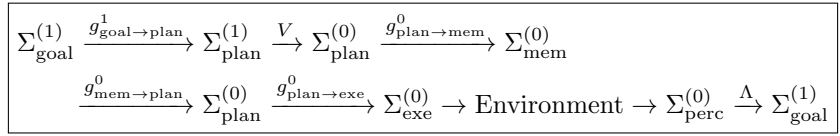
\begin{empheq}[box=\fbox]{align*}
	& \Sigma_{\text{goal}}^{(1)} \xrightarrow{g_{\text{goal} \rightarrow \text{plan}}^1} \Sigma_{\text{plan}}^{(1)} \xrightarrow{V} \Sigma_{\text{plan}}^{(0)} \xrightarrow{g_{\text{plan} \rightarrow \text{mem}}^0} \Sigma_{\text{mem}}^{(0)} \\
	& \qquad \xrightarrow{g_{\text{mem} \rightarrow \text{plan}}^0} \Sigma_{\text{plan}}^{(0)} \xrightarrow{g_{\text{plan} \rightarrow \text{exe}}^0} \Sigma_{\text{exe}}^{(0)} \rightarrow \text{Environment} \rightarrow \Sigma_{\text{perc}}^{(0)} \xrightarrow{\Lambda} \Sigma_{\text{goal}}^{(1)}
\end{empheq}

\textit{Process Analysis:}
\begin{enumerate}
	\item An abstract goal resides in $\Sigma_{\text{goal}}^{(1)}$ (as per Appendix~\ref{app:sectors_abstraction_granularity}).
	\item This goal influences abstract planning in $\Sigma_{\text{plan}}^{(1)}$ via $g_{\text{goal} \rightarrow \text{plan}}^1$.
	\item The abstract plan is elaborated by $V$ to a concrete plan in $\Sigma_{\text{plan}}^{(0)}$.
	\item The concrete plan augments concrete memory $\Sigma_{\text{mem}}^{(0)}$ (via $g_{\text{plan} \rightarrow \text{mem}}^0$).
	\item Retrieved concrete memories refine the plan in $\Sigma_{\text{plan}}^{(0)}$ (via $g_{\text{mem} \rightarrow \text{plan}}^0$).
	\item The refined concrete plan guides execution through $\Sigma_{\text{exe}}^{(0)}$. 
	\item Environmental feedback is perceived by $\Sigma_{\text{perc}}^{(0)}$, then abstracted by $\Lambda$ to potentially update the abstract goal in $\Sigma_{\text{goal}}^{(1)}$.
\end{enumerate}

\subsubsection{Multi-Level Reflection and Iterative Re-Planning} \label{app_seq:multilevel_reflection_replanning}
This sequence illustrates a deep reflective process, where initial reflections are progressively abstracted to higher levels for more profound consideration before leading to a high-level plan.
\[ \boxed{\Sigma_{\text{perc}}^{(0)} \xrightarrow{g_{\text{perc} \rightarrow \text{refl}}^0} \Sigma_{\text{refl}}^{(0)} \xrightarrow{\Lambda} \Sigma_{\text{refl}}^{(1)} \xrightarrow{g_{\text{refl} \rightarrow \text{refl}}^1} \Sigma_{\text{refl}}^{(1)} \xrightarrow{\Lambda} \Sigma_{\text{refl}}^{(2)} \xrightarrow{g_{\text{refl} \rightarrow \text{plan}}^2} \Sigma_{\text{plan}}^{(2)} \xrightarrow{V} \Sigma_{\text{plan}}^{(1)} \xrightarrow{V} \Sigma_{\text{exe}}^{(0)} \rightarrow \text{Environment} }\]
\textit{Process Analysis:}
\begin{enumerate}
	\item Concrete perception in $\Sigma_{\text{perc}}^{(0)}$ influences concrete reflection in $\Sigma_{\text{refl}}^{(0)}$ via $g_{\text{perc} \rightarrow \text{refl}}^0$.
	\item This reflection is abstracted by $\Lambda$ to $\Sigma_{\text{refl}}^{(1)}$. 
	\item Sustained abstract reflection occurs in $\Sigma_{\text{refl}}^{(1)}$, potentially involving intra-sectoral recurrence $g_{\text{refl} \rightarrow \text{refl}}^1$.
	\item Further abstraction by $\Lambda$ elevates reflection to $\Sigma_{\text{refl}}^{(2)}$ for deeper consideration.
	\item Highly abstract reflection in $\Sigma_{\text{refl}}^{(2)}$ influences planning in $\Sigma_{\text{plan}}^{(2)}$ via $g_{\text{refl} \rightarrow \text{plan}}^2$.
	\item The abstract plan is then progressively detailed by $V$ through $\Sigma_{\text{plan}}^{(1)}$ to $\Sigma_{\text{exe}}^{(0)}$ for execution.
\end{enumerate}

\subsection{Learning and Adaptation Loops}
These sequences illustrate how an agent might learn from experience, updating its knowledge or memory.

\subsubsection{Experiential Learning (Outcome Evaluation Modifies Abstract Knowledge)} \label{app_seq:experiential_learning_outcome}
The following process shows how an agent might learn from the outcomes of its actions by reflecting on them and updating its abstract knowledge base.
\[\boxed{ \Sigma_{\text{perc}}^{(0)} \xrightarrow{g_{\text{perc} \rightarrow \text{plan}}^0} \Sigma_{\text{plan}}^{(0)} \xrightarrow{g_{\text{plan} \rightarrow \text{exe}}^0} \Sigma_{\text{exe}}^{(0)} \rightarrow \text{Environment} \rightarrow \Sigma_{\text{perc}}^{(0)} \xrightarrow{g_{\text{perc} \rightarrow \text{refl}}^0} \Sigma_{\text{refl}}^{(0)} \xrightarrow{\Lambda} \Sigma_{\text{refl}}^{(1)} \xrightarrow{g_{\text{refl} \rightarrow \text{know}}^1} \Sigma_{\text{know}}^{(1)} }\]
\textit{Process Analysis:}
\begin{enumerate}
	\item Agent perceives input in $\Sigma_{\text{perc}}^{(0)}$, plans in $\Sigma_{\text{plan}}^{(0)}$, and executes an action via $\Sigma_{\text{exe}}^{(0)}$.
	\item The action leads to an outcome in the Environment. 
	\item The agent perceives this outcome in $\Sigma_{\text{perc}}^{(0)}$.
	\item The perceived outcome influences concrete reflection in $\Sigma_{\text{refl}}^{(0)}$ for evaluation. 
	\item This evaluation is abstracted by $\Lambda$ to $\Sigma_{\text{refl}}^{(1)}$.
	\item Abstract reflection influences the abstract knowledge base $\Sigma_{\text{know}}^{(1)}$ (as per Appendix~\ref{app:sectors_abstraction_granularity}) via $g_{\text{refl} \rightarrow \text{know}}^1$, leading to an update of knowledge.
\end{enumerate}

\subsubsection{Skill Refinement via Execution Trace Analysis} \label{app_seq:skill_refinement_trace_analysis}
This sequence details a skill refinement mechanism where feedback from action execution is analyzed and used to update memory related to skill components.
\[\boxed{ \Sigma_{\text{plan}}^{(1)} \xrightarrow{V} \Sigma_{\text{plan}}^{(0)} \xrightarrow{g_{\text{plan} \rightarrow \text{exe}}^0} \Sigma_{\text{exe}}^{(0)} \rightarrow \Sigma_{\text{perc}}^{(0)} \xrightarrow{g_{\text{perc} \rightarrow \text{refl}}^0} \Sigma_{\text{refl}}^{(0)} \xrightarrow{g_{\text{refl} \rightarrow \text{mem}}^0} \Sigma_{\text{mem}}^{(0)} }\]
\textit{Process Analysis:}
\begin{enumerate}
	\item An abstract skill or plan from $\Sigma_{\text{plan}}^{(1)}$ is elaborated by $V$ into concrete steps in $\Sigma_{\text{plan}}^{(0)}$.
	\item These steps are executed via $\Sigma_{\text{exe}}^{(0)}$, generating performance data. 
	\item Performance data (e.g., errors, efficiency) are perceived in $\Sigma_{\text{perc}}^{(0)}$.
	\item Perceived performance data influence concrete reflection in $\Sigma_{\text{refl}}^{(0)}$ for analysis.
	\item The analysis influences $\Sigma_{\text{mem}}^{(0)}$ via $g_{\text{refl} \rightarrow \text{mem}}^0$ to store or refine concrete skill components or execution traces.
\end{enumerate}

\subsection{Meta-Cognitive and Self-Regulatory Loops}
These loops focus on the agent's capacity to monitor and regulate its own cognitive processes.

\subsubsection{Cognitive Load Management and Resource Re-allocation} \label{app_seq:cog_load_management}
The following sequence describes a mechanism for managing cognitive load by monitoring sectoral activity and adjusting resource allocation.
\[ \boxed{ \Sigma_X^{(k)} \xrightarrow{M} \Sigma_{\text{refl}}^{(k)} \xrightarrow{\Lambda} \Sigma_{\text{refl}}^{(k+1)} \xrightarrow{V} \Sigma_{\text{refl}}^{(k)} \xrightarrow{g_{\text{refl} \rightarrow X}^k} \Sigma_X^{(k)} }
\]
\textit{Process Analysis:}
\begin{enumerate}
	\item High activity (cognitive load $\lambda$) is detected in an operational sector $\Sigma_X^{(k)}$.
	\item This triggers Meta-Assimilation ($M$) of this load information into $\Sigma_{\text{refl}}^{(k)}$ as a meta-belief (e.g., "High load in $\Sigma_X^{(k)}$").
	\item The meta-belief is abstracted by $\Lambda$ to $\Sigma_{\text{refl}}^{(k+1)}$ for higher-level policy consideration (e.g., forming policy "Reduce effort to $\Sigma_X^{(k)}$").
	\item The policy is elaborated by $V$ back to $\Sigma_{\text{refl}}^{(k)}$ as a specific directive.
	\item This directive influences $\Sigma_X^{(k)}$ via an (e.g., inhibitory) intra-level coupling $g_{\text{refl} \rightarrow X}^k$, reducing its activity or allocated effort.
\end{enumerate}

\subsubsection{Abstract Coherence Monitoring and Belief Revision} \label{app_seq:coherence_monitoring_revision}
This sequence outlines how an agent might maintain belief coherence by reflectively monitoring for inconsistencies and triggering corrective processes.
\[ \boxed{ (\Sigma_{\text{know}}^{(1)} + \Sigma_{\text{perc}}^{(1)}) \xrightarrow{g_{\text{know} \rightarrow \text{refl}}^1, g_{\text{perc} \rightarrow \text{refl}}^1} \Sigma_{\text{refl}}^{(1)} \xrightarrow{g_{\text{refl} \rightarrow \text{know}}^1} \Sigma_{\text{know}}^{(1)} } \]
\textit{Process Analysis:}
\begin{enumerate}
	\item Existing abstract knowledge in $\Sigma_{\text{know}}^{(1)}$ (e.g., belief $\varphi_A$) and newly abstracted perceptual information in $\Sigma_{\text{perc}}^{(1)}$ (e.g., belief $\varphi_B$, derived from $\Lambda(\Sigma_{\text{perc}}^{(0)})$) are present.
	\item Both influence abstract reflection in $\Sigma_{\text{refl}}^{(1)}$ via respective couplings. 
	\item $\Sigma_{\text{refl}}^{(1)}$ detects incoherence ($\kappa$) between $\varphi_A$ and $\varphi_B$.
	\item This detected incoherence influences $\Sigma_{\text{know}}^{(1)}$ via $g_{\text{refl} \rightarrow \text{know}}^1$, triggering corrective assimilation ($A_{\text{corr}}$) to revise beliefs and restore coherence.
\end{enumerate}

\subsection{Affect-Driven and Social Loops}
These sequences incorporate affective states or reasoning about other agents.

\subsubsection{Abstract Threat Assessment and Anxiety-Modulated Planning} \label{app_seq:abstract_threat_anxiety_planning}
This sequence shows how abstracted perceptions can trigger affective states that, in turn, modulate higher-level planning.
\[ \boxed{ \Sigma_{\text{perc}}^{(0)} \xrightarrow{\Lambda} \Sigma_{\text{perc}}^{(1)} \xrightarrow{g_{\text{perc} \rightarrow \text{affect}}^1} \Sigma_{\text{affect}}^{(1)} \xrightarrow{g_{\text{affect} \rightarrow \text{plan}}^1} \Sigma_{\text{plan}}^{(1)} \xrightarrow{V} \Sigma_{\text{plan}}^{(0)} \rightarrow \Sigma_{\text{exe}}^{(0)} } \]
\textit{Process Analysis:}
\begin{enumerate}
	\item Ambiguous or complex sensory input from $\Sigma_{\text{perc}}^{(0)}$ is abstracted by $\Lambda$ to $\Sigma_{\text{perc}}^{(1)}$.
	\item In $\Sigma_{\text{perc}}^{(1)}$, an abstract cue (e.g., "potential threat") is identified.
	\item This cue influences the abstract affective sector $\Sigma_{\text{affect}}^{(1)}$ via $g_{\text{perc} \rightarrow \text{affect}}^1$, generating an affective state (e.g., "anxiety").
	\item The affective state in $\Sigma_{\text{affect}}^{(1)}$ biases abstract planning in $\Sigma_{\text{plan}}^{(1)}$ via $g_{\text{affect} \rightarrow \text{plan}}^1$ (e.g., towards cautious or avoidant strategies).
	\item The biased abstract plan is elaborated by $V$ to $\Sigma_{\text{plan}}^{(0)}$ and subsequently to $\Sigma_{\text{exe}}^{(0)}$ for action.
\end{enumerate}

\subsubsection{Simplified Social Interaction (Inferring Intent and Responding)} \label{app_seq:social_interaction_intent}
The following describes a simplified social interaction involving inferring another agent's intent and formulating a communicative response.
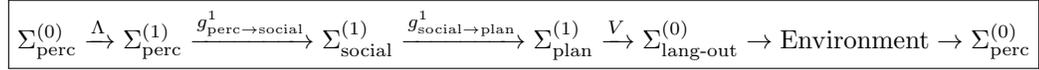
\[ \boxed{ \Sigma_{\text{perc}}^{(0)} \xrightarrow{\Lambda} \Sigma_{\text{perc}}^{(1)} \xrightarrow{g_{\text{perc} \rightarrow \text{social}}^1} \Sigma_{\text{social}}^{(1)} \xrightarrow{g_{\text{social} \rightarrow \text{plan}}^1} \Sigma_{\text{plan}}^{(1)} \xrightarrow{V} \Sigma_{\text{lang-out}}^{(0)} \rightarrow \text{Environment} \rightarrow \Sigma_{\text{perc}}^{(0)} } \]
\textit{Process Analysis:}
\begin{enumerate}
	\item The agent perceives another agent's action in $\Sigma_{\text{perc}}^{(0)}$.
	\item This perception is abstracted by $\Lambda$ to $\Sigma_{\text{perc}}^{(1)}$. 
	\item The abstracted percept influences a social cognition sector $\Sigma_{\text{social}}^{(1)}$ (as per Appendix~\ref{app:sectors_abstraction_granularity}) via $g_{\text{perc} \rightarrow \text{social}}^1$, leading to an inference about the other agent's intent (e.g., $\varphi_{\text{intent}}$).
	\item This inferred intent guides abstract planning in $\Sigma_{\text{plan}}^{(1)}$ via $g_{\text{social} \rightarrow \text{plan}}^1$.
	\item The responsive plan is elaborated by $V$ to the language output sector $\Sigma_{\text{lang-out}}^{(0)}$ to generate a concrete utterance.
	\item The utterance affects the Environment (specifically, the other agent).
	\item The other agent's reaction is perceived by $\Sigma_{\text{perc}}^{(0)}$, closing the social loop.
\end{enumerate}

\subsection{Communicative and Creative Loops}
These sequences involve language generation or novel idea formation.

\subsubsection{Explanation of Abstract Reflective Insights} \label{app_seq:explanation_abstract_insights}
This sequence outlines how an agent might articulate its abstract reflective insights through language.
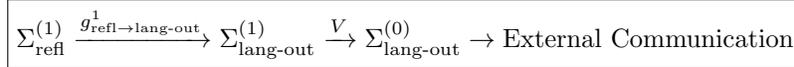
\[ \boxed{ \Sigma_{\text{refl}}^{(1)} \xrightarrow{g_{\text{refl} \rightarrow \text{lang-out}}^1} \Sigma_{\text{lang-out}}^{(1)} \xrightarrow{V} \Sigma_{\text{lang-out}}^{(0)} \rightarrow \text{External Communication} } \]
\textit{Process Analysis:}
\begin{enumerate}
	\item An abstract insight or line of reasoning (e.g., $\varphi_{\text{insight}}$) is formed in $\Sigma_{\text{refl}}^{(1)}$.
	\item This insight influences the language output sector at an abstract level, $\Sigma_{\text{lang-out}}^{(1)}$, via $g_{\text{refl} \rightarrow \text{lang-out}}^1$, preparing an abstract message.
	\item The abstract message is elaborated by $V$ into a concrete linguistic formulation in $\Sigma_{\text{lang-out}}^{(0)}$.
	\item The formulated explanation is conveyed as External Communication. 
\end{enumerate}

\subsubsection{Creative Ideation and Hypothesis Generation (using $\Sigma_{\text{create}}$)}

\label{app_seq:creative_ideation_hypothesis}
The following illustrates a creative process where information from memory and perception fuels idea generation, which is then evaluated and planned.
\[ \boxed{ (\Sigma_{\text{mem}}^{(0..1)} + \Sigma_{\text{perc}}^{(0..1)}) \xrightarrow{g_{\text{mem} \rightarrow \text{create}}^{0..1}, g_{\text{perc} \rightarrow \text{create}}^{0..1}} \Sigma_{\text{create}}^{(1)} \xrightarrow{\Lambda} \Sigma_{\text{refl}}^{(2)} \xrightarrow{V} \Sigma_{\text{plan}}^{(1)} } \]
\textit{Process Analysis:}
\begin{enumerate}
	\item Information from memory ($\Sigma_{\text{mem}}$) and perception ($\Sigma_{\text{perc}}$), potentially across concrete and abstract levels ($k=0,1$), influences a creative ideation sector $\Sigma_{\text{create}}^{(1)}$ (as per Appendix~\ref{app:sectors_abstraction_granularity}) via various intra-level couplings.
	\item A novel idea or hypothesis (e.g., $\varphi_{\text{new}}$) is generated in $\Sigma_{\text{create}}^{(1)}$.
	\item This idea is abstracted by $\Lambda$ to $\Sigma_{\text{refl}}^{(2)}$ for evaluation of novelty, plausibility, or potential.
	\item Promising evaluated ideas are elaborated by $V$ to $\Sigma_{\text{plan}}^{(1)}$ to develop plans for testing or utilizing the new idea.
\end{enumerate}

\subsection{Complex Multi-Sector, Multi-Level Integrative Loops}
These sequences demonstrate more intricate interactions involving multiple sectors and abstraction levels for complex tasks.

\subsubsection{Problem Solving with Internal Simulation and Iterative Refinement}

\label{app_seq:problem_solving_simulation}

This sequence describes problem-solving that involves internal simulation of planned strategies and iterative refinement based on simulated outcomes.
\begin{empheq}[box=\fbox]{align*}
	& \Sigma_{\text{goal}}^{(1)} \xrightarrow{g_{\text{goal} \rightarrow \text{plan}}^1} \Sigma_{\text{plan}}^{(1)} \xrightarrow{g_{\text{plan} \rightarrow \text{sim}}^1} \Sigma_{\text{sim}}^{(1)} \rightarrow \Sigma_{\text{perc (sim)}}^{(1)} \\
	& \qquad \xrightarrow{g_{\text{perc(sim)} \rightarrow \text{refl}}^1} \Sigma_{\text{refl}}^{(1)} \xrightarrow{g_{\text{refl} \rightarrow \text{plan}}^1} \Sigma_{\text{plan}}^{(1)} \xrightarrow{V} \Sigma_{\text{exe}}^{(0)} \rightarrow \text{Environment}
\end{empheq}
\textit{Process Analysis:}
\begin{enumerate}
	\item An abstract goal in $\Sigma_{\text{goal}}^{(1)}$ influences abstract planning in $\Sigma_{\text{plan}}^{(1)}$.
	\item The current abstract plan (or strategy) in $\Sigma_{\text{plan}}^{(1)}$ is provided as input to an internal simulation sector $\Sigma_{\text{sim}}^{(1)}$ (as per Appendix~\ref{app:sectors_abstraction_granularity}) via $g_{\text{plan} \rightarrow \text{sim}}^1$.
	\item An internal simulation ($\gamma_{\text{sim}}$) is run within $\Sigma_{\text{sim}}^{(1)}$. 
	\item The outcome of the simulation is "perceived" as if it were real, populating a simulated perception sector $\Sigma_{\text{perc (sim)}}^{(1)}$.
	\item This simulated outcome influences abstract reflection in $\Sigma_{\text{refl}}^{(1)}$ via $g_{\text{perc(sim)} \rightarrow \text{refl}}^1$ for evaluation.
	\item The evaluation influences $\Sigma_{\text{plan}}^{(1)}$ via $g_{\text{refl} \rightarrow \text{plan}}^1$, leading to refinement of the abstract strategy.
	\item The refined strategy is eventually elaborated by $V$ to $\Sigma_{\text{exe}}^{(0)}$ for execution in the actual Environment.
\end{enumerate}

\subsubsection{Ethical Deliberation Influencing Concrete Action (using $\Sigma_{\text{ethic}}$)}

\label{app_seq:ethical_deliberation_action}
The next sequence outlines how ethical considerations might be integrated into the planning process through abstraction and reflective evaluation.
\[ \boxed{ \Sigma_{\text{plan}}^{(0)} \xrightarrow{\Lambda} \Sigma_{\text{plan}}^{(1)} \xrightarrow{g_{\text{plan} \rightarrow \text{ethic}}^1} \Sigma_{\text{ethic}}^{(1)} \xrightarrow{g_{\text{ethic} \rightarrow \text{refl}}^1} \Sigma_{\text{refl}}^{(1)} \xrightarrow{V} \Sigma_{\text{refl}}^{(0)} \xrightarrow{g_{\text{refl} \rightarrow \text{plan}}^0} \Sigma_{\text{plan}}^{(0)} \rightarrow \Sigma_{\text{exe}}^{(0)} } \]
\textit{Process Analysis:}
\begin{enumerate}
	\item A proposed concrete action or plan is formed in $\Sigma_{\text{plan}}^{(0)}$.
	\item This plan is abstracted by $\Lambda$ to its abstract concept in $\Sigma_{\text{plan}}^{(1)}$.
	\item The abstract action concept influences an ethical reasoning sector $\Sigma_{\text{ethic}}^{(1)}$ (as per Appendix~\ref{app:sectors_abstraction_granularity}) via $g_{\text{plan} \rightarrow \text{ethic}}^1$ for evaluation against ethical principles.
	\item The outcome of ethical evaluation influences abstract reflection in $\Sigma_{\text{refl}}^{(1)}$ via $g_{\text{ethic} \rightarrow \text{refl}}^1$, forming a meta-belief about the ethicality of the plan.
	\item This meta-belief is elaborated by $V$ to $\Sigma_{\text{refl}}^{(0)}$ as a concrete judgment or injunction.
	\item The concrete judgment influences the original plan in $\Sigma_{\text{plan}}^{(0)}$ via $g_{\text{refl} \rightarrow \text{plan}}^0$, potentially gating or modifying it.
	\item The (potentially modified) final action is passed to $\Sigma_{\text{exe}}^{(0)}$ for execution.
\end{enumerate}

\subsubsection{Narrative Construction for Self-Understanding and Identity Update}
\label{app_seq:narrative_construction_identity}
This process describes how an agent might construct narratives from episodic memories to achieve self-understanding and update its core identity.
\[ \boxed{ \Sigma_{\text{mem}}^{(0)} \xrightarrow{\Lambda} \Sigma_{\text{mem}}^{(1)} \xrightarrow{g_{\text{mem} \rightarrow \text{narr}}^1} \Sigma_{\text{narr}}^{(1)} \xrightarrow{\Lambda} \Sigma_{\text{narr}}^{(2)} \xrightarrow{g_{\text{narr} \rightarrow \text{refl}}^2} \Sigma_{\text{refl}}^{(2)} \rightarrow \text{Update Epistemic Identity } \vec{\eta} } \]
\textit{Process Analysis:}
\begin{enumerate}
	\item A collection of concrete episodic memories from $\Sigma_{\text{mem}}^{(0)}$ is processed.
	\item These are abstracted by $\Lambda$ to $\Sigma_{\text{mem}}^{(1)}$, where generalized event patterns might be identified.
	\item These patterns influence a narrative construction sector $\Sigma_{\text{narr}}^{(1)}$ (as per Section~\ref{subsec:intro_foundation}) via $g_{\text{mem} \rightarrow \text{narr}}^1$, forming narrative segments.
	\item These narrative segments are further abstracted by $\Lambda$ and integrated into a broader life story or self-model in $\Sigma_{\text{narr}}^{(2)}$.
	\item This high-level narrative understanding influences highly abstract self-reflection in $\Sigma_{\text{refl}}^{(2)}$ via $g_{\text{narr} \rightarrow \text{refl}}^2$.
	\item The outcome is an abstract self-assessment or insight that contributes to updating the agent's Epistemic Identity $\vec{\eta}$ (\cite{Dumbrava2025TheoreticalFoundations}, Chapter 24).
\end{enumerate}

\subsubsection{Proactive Information Seeking Driven by Identified Knowledge Gaps}
\label{app_seq:proactive_info_seeking}
Finally, this sequence details how an agent might proactively seek information upon identifying a knowledge gap, involving goal setting, planning, and action.
\begin{empheq}[box=\fbox]{align*}
	& \Sigma_{\text{know}}^{(1)} \xrightarrow{g_{\text{know} \rightarrow \text{refl}}^1} \Sigma_{\text{refl}}^{(1)} \xrightarrow{g_{\text{refl} \rightarrow \text{goal}}^1} \Sigma_{\text{goal}}^{(1)} \xrightarrow{g_{\text{goal} \rightarrow \text{plan}}^1} \Sigma_{\text{plan}}^{(1)} \xrightarrow{V} \Sigma_{\text{exe}}^{(0)} \\
	& \qquad \rightarrow \text{Environment} \rightarrow \Sigma_{\text{perc}}^{(0)} \xrightarrow{\Lambda} \Sigma_{\text{perc}}^{(1)} \xrightarrow{g_{\text{perc} \rightarrow \text{know}}^1} \Sigma_{\text{know}}^{(1)}
\end{empheq}
\textit{Process Analysis:}
\begin{enumerate}
	\item The knowledge sector $\Sigma_{\text{know}}^{(1)}$ (as per Appendix~\ref{app:sectors_abstraction_granularity}) identifies a knowledge gap (e.g., missing information $\Delta\varphi$).
	\item This identification influences abstract reflection in $\Sigma_{\text{refl}}^{(1)}$ via $g_{\text{know} \rightarrow \text{refl}}^1$, leading to a belief like "I need information X".
	\item This reflective state influences the goal management sector $\Sigma_{\text{goal}}^{(1)}$ via $g_{\text{refl} \rightarrow \text{goal}}^1$, setting a goal to "Acquire information X".
	\item The new goal influences abstract planning in $\Sigma_{\text{plan}}^{(1)}$ via $g_{\text{goal} \rightarrow \text{plan}}^1$ to devise information-seeking strategies.
	\item The plan is elaborated by $V$ to $\Sigma_{\text{exe}}^{(0)}$ for concrete actions like executing a query or search.
	\item Action on the Environment occurs. 
	\item New information from the Environment is perceived by $\Sigma_{\text{perc}}^{(0)}$.
	\item This new information is abstracted by $\Lambda$ to $\Sigma_{\text{perc}}^{(1)}$.
	\item The abstracted new information influences the knowledge base $\Sigma_{\text{know}}^{(1)}$ via $g_{\text{perc} \rightarrow \text{know}}^1$, ideally filling the identified gap $\Delta\varphi$.
\end{enumerate}

\section{Continuous Coupling Propagation}
\label{app:continuous_coupling_propagation}

While Section~\ref{subsec:coupling_propagation} models hierarchical coupling propagation using a discrete, linear transformation $$\vec{G}^{(k+1)} = \mathbf{M}_k \vec{G}^{(k)},$$ it is conceptually useful to consider an idealized scenario where the abstraction level $k$ is a continuous variable. This allows for the application of differential equations to model the "flow" of the coupling profile $G^{(k)}$, providing a different lens through which to understand its dynamics. This appendix offers:
\begin{enumerate}
	\item a discussion of how continuous abstraction levels can be bridged with "cognitive beta functions" to model coupling propagation (Subsection~\ref{app_subsec:cognitive_beta_functions});
	\item a discussion of hypotheses 
	\begin{enumerate}
		\item concerning fixed-point profiles $G^*$ representing scale-invariant features in coupling propagation,
		\item and concerning long-range behavior of couplings in the limit $k \rightarrow \infty$ (Subsection~\ref{app_subsec:fixed_points_continuous});
	\end{enumerate}
	\item and a discussion of "relevant" and "irrelevant" couplings that characterizes the behavior of couplings in the limit $k \rightarrow \infty$ (Subsection~\ref{app_subsec:relevant_irrelevant_continuous}).
	
\end{enumerate}

\subsection{Conceptualizing Cognitive "Beta Functions" ($\beta_{ij}(G^{(k)})$)}
\label{app_subsec:cognitive_beta_functions}

If we treat the abstraction level $k$ as continuous, the propagation of a specific intra-level coupling constant $g_{ij}^{(k)}$ can be described by a differential equation:

\begin{equation*}
	\frac{d g_{ij}^{(k)}}{d k} = \beta_{ij}(G^{(k)}).
\end{equation*}
This allows for the calculation of the coupling at any level $k$ via integration:

\begin{equation*}
	g^{(k)}_{ij} = g^{(0)}_{ij} + \int_{0}^{k} \beta_{ij}(G^{(\kappa)}) \, d\kappa.
\end{equation*}
Here, $\beta_{ij}$ is the \textbf{"cognitive beta function"} for the specific coupling $g_{ij}$.
Its value depends on the entire set of couplings $G^{(k)}$ at that scale, encapsulating how the interactions between all sectors collectively influence the evolution of one particular coupling's strength as abstraction increases.

\subsection{The Fixed Point Hypothesis}
\label{app_subsec:fixed_points_continuous}

Within this continuous framework, a \textbf{fixed-point profile ($G^*$)} is a configuration where the flow of all couplings ceases.
This occurs when the beta functions for all couplings are simultaneously zero:

\begin{equation*}
	\beta_{ij}(G^*) = 0 \quad \forall i,j.
\end{equation*}
At such a fixed point, further abstraction no longer changes the effective profile of sectoral coupling strengths.
These $G^*$ profiles represent scale-invariant features of the agent's cognitive architecture.

\bigskip
\textbf{Coupling Convergence Hypothesis.} We posit that, \textbf{as the abstraction level $k$ tends towards infinity, all intra-level sectoral influences vanish}:
\[\boxed{\lim_{k \rightarrow \infty} g^k_{ij} = 0.}\]
That is, the only stable fixed point that the cognitive architecture approaches at its highest levels is $$G^* = \{0\}.$$
This implies that at the ultimate level of abstract thought, all specialized cognitive functions become fully decoupled, and processing is no longer characterized by inter-sectoral dynamics. While other, non-trivial fixed points where specific couplings retain finite values might exist as transient or unstable features at intermediate scales, our hypothesis posits that they do not represent the final, scale-invariant state of the system. While a formal proof is outside our scope, the hypothesis is motivated by fundamental physical constraints presented below.

\bigskip

\textit{Coupling Convergence Hypothesis Justification: Abstract thought is bounded in a physical system.} 
Let us test our hypothesis by supposing that there exists an agent capable of unbounded abstract thought.

Such a capacity would require either infinite resources or infinite time. The first case, requiring infinite memory or energy, is physically impossible; by the principle of mass-energy equivalence, it implies infinite mass, which would collapse the agent into a black hole. The second case, requiring the agent to persist for infinite time, is also impossible; to counteract the relentless increase of entropy, the agent would need to function as a perpetual motion machine of the second kind, a construct forbidden by the Second Law of Thermodynamics.

Since both scenarios violate fundamental physical constraints, the initial premise is untenable. Therefore, any physically realizable agent must be incapable of unbounded abstract thought. This physical limitation implies that the hierarchy of abstraction layers must have an effective upper bound, a level $N$ beyond which the agent can maintain no belief fragments: $$\phi^{(k>N)} = \emptyset.$$ In these empty, inaccessible layers, the sectors contain no content and thus can exert no influence. Consequently, \textbf{all intra-level couplings thereof must be trivially zero}: $$g_{ij}^{(k>N)}=0.$$

\subsection{Relevant and Irrelevant Couplings}
\label{app_subsec:relevant_irrelevant_continuous}

The sign of the beta function determines the fate of a coupling as abstraction ($k$) increases:

\begin{itemize}
	\item \textbf{Irrelevant Couplings}: If $\beta_{ij} < 0$ (for a positive $g_{ij}$), the coupling $g_{ij}^{(k)}$ will decrease as $k$ increases.
	These are fine-grained interactions whose effects are localized to lower levels and "wash out" at coarser scales of cognition.
	
	\item \textbf{Relevant Couplings}: If $\beta_{ij} > 0$ (for a positive $g_{ij}$), the coupling $g_{ij}^{(k)}$ grows with $k$.
	These couplings represent interaction patterns that become increasingly dominant in high-level, strategic, or abstract thought processes.
	
\end{itemize}
Our central hypothesis---that all couplings ultimately decay to zero---has a profound implication for these roles. For any coupling $g_{ij}^{(k)}$ to vanish as $k \rightarrow \infty$, its corresponding beta function $\beta_{ij}$ must, at least for sufficiently large $k$, be negative (for positive $g_{ij}$) or positive (for negative $g_{ij}$) to drive the coupling towards the trivial fixed point at $g_{ij}=0$.

This does not preclude couplings from being "relevant" and growing across certain ranges of abstraction (e.g., from concrete to tactical levels). However, it implies that any such growth must eventually be counteracted, with all couplings ultimately becoming "irrelevant" to ensure they decay towards zero. Under this hypothesis, the macroscopic cognitive organization of a complex agent is governed by a hierarchy of interactions that progressively fade, leaving only decoupled, abstract principles at the highest scale.

\section{Experimental Strategies for G-Map Development}
\label{app:g_mapper_strategies}

The G-Map, introduced in Section~\ref{subsec:future_learning_g}, is a conceptual neural network designed to learn the complex transformation from an agent's foundational neural network weights $W$ to its emergent, high-level intra-level sectoral coupling profile $G$. The successful development of such a G-Map would be pivotal for enabling "meta-tuning"---a principled approach to adjusting an agent's underlying weights $W$ to achieve a desired functional profile $G_{\text{target}}$. This appendix outlines:
\begin{enumerate}
	\item leveraging concepts such as gauge equivalence in agent architectures for G-Map development (Subsection~\ref{app_subsec:g_mapper_gauge_equivalence});
	\item potential experimental strategies for the development, training, and analysis of a G-Map (Subsection~\ref{app_subsec:g_mapper_experimental_phases});
	\item specific theoretical hurdles to overcome during G-Map development (Subsection~\ref{app_subsec:g_mapper_challenges});
	\item and the outcomes that can be expected from pursuing the development of a G-Map (Subsection~\ref{app_subsec:g_mapper_outcomes}).
\end{enumerate}
potential experimental strategies for the development, training, and analysis of a G-Map, including 

\subsection{Conceptual Basis: Gauge Equivalence and G-Map}
\label{app_subsec:g_mapper_gauge_equivalence}
In the context of this framework, we define \textbf{operational gauge equivalence} as a scenario where multiple distinct foundational neural network weight configurations $$W_1, W_2, \dots, W_N$$ for an agent all result in the same (or functionally indistinguishable) emergent intra-level coupling profile $G$. That is, if $$W_a \sim W_b$$ (denoting gauge equivalence), then it is expected that $$\text{G-Map}(W_a) \approx \text{G-Map}(W_b) \approx G_{\text{shared}}.$$ For supplementary reading on gauge equivalence, see \cite{Dumbrava2025TheoreticalFoundations} (Chapter 23).

This concept implies that the G-Map captures a functionally salient representation of the agent's cognitive interaction patterns irrespective of certain micro-architectural variations in $W$. This has significant implications for data generation: if one instance of $(W_i, G_{\text{shared}})$ is known, and methods can be found to generate or identify other $W_j$ that are gauge-equivalent to $W_i$, this can augment the dataset for training the G-Map with multiple $(W, G_{\text{shared}})$ pairs derived from a single known functional profile.

\subsection{Experimental Phases for G-Map Development}
\label{app_subsec:g_mapper_experimental_phases}
The development of a G-Map can be envisioned in three primary phases:

\subsubsection{Phase 1: Dataset Generation --- Leveraging Target Profiles and Gauge Equivalence}
\label{app_subsubsec:g_mapper_phase1_dataset}
The generation of a robust dataset of ($W, G$) pairs is the most critical prerequisite.

\begin{itemize}
	\item \textbf{Step J.1.1: Defining Target $G_{\text{target}}$ Profiles.}
	Establish a diverse set of target coupling profiles $G_{\text{target}}$. These profiles could represent:
	\begin{itemize}
		\item Hypothesized "healthy" or archetypal cognitive styles (e.g., "Reflective/Deliberative" or "Reactive" as per Section~\ref{subsec:cognitive_styles}).
		\item Profiles associated with specific optimized task performances.
		\item Profiles exhibiting known "pathological" dynamics (e.g., excessive rumination due to strong recurrent $g_{ii}^k$ as per Section~\ref{subsec:conceptualizing_pathways}).
		\item Systematically varied profiles (e.g., incrementally changing specific $g_{ij}^k$ values to observe effects).
	\end{itemize}
	
	\item \textbf{Step J.1.2: Engineering Foundational Architectures ($W$) for $G_{\text{target}}$.}
	For each $G_{\text{target}}$, one or more foundational weight configurations $W$ must be obtained. This might involve:
	\begin{itemize}
		\item \textit{Direct Design \& Heuristics:} For simple agents or subsystems, hand-crafting $W$ to approximate $G_{\text{target}}$.
		\item \textit{Iterative Agent Training:} Training agents (e.g., using reinforcement learning) on tasks hypothesized to induce a specific $G_{\text{target}}$, then estimating $G_{actual}$ from the learned $W$.
		\item \textit{Evolutionary Algorithms:} Searching $W$-space with a fitness function rewarding proximity to $G_{\text{target}}$.
	\end{itemize}
	
	\item \textbf{Step J.1.3: Exploiting Gauge Equivalence for Data Augmentation.}
	If a configuration $W_0$ is known/estimated to produce $G_{\text{target}}$, attempt to find other configurations $W_1, \dots, W_N$ gauge-equivalent to $W_0$. This could involve:
	\begin{itemize}
		\item Identifying symmetries or transformations in $W$-space hypothesized to leave $G$ invariant.
		\item Applying "neutral" network mutations or re-parameterizations to $W_0$.
	\end{itemize}
	This step, while powerful, requires significant theoretical insight into the $W \mapsto G$ relationship.
	
	\item \textbf{Step J.1.4: Estimation and Verification of $G_{actual}$ from $W_i$.}
	For each $W_i$, estimate the actual emergent $G_{actual}$ using methodologies from Section~\ref{sec:measuring_g}. The pair $(W_i, G_{actual})$ is added to the training set.
\end{itemize}

\subsubsection{Phase 2: G-Map Training}
\label{app_subsubsec:g_mapper_phase2_training}
\begin{itemize}
	\item \textbf{Step J.2.1: Model Architecture for G-Map.}
	Design a neural network architecture for the G-Map that takes $W$ (or a suitable representation) as input and outputs $G$. This might involve architectures capable of handling graph-structured input (if $W$ is from a GNN-based agent) or very high-dimensional vectors, and outputting a structured tensor representing $$G = \{g_{ij}^k\}.$$
	
	\item \textbf{Step J.2.2: Training Process and Loss Function.}
	Train the G-Map on ($W, G_{actual}$) pairs. The loss function would measure $$L = \| \text{G-Map}(W) - G_{actual} \|,$$ potentially using metrics like the Frobenius norm for the $G^{(k)}$ matrices or other structural similarity measures.
\end{itemize}

\subsubsection{Phase 3: G-Map Validation and Analysis}
\label{app_subsubsec:g_mapper_phase3_validation}
\begin{itemize}
	\item \textbf{Step J.3.1: Predictive Accuracy on Hold-Out Data.}
	Evaluate the G-Map's ability to predict $G$ for unseen $W$ configurations.
	
	\item \textbf{Step J.3.2: Analysis of Sensitivity to Perturbations in $W$-space.}
	Use the trained G-Map to study $$G_{\text{pred}} = \text{G-Map}(W_0 + \Delta W).$$ This can identify critical components in $W$ and regions of stability or high sensitivity in the $W \mapsto G$ map.
	
	\item \textbf{Step J.3.3: Exploring Interpolations in $W$-space and their Effect on $G$.}
	Given $(W_1, G_1)$ and $(W_2, G_2)$, analyze $$G_{\text{pred}}(W_{\alpha} = \alpha W_1 + (1-\alpha) W_2).$$ Does this $G_{\text{pred}}(W_{\alpha})$ correspond to a meaningful interpolation of cognitive characteristics?
	
	\item \textbf{Step J.3.4: Testing Consistency with Known Gauge Equivalences.}
	If $(W_a, G_s)$ and $(W_b, G_s)$ are known gauge-equivalent pairs, verify that $$\text{G-Map}(W_a) \approx \text{G-Map}(W_b) \approx G_s.$$
\end{itemize}

\subsection{Specific Experimental Considerations and Challenges}
\label{app_subsec:g_mapper_challenges}
\begin{itemize}
	\item \textbf{Practical Generation/Identification of Gauge-Equivalent $W$ Configurations:} This remains a significant theoretical hurdle, crucial for effective data augmentation.
	\item \textbf{High-Dimensionality of $W$ and $G$:} Representing and processing these high-dimensional objects efficiently is challenging. Dimensionality reduction or feature selection techniques may be necessary.
	\item \textbf{Defining Appropriate Metrics for G-Map Comparison:} Beyond element-wise differences (e.g., Frobenius norm), metrics capturing functional or structural similarity of $G$ profiles (e.g., based on graph-theoretic properties of $G^{(k)}$ for each layer) are needed.
	\item \textbf{Computational Resources:} Simulating agents to generate $G_{actual}$ for many $W_i$, and subsequently training a potentially large G-Map network, will require substantial computational power.
	\item \textbf{Iterative Refinement Loop:} The estimation of $G_{actual}$ (for dataset generation) and the training of the G-Map might be an iterative process, where improvements in one facilitate advances in the other.
\end{itemize}

\subsection{Expected Outcomes and Contributions}
\label{app_subsec:g_mapper_outcomes}
These experimental strategies aim to:
\begin{itemize}
	\item Provide a practical, albeit challenging, pathway towards developing and validating the G-Map.
	\item Offer empirical insights into the $W \mapsto G$ mapping, potentially revealing how high-level functional cognitive architectures emerge from lower-level network parameters.
	\item Test the utility of the gauge equivalence concept for understanding and modeling AI cognitive architectures.
	\item Pave the way for the "meta-tuning" paradigm, contributing to the principled design of AI systems with more predictable, interpretable, and potentially safer cognitive profiles.
\end{itemize}
The successful execution of such experiments would represent a significant step in transforming the theoretical G-Map into an empirically grounded and practically applicable tool for AI research and development.

\bibliographystyle{plain}
\bibliography{references}

\begin{thebibliography}{1}

\bibitem{Dumbrava2025BeliefFiltering}
Sebastian Dumbrava.
\newblock Belief filtering for epistemic control in linguistic state space.
\newblock {\em arXiv preprint arXiv:2505.04927}, 2025.

\bibitem{Dumbrava2025BeliefInjection}
Sebastian Dumbrava.
\newblock Belief injection for epistemic control in linguistic state space.
\newblock {\em arXiv preprint arXiv:2505.07693}, 2025.

\bibitem{Dumbrava2025TheoreticalFoundations}
Sebastian Dumbrava.
\newblock Theoretical foundations for semantic cognition in artificial
  intelligence.
\newblock {\em arXiv preprint arXiv:2504.21218}, 2025.

\end{thebibliography}

\end{document}